\documentclass[
  superscriptaddress,
  amsmath,
  amssymb,
  aps,
  reprint,
  prx,
  twocolumn,
  nofootinbib,
]{revtex4-2}

\usepackage{caption}
\usepackage[utf8]{inputenc}
\usepackage[T1]{fontenc}

\usepackage{amsmath, amsfonts, amssymb, amsthm, bm, bbm, mathtools}
\usepackage{physics}
\usepackage{upgreek}
\usepackage{dsfont}

\usepackage{array, booktabs, tabulary, dcolumn}

\usepackage{graphicx}

\usepackage{xcolor}
\usepackage{color}
\usepackage[colorlinks]{hyperref}
\usepackage{soul}

\usepackage{algorithm}
\usepackage{algpseudocode}

\usepackage[toc,page]{appendix}
\usepackage{natbib}

\DeclareRobustCommand{\rev}[1]{\textcolor{black}{#1}}

\newenvironment{revblock}
    {\par\begingroup\color{black}}
    {\par\endgroup}

\newcommand{\bea}{\begin{eqnarray}}
\newcommand{\eea}{\end{eqnarray}}

\newcommand{\pgt}{G_2}
\newcommand{\pgo}{G_1}
\newcommand{\pmo}{M_1}
\newcommand{\pmt}{M_2}

\newcommand{\be}{\begin{equation}\begin{aligned}}
\newcommand{\ee}{\end{aligned}\end{equation}}
\def\eqref#1{equation~\ref{#1}}

\usepackage{lineno}
\usepackage{caption}
\usepackage{subcaption}
\begin{document}
\title[]{Distinct mechanisms underlying in-context learning in transformers}

\author{Cole Gibson}
\thanks{Authors contributed equally}
\affiliation{Lewis-Sigler Institute for Integrative Genomics, Princeton University, Princeton, NJ 08544, USA}
\affiliation{Department of Physics, Princeton University, Princeton, NJ 08544, USA}
\author{Wenping Cui}
\thanks{Authors contributed equally}
\affiliation{Department of Physics, Princeton University, Princeton, NJ 08544, USA}
\author{Gautam Reddy}
\email{greddy@princeton.edu}
\affiliation{Department of Physics, Princeton University, Princeton, NJ 08544, USA}

\begin{abstract}
Modern distributed networks, notably transformers, acquire a remarkable ability (termed `in-context learning') to adapt their computation to input statistics, such that a fixed network can be applied to data from a broad range of systems. Here, we provide a mechanistic characterization of this behavior in transformers trained on a finite set $\mathcal{S}$ of discrete Markov chains. The transformer displays four algorithmic phases, characterized by whether the network memorizes or generalizes, and whether it uses 1-point or 2-point statistics.
We show that the four phases are implemented by multi-layer subcircuits that exemplify two qualitatively distinct mechanisms for implementing context-adaptive computations. Minimal models isolate the key features of both motifs. Memorization and generalization phases are delineated by two boundaries that depend on data diversity, $K = |\mathcal{S}|$. The first ($K_1^\ast$) is set by a kinetic competition between subcircuits and the second ($K_2^\ast$) is set by a representational bottleneck. A symmetry-constrained theory of a transformer's training dynamics explains the sharp transition from 1-point to 2-point generalization and identifies key features of the loss landscape that allow the network to generalize. Put together, we show that transformers develop distinct subcircuits to implement in-context learning and identify conditions that favor certain mechanisms over others. 
\end{abstract}

\maketitle

\section{Introduction}
Traditionally, learning involves tuning parameters so that a learning system encodes statistical regularities of a particular dataset. A recurrent network trained on trajectories from one dynamical system, for example, internalizes that system's correlations and then extrapolates using these statistics. Presenting it with trajectories from a different system typically degrades the accuracy with which it can predict future states.  Modern machine learning systems, most notably transformers~\cite{vaswani2023attentionneed}, display an alternative mode of adaptation once trained on a large dataset. When given a partial sequence or a short list of example input-output pairs, a network is often able to infer the rule that generated the data and apply it to a new input without additional tuning of its parameters. That is, the burden of learning shifts from slow and often expensive parameter updates to a fast computation implemented by network dynamics. This capacity to generalize has been termed `in-context learning (ICL)'~\cite{brown2020language, garg2022can,chan2022data,olsson2022incontextlearninginductionheads,kirsch2022general, reddy2023mechanistic, von2023uncovering, bietti2023birthtransformermemoryviewpoint, ortega2019meta}. While this feature was first highlighted in natural language processing \cite{brown2020language}, similar behavior has been recapitulated across diverse paradigms, including nonlinear regression, small tabular datasets, zero-shot forecasting of chaotic dynamical systems, imitation learning and control systems~\cite{garg2022can, von2023transformers, ahn2023transformers, zhang2023trained,hollmann2022tabpfn,zhang2024zero,bao2026transformers,fu2024context, laskin2022context, lee2023supervised, moeini2025survey}, in turn motivating the development of so-called foundation models across many domains of science~\cite{bommasani2021opportunities,brixi2026genome,  zhang2024comprehensive, moor2023foundation, subramanian2023towards, pyzer2025foundation, dalla2025nucleotide}. 

A natural question is if there are particular features of data or network architecture that favor such context-adaptive computations. Previous work has shown that inducing this ability depends crucially on the diversity and scale of data the model encounters during training~\cite{chan2022data,kirsch2022general,raventos2023pretraining,wu2023many, reddy2023mechanistic, nguyen2024differential, park2024competition, wurgaft2025context, lu2024asymptotic, goddard2025can}. It is not known what sets this scale, though two distinct hypotheses have been proposed \cite{nguyen2024differential, lu2024asymptotic}. 

Why might transformers, in particular, support this behavior? A transformer operates in parallel on an input sequence of vectors~\cite{vaswani2023attentionneed}. The attention operation mediates interactions by comparing each vector to the rest of the vectors in the sequence. These comparisons let the network retrieve relevant vectors from the context. Retrieved vectors are subsequently acted upon by a nonlinear operation parameterized as a multi-layer perceptron (MLP). 
One hypothesis is that the transformer leverages retrieved information in earlier layers of the network to adapt its computations in later layers based on sequence-level statistics, but it is not known how different transformer elements interact to enable this behavior. 

\begin{figure*}[ht!]
\begin{nolinenumbers}
\centering
\includegraphics[width=0.95\textwidth]{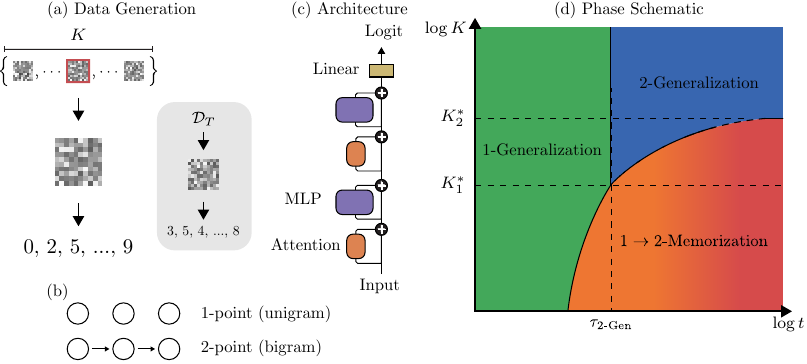}
    \caption{(a) Data generating process for training and \rev{generalization} sequences. Training sequences are generated from transition matrices drawn from the $K$ matrices in $\mathcal{S}$ sampled prior to training. \rev{Generalization} sequences are generated from transition matrices drawn directly from $\mathcal{D}_T$. (b) An illustration of 1-point (unigram) statistics and 2-point (bigram) directed, nearest-neighbor statistics. (c) The 2-layer transformer architecture, which contains one attention block and one multi-layer perceptron (MLP) block in each layer. After the operation of two layers, the linear block maps the embedded representation to a discrete distribution over states that is then normalized to form the logit. (d) Schematic of the four algorithmic phases as the number of chains ($K$) and training time ($t$) are varied. Phases are consistent with either memorizing or generalizing predictors, which use either 1-point or 2-point statistics, denoted by 1- and 2- prefixes respectively. Orange to red gradient in memorization phase indicates a gradual transition from $\pmo$ to $\pmt$ in contrast to the abrupt transition from $\pgo$ to $\pgt$. Three quantities characterize the transition between the different phases: (i) $\tau_{\text{2-Gen}}$, which is the training time to transition from $\pgo$ to $\pgt$ given $K > K_1^*$, (ii) $K_1^*$, which together with $\tau_{\text{2-Gen}}$ defines the `triple point' between $\pgo$, $\pgt$ and $\pmo$, and (iii) $K_2^*$, which delineates $\pgt$ and $\pmt$. The dashed phase boundary indicates an extrapolation as $t\to\infty$.}
    \label{fig:Background}
\end{nolinenumbers}
\end{figure*}

\section{Background}
We consider a setting that has been empirically shown to capture many key features of ICL observed across synthetic data and natural language paradigms~\cite{park2024competition, wurgaft2025context, chan2022data, raventos2023pretraining, kirsch2022general, lin2024dual} \rev{including induction heads~\cite{olsson2022incontextlearninginductionheads,edelman2024evolutionstatisticalinductionheads, reddy2023mechanistic}, learning kinetics~\cite{nguyen2024differential}, and transient generalization~\cite{singh2023transient}}. In this setting~\cite{park2024competition}, the transformer is provided an input sequence of states sampled from a finite set $\mathcal{S}$ of $K$ stationary Markov chains, where each chain is over $C$ discrete states. The network must then output a predictive distribution over the next state. The $K$ transition matrices are drawn once from a symmetric Dirichlet ensemble $\mathcal{D}_T$ and frozen thereafter (Figure~\ref{fig:Background}a) (further details in Appendix~\ref{sec:data_gen}). $K$ serves as a quantitative measure of data diversity. 

Since $K$ is finite, it is possible for the network to develop distinct generalization and memorization strategies depending on $K$. In one case, 
the network may predict the next state by simply reproducing the empirical statistics of the presented sequence.
This strategy would allow the network to generalize to sequences drawn from Markov chains not in $\mathcal{S}$. Because Markov chains are characterized by transition statistics between neighboring states (Figure~\ref{fig:Background}b), the network does not benefit from using correlations of order greater than two. We term the two such generalizing strategies that use 1-point or 2-point information as 1-Gen and 2-Gen respectively, and the corresponding phases $\pgo$ and $\pgt$. 

Alternatively, the network may use the sequence context to identify which of the $K$ chains the sequence is being sampled from and retrieve the corresponding transition probabilities, even if the network never encounters the same sequence during training. Identification is more sample efficient than generalization (i.e., needs a shorter sequence to predict well), but requires the network to encode each of the $K$ transition matrices in its parameters. The network may use 1-point or 2-point statistics (and possibly intermediate strategies) to identify the chain. We term these two memorizing strategies 1-Mem and 2-Mem respectively, and the corresponding phases $\pmo$ and $\pmt$.  

We consider a two-layer transformer with one attention block and one MLP block per layer followed by a linear readout (Figure~\ref{fig:Background}c, further details in Appendix~\ref{sec:transformerinfo}). This is the simplest network architecture that has been empirically shown to implement all four strategies across different $K$ \cite{park2024competition}. An attention head allows a state at one position in the sequence to retrieve information about other states (known as values) either based on their position or their content (by implementing a key-query dot product). An MLP is a feedforward neural network that processes the information produced by earlier blocks. Information from attention heads and MLPs is written into a residual stream (vertical path from input to logit in Figure~\ref{fig:Background}c), which tracks the $D$-dimensional representation of a state as it evolves over the network's layers due to the attention and MLP operations. We fix $C = 10$ and $D = 64$ (refer to Appendix~\ref{sec:summaryvariables} for a complete summary of values). The network is trained autoregressively to predict the next state $s_{n+1}$ after observing a sequence of discrete states $s_0,s_1,\dots,s_n$ up to a maximum sequence length $N$.

We now summarize known phenomenology for two-layer transformers trained on different numbers of chains $K$ over the course of training (illustrated in Figure~\ref{fig:Background}d) \rev{\cite{park2024competition}}.

Soon after training begins, the network learns to implement 1-Gen (phase \(\pgo\), green region in Figure~\ref{fig:Background}d). When $K$ is below a certain threshold value (denoted $K_1^*$), the network transitions from $\pgo$ to the memorization phases, first learning 1-Mem in phase $\pmo$ and then gradually implementing 2-Mem in phase $\pmt$ (orange-red regions in Figure~\ref{fig:Background}d). In both $\pmo$ and $\pmt$, the network does not generalize to sequences sampled from a chain not in $\mathcal{S}$. When $K > K_1^*$, the network follows a different trajectory and abruptly transitions from $\pgo$ by rapidly learning 2-Gen (phase \(\pgt\), blue region in Figure~\ref{fig:Background}d). In this phase, the network optimally generalizes to sequences from unseen chains. The number of iterations after which the \(\pgt\) generalization phase appears, $\tau_{\text{2-Gen}}$, is independent of $K$ provided $K > K_1^*$.

\section{Summary of Contributions}
We present four main results that together identify the circuits that implement each of the aforementioned phases and the factors that govern the transitions between them.


First, we use a circuit tracing technique to identify the four subcircuits. 
Notably, 2-Mem is implemented by a novel encoder-pool-decoder subcircuit, which builds a latent representation of a chain and uses this representation to modulate network response in later layers. This circuit motif requires both pairs of attention heads and MLPs, which explains why a two-layer transformer is the minimal transformer necessary to recapitulate distinct algorithmic regimes.

Second, we characterize the sharp transition from \(\pgo\) to \(\pgt\) (as \(K \rightarrow \infty\)). To do so, we introduce the symmetry-constrained attention-only transformer (the \emph{SA-transformer}) architecture, which exploits permutation symmetry in task structure to significantly simplify the standard transformer. We analyze the SA-transformer to develop an effective theory for the learning dynamics of the statistical induction head underpinning \(\pgt\). The theory explains its abrupt formation, identifies two symmetry-breaking biases that guide the optimization dynamics, and predicts a scaling relationship between the formation time \(\tau_\text{2-Gen}\) and the sequence length \(N\) which we verify empirically with a standard transformer.



\begin{revblock}
Third, we characterize the memorization-generalization transition at $K = K_1^\ast$ as resulting from a kinetic competition between subcircuits.

Finally, we provide evidence of a second threshold, $K_2^\ast$, and suggest its mechanistic origin. For $K_1^* < K < K_2^*$, the interval $\Delta \tau_K = \tau_{\text{2-Mem}} - \tau_{\text{2-Gen}}$ between the initial appearance of $\pgt$ until it disappears in favor of $\pmt$ is observed to scale as a power law, $\Delta \tau_K \sim (K_2^* - K)^{-\gamma}$, with an exponent $\gamma \approx 2$. Thus, we hypothesize that this threshold marks the upper boundary of $\pmt$: when $K_1^* < K < K_2^*$, the network eventually memorizes and transitions from $\pgt$ to $\pmt$ with sufficient training, whereas when $K > K_2^*$, the network remains in $\pgt$ indefinitely (Figure~\ref{fig:Background}d). The transient behavior of $\pgt$ in the former case can then be interpreted as a form of overfitting.

An analysis of chain-level memorization suggests that the model cannot reliably encode and retrieve any of the $K$ chains when \(K > K_2^\ast\). Thus, we quantify how $K_2^*$ depends on the expressivity of the encoder and decoder via a minimal model of the 2-Mem subcircuit. We show that this circuit motif is also capable of 2-Gen within specific architectural regimes, thereby demonstrating that transformers may in principle implement two distinct in-context mechanisms that achieve optimal generalization.

\end{revblock}

\section{Results}


\begin{figure*}[ht!]
\begin{nolinenumbers}
\centering
\includegraphics[width=0.75\textwidth]{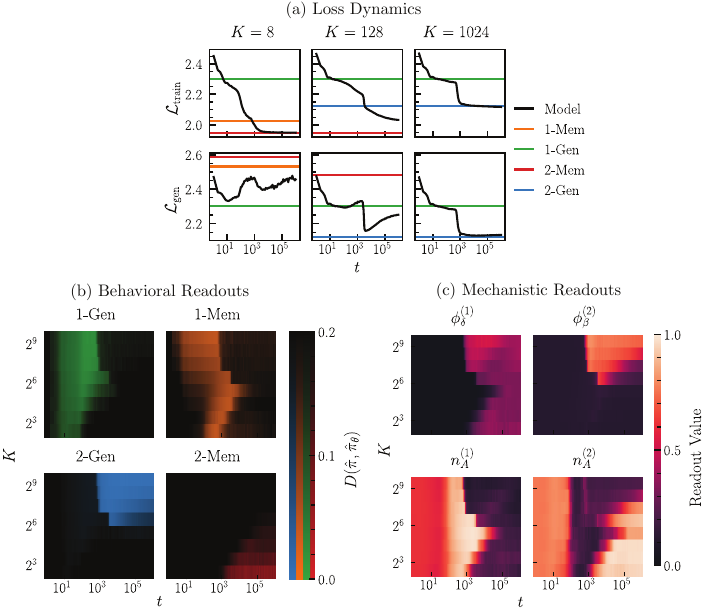}
    \caption{(a) Training dynamics of the transformer training loss (top) and generalization loss (bottom) at low, medium, and high $K$. Notice the plateaus during training that occur near the predictor training and \rev{generalization} losses. The model dynamics and predictor losses have been produced by evaluating model checkpoints and the predictors on a large batch of sequences. We have omitted for clarity the predictor losses near which each model does not exhibit plateaus. (b) Behavioral readouts of the transformer through training across a range of $K$. Color intensity is inversely proportional to the divergence (in nats) of the transformer predictions from the indicated predictor (Appendix \ref{sec:behave_readouts}). Simultaneous similarity of the model to both 1-Gen and 1-Mem predictors at large $K$ results from the relatively high similarity of these predictors to the uniform predictor over states. (c) Mechanistic readouts and order parameters of the transformer through training across a range of $K$ (Appendix \ref{sec:mech_readouts} for details). Here, \(n_A^{(1)}\) and \(n_A^{(2)}\) have been normalized by the sequence length, \(N=256\). The correspondence between phases and mechanistic readouts are as follows: \(\pgo\) and \(\pmo\): high \(n_A^{(1)}\) and low \(\phi_\delta^{(1)}\); \(\pgt\): high \(\phi_\delta^{(1)}\) and \(\phi_\beta^{(2)}\); \(\pmt\): high \(\phi_\delta^{(1)}\) and \(n_A^{(2)}\).}
    \label{fig:Phases}
\end{nolinenumbers}
\end{figure*}

\subsection{Identifying in-context memorization and generalization regimes}\label{sec:phase_identification}
\rev{We begin by reproducing the algorithmic phases reported in \cite{park2024competition} and illustrated in Figure~\ref{fig:Background}d.}
To determine which algorithm the network implements at a given training time ($t$) and number of chains in $\mathcal{S}$ ($K$), we compare its predictions to four Bayes predictors corresponding to the four strategies (Appendix~\ref{app:4phases}). The two generalizing predictors assume knowledge only of $\mathcal{D}_T$: $\hat\pi^{\text{1-Gen}}$ predicts from empirical 1-point frequencies, whereas $\hat\pi^{\text{2-Gen}}$ predicts from empirical nearest-neighbor 2-point frequencies. The two memorizing predictors instead assume knowledge of $\mathcal{S}$
: $\hat\pi^{\text{1-Mem}}$ and $\hat\pi^{\text{2-Mem}}$ infer which of the $K$ training chains generated the context using 1-point or 2-point statistics respectively, and predict the next state with the corresponding transition matrix. These four predictors isolate the two central choices faced by the network: memorization versus generalization, and 1-point versus 2-point statistics. Because the data are first-order Markovian, $\hat\pi^{\text{2-Gen}}$ is the optimal generalizing solution on $\mathcal{D}_T$, whereas $\hat\pi^{\text{2-Mem}}$ is the optimal memorizing solution on $\mathcal{S}$. 

The network passes through discrete strategies during training, which manifest as plateaus in the training loss near the loss of certain predictors (Figure~\ref{fig:Phases}a, upper). At low $K$, the model quickly improves and\rev{,} after a transient 1-Mem stage (left panel, brief flattening of black curve near orange line)\rev{,} converges to the optimal memorizing 2-Mem loss (convergence to red line). At high $K$, the model instead undergoes an abrupt drop to the loss incurred by 2-Gen (right panel, steep drop of black curve to blue line). Intermediate $K$ displays both features, that is, the model can transiently generalize (middle panel, steep drop of black curve to blue line) and later continue lowering the training loss by overfitting to $\mathcal{S}$ (convergence towards red line). The generalization loss (Figure~\ref{fig:Phases}a, lower) displays behavior consistent with these interpretations.

We characterize these transitions in more detail by measuring the similarity between the model's output $\hat\pi_\theta$ and the expected output from each predictor $\hat\pi$. This similarity is measured using
\begin{equation}
D(\hat\pi,\hat\pi_\theta)
\equiv
\frac{1}{2}\Big\langle D_{\mathrm{KL}}^{S_N}(\hat\pi\Vert\hat\pi_\theta)\Big\rangle_{\mathcal{S}}
+
\frac{1}{2}\Big\langle D_{\mathrm{KL}}^{S_N}(\hat\pi\Vert\hat\pi_\theta)\Big\rangle_{\mathcal{D}_T},
\end{equation}
where $\langle\cdot\rangle_{\mathcal{S}}$ and $\langle\cdot\rangle_{\mathcal{D}_T}$ denote averages over sequences drawn from $\mathcal{S}$ and $\mathcal{D}_T$, respectively, and $D_{\mathrm{KL}}^{S_N}$ is the autoregressive KL divergence averaged over sequence positions (definitions are given in Appendix \ref{app:phase_idenfication}). Figure~\ref{fig:Phases}b plots this divergence across $t$ and $K$. Early in training, model output across essentially all $K$ is closest to $1$-Gen. For small $K$, the model's output is consistent with a transition to $\pmo$, and then gradually to $\pmt$. For sufficiently large $K$, the network bypasses memorization and transitions sharply into $\pgt$. Between these limits is a window in which $\pgt$ appears transiently before eventually drifting into $\pmt$.

\begin{figure*}
\centering
\begin{nolinenumbers}
\includegraphics[width=0.95\linewidth]{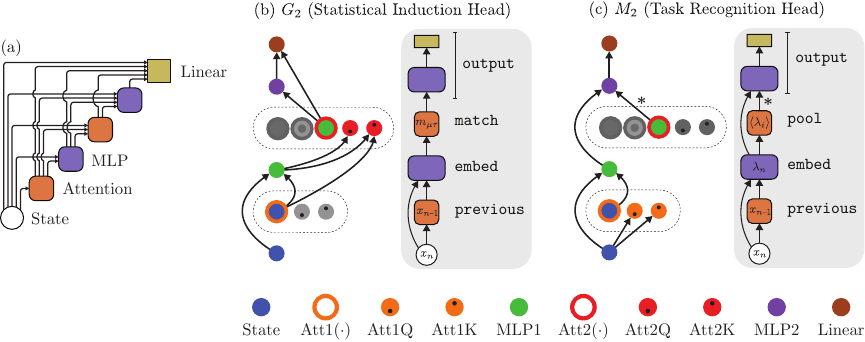}
    \caption{
    Circuit tracing. (a) The transformer may be unrolled into a directed graph with each block taking input from the previous blocks. (b-c) \rev{
    Left: Visualization of the circuit traced for the phase. See Appendix~\ref{sec:tracing} for the definitions of each node in the circuit and an interpretation of the connections.
    Any nodes without incoming and outgoing edges  are depicted as desaturated.
    }
    Right: Interpretative schematic of the circuit traced for the phase, with arrows indicating input-output relationships and layer function labeled. See the main text for more details. Isolated layers are omitted entirely. In panel (c), the asterisk \(\ast\) denotes the task vector discussed in Section \ref{sec:taskvector}.}
    \label{fig:Circuits}
\end{nolinenumbers}
\end{figure*}

We next turn to four mechanistic readouts that probe the attention patterns within the transformer (Appendix~\ref{sec:mech_readouts}). The first is the previous-state parameter $\phi_\delta^{(1)}$, which measures how much the first attention layer attends to the immediately preceding state. Because nearest-neighbor correlations are required for the two 2-point strategies, a rise in $\phi_\delta^{(1)}$ marks the onset of $\pmt$ or $\pgt$. The second is the induction head parameter $\phi_\beta^{(2)}$, which measures whether the second attention layer selectively attends to states that follow previous occurrences of the current state. This is the hallmark of a statistical induction head \cite{olsson2022incontextlearninginductionheads, bietti2023birthtransformermemoryviewpoint, reddy2023mechanistic, edelman2024evolutionstatisticalinductionheads, singh2024needs} (described in further detail below) and specifically delineates $\pgt$. The remaining two readouts, $n_A^{(1)}$ and $n_A^{(2)}$, are entropy-based measures of the effective number of sequence positions attended to by the two attention layers. Large $n_A^{(\ell)}$ indicates diffuse pooling over the sequence, whereas small $n_A^{(\ell)}$ indicates focused retrieval.

These mechanistic observables recapitulate the phase diagram (Figure~\ref{fig:Phases}c). In the 1-point phases, $\phi_\delta^{(1)}$ is small and $n_A^{(1)}$ is large, showing that the first layer broadly pools over the sequence rather than extracting local transitions. Entry into either 2-point phase is marked by a sharp increase in $\phi_\delta^{(1)}$ together with a collapse of $n_A^{(1)}$, indicating that the first layer \rev{extracts previous states}. The split between the 2-point phases is then resolved in the second layer. In $\pgt$, $\phi_\beta^{(2)}$ becomes large and $n_A^{(2)}$ decreases, whereas
in $\pmt$, $\phi_\beta^{(2)}$ remains small while $n_A^{(2)}$ stays large. 

Taken together, Figure~\ref{fig:Phases} shows that, as $t$ and $K$ are varied, the network traverses a small number of discrete algorithmic phases, each with a distinct predictive signature and a distinct internal mechanism. This correspondence provides the empirical basis for the circuit analyses that follow.

\subsection{Tracing circuits}
We next identify how the computation in each phase is implemented (see Appendix~\ref{sec:tracing} for full details). Because every computational block in a transformer writes additively to the residual stream (Figure~\ref{fig:Background}c), the two-layer transformer can be unrolled into a directed graph whose nodes are the state representation, the two attention blocks, the two MLP blocks, and the final linear readout (Figure~\ref{fig:Circuits}a). For the attention blocks, we further separate the query, key and value operations. We then trace \rev{a} circuit by caching the vector transmitted along each edge in an unperturbed forward pass, ablating one edge at a time by replacing its transmitted vector with its batch-mean value, and measuring the resulting change in the model prediction by the KL divergence from the unperturbed output.
\rev{Minor alterations to these traced circuit prototypes are sufficient for fully circuit-constrained models to retain much of their performance.
}
These results together with the readouts in Figure~\ref{fig:Phases}b-c allow us to identify four sparse circuits that emulate each of the four Bayes predictors. The operations implemented by these circuits are presented in Figure~\ref{fig:Circuits}b-c for \(\pgt\) and \(\pmt\), and in Figure~\ref{fig:add_circuits} for \(\pgo\) and \(\pmo\).

In $\pgo$, the first attention layer pools almost uniformly over the sequence ($\phi_\delta^{(1)}$, $n_A^{(1)}$ in Figure~\ref{fig:Phases}c), producing a representation of the empirical 1-point frequencies, and MLP1 maps this representation directly to produce the logits (Figure~\ref{fig:add_circuits}a). The current state plays essentially no role, as expected for an algorithm that predicts based on the stationary distribution of a Markov chain (rather than its transition matrix). In $\pmo$, the same pooled 1-point representation is computed ($\phi_\delta^{(1)}$, $n_A^{(1)}$ in Figure~\ref{fig:Phases}c ) but it is now combined with the current state and decoded through MLP1 and MLP2 to retrieve a memorized transition matrix (Figure~\ref{fig:add_circuits}b). 

In $\pgt$, the first attention layer pays attention to the previous state, so that the residual stream at each position carries information about neighboring pairs ($\phi_\delta^{(1)}$ in Figure~\ref{fig:Phases}c). Pair information is then presented to the second attention layer which implements a matching operation. This two-layer circuit, known as the induction head, has been well-characterized: the representation of the current state is used to match earlier occurrences of that state in the context, and the second attention layer reads out the states that followed those matches. Pooling over these matches yields an estimate of the empirical conditional distribution $\hat{P}(s_{n+1}=\tau\mid s_n=\mu)$, which is then mapped to the output logits. In our traced circuit, the relevant query, key, and value representations run primarily through MLP1 (Figure~\ref{fig:Circuits}c) which is not necessary to implement an induction head, but the essential computation is preserved. A more detailed analysis of $G_2$ is presented further below. Next, we examine the $M_2$ circuit in greater detail, which involves a qualitatively different circuit motif. 

\subsection{The task recognition head and the formation of task vectors}
\label{sec:taskvector}

In the $\pmt$ circuit, the first layer again extracts the previous state (Figure~\ref{fig:Phases}c), but the second attention layer no longer performs content-based matching. Instead, it attends diffusely across the sequence and averages the 2-point embeddings produced by MLP1 (Figure~\ref{fig:Phases}c, Figure~\ref{fig:Circuits}c). We infer that the $M_2$ circuit has an encoder-pool-decoder structure.  Specifically, Att1 and MLP1 encode each neighboring pair $(s_{i-1},s_{i})$ into a pair embedding $\lambda(s_{i-1}, s_{i})$, Att2 pools these pair embeddings into a vector, and MLP2 decodes that vector together with the local state information carried by MLP1 to produce the next-state distribution,
\begin{align}
    \varphi_n &= \frac{1}{n}\sum_{i=1}^{n} \lambda(s_{i-1}, s_{i}),  \quad \hat{\pi}_n = \psi(s_n, \varphi_n), \label{eq:2memcirc}
\end{align}
where $\hat{\pi}_n$ is the predicted probability distribution over the next state, $s_{n+1}$. 

This encoder-pool-decoder picture has an intuitive interpretation. For memorization, the network estimates a compact statistic $\varphi_n$ that distinguishes among the finite set of chains stored in the weights. Averaging over pair embeddings is a natural way to build such a statistic. For this reason, we refer to $\varphi_n$ as a \emph{task vector} and $\pmt$ circuit as a \emph{task recognition head}, where a `task' corresponds to prediction on sequences from one chain.
The embedding $\lambda$ must be nonlinear, as averaging any linear representation of neighboring pairs collapses to a function of 1-point frequencies and does not encode the 2-point information necessary to implement 2-Mem.

\begin{figure}[t!]
\begin{nolinenumbers}
\centering
\includegraphics[width=0.95\linewidth]{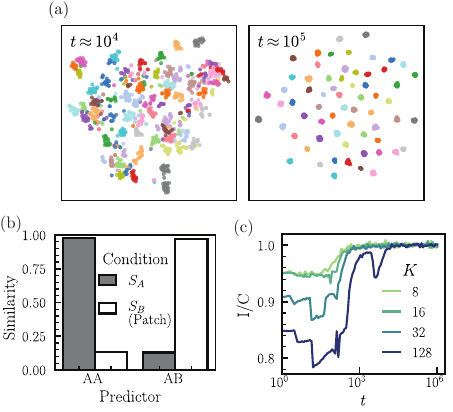}
    \caption{(a) A 2-D t-SNE visualization of task vectors for the $K=64$ model demonstrates increased separability through training. (b) Results of task vector patching experiment. Replacing the task vector formed on a sequence from transition matrix A with that formed on a sequence from transition matrix B results in the transformer predicting according to the transition matrix B. See Appendix \ref{sec:tv_patch} for details. (c) The information content of the task vectors is optimized through training. Mutual information is presented as a fraction of the channel capacity \(\log_2 K\). See Appendix \ref{sec:tv_info} for details.}
    \label{fig:task_vector}
\end{nolinenumbers}
\end{figure}

We perform three specific experiments to further support this interpretation (Appendix~\ref{sec:Memorization}). First, task vectors from the same chain become increasingly separable through training: a nonlinear projection shows diffuse clouds early in training and well-separated task-specific clusters later on (Figure~\ref{fig:task_vector}a). Second, we patch the Att2 output from one sequence into another sequence, i.e., the task vector $\varphi_B$ from a sequence sampled from task $B$ is patched into a sequence sampled from task $A$. The patch causes the model that sees the sequence from $A$ to predict according to sequence $B$'s transition matrix (Figure~\ref{fig:task_vector}b). \rev{These patching results support the interpretation that MLP2 predicts using task-dependent information carried by the statistic $\varphi$ together with current-state information supplied outside the averaging operation, consistent with \eqref{eq:2memcirc}}. Third, the mutual information between $\varphi$ and the latent task identity grows toward the channel capacity $\log_2 K$ during training (Figure~\ref{fig:task_vector}c), indicating that the task vector $\varphi$ is specifically constructed by earlier layers to discriminate tasks. \rev{This is corroborated by a linear decoding analysis which finds that task vectors represent the task posterior rather than directly encode 2-point statistics (Figure~\ref{fig:lin_decoders}).} More so, performing the patch experiment on the \(\pmo\) circuit (Figure~\ref{fig:1mem_patching}) yields similar results to those seen for \(\pmt\), indicating that the task vector construction is used generically during memorization. 

Put together, this analysis characterizes the circuit motifs that underpin each of the four phases. Next, we examine the factors that determine the three boundaries delineating these phases (Figure \ref{fig:Background}\rev{d}): (i) the vertical boundary from $\pgo$ to $\pgt$ for $K > K_1^*$ characterized by the transition time $\tau_{\text{2-Gen}}$, (ii) the `triple point' between $\pgo$, $\pmo$ and $\pgt$ set by $K_1^*$ and $\tau_{\text{2-Gen}}$, and (iii) the boundary between $\pgt$ and $\pmt$ set by $K_2^*$. 

\begin{figure*}[ht!]
\begin{nolinenumbers}
\centering
\includegraphics[width=1.0\linewidth]{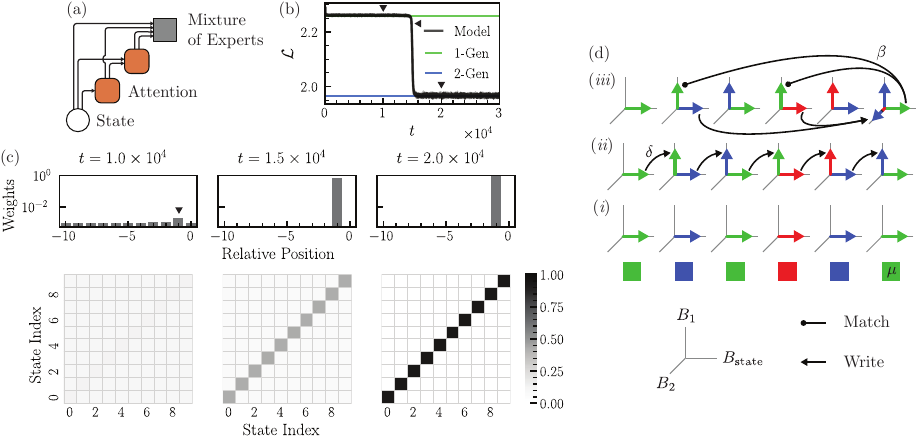}
  \caption{(a) Schematic of the SA-Transformer for generalization.
    (b) Loss dynamics for the SA-Transformer trained in the limit \(K \rightarrow \infty\) display the same sharp transition from \(\pgo\) (green line) to \(\pgt\) (blue line). Data is shown from 5 model seeds, with one darkened for clarity. 
    (c) Model dynamics are dominated by the previous-state attention and the diagonal of one block of the key-query matrix. Top: Normalized weights from the relative positional bias in the first attention layer at different iterations. Bottom: Normalized state-wise attention derived from the bottom-left block of the key/query matrix in the second attention layer at different iterations. (d) A schematic of the induction-head mechanism underlying the model prediction in \(\pgt\) (refer to Appendix~\ref{sec:Induction_head} for a detailed explanation). (i) Given the state embeddings in subspace \(B_\text{state}\), (ii) the first attention block selectively attends to the previous state (with specificity \(\propto \delta\)) and copies that state to an orthogonal subspace \(B_1\), and (iii) the second attention block produces the empirical next-state distribution by comparing the current state with the contents of \(B_1\) (with specificity \(\propto \beta\)) and copying the contents of \(B_2\) from matching positions.}
    \label{fig:Loss}
\end{nolinenumbers}
\end{figure*}

\subsection{The transition from $\pgo$ to $\pgt$}
\label{sec:sa_2gen}
To develop a theory for the transition to $\pgt$, we derive a simpler trainable network that exploits symmetries in $\mathcal{S}$ as $K \to \infty$ (so that $\mathcal{S}$ is equivalent to $\mathcal{D}_T$) and that could implement the statistical induction head isolated in the full transformer  (see Appendix~\ref{sec:MFT_model} for full details). 
After a series of modifications, the content-dependent part of the key-query computation in an attention head, denoted $M^{(\ell)}$, is reduced to a block diagonal form. The key-query computation is then parameterized by one scalar $\beta_1^{(1)}$ in the first attention layer and four scalars $\beta_1^{(2)}, \beta_2^{(2)}, \beta_3^{(2)}, \beta_4^{(2)}$ in the second. The readout contributes three independent mixture weights. The resulting symmetry-constrained, attention-only transformer (\emph{SA-transformer}) therefore has only eight independent parameters, in addition to learned relative positional biases $P^{(\ell)}$ for the two layers. 

Since there are four paths that an input state can take through the computational graph of a two-layer attention-only transformer, this leads to four terms that contribute to the final prediction $\hat{\pi}$,
\begin{align}
\hat{\pi} &= w_A x_N + w_B \sum_{i\le N} A^{(1)}_{iN} x_i +
w_C \sum_{i\le N} A^{(2)}_{iN} x_i \nonumber \\
&+ w_D \sum_{i\le N} \sum_{j\le i} A^{(2)}_{iN} A^{(1)}_{ji} x_j, \quad \text{where}
\label{eq:sa_main}
\\
\log A^{(1)}_{ji} &= \beta^{(1)}_1 \delta_{s_i s_j} + P^{(1)}_{j-i} - \log Z^{(1)},
\nonumber\\
\log A^{(2)}_{ji} &= \beta^{(2)}_1 \delta_{s_i s_j} + \beta^{(2)}_2 \sum_{k\le i} A^{(1)}_{ki}\delta_{s_k s_j} + \beta^{(2)}_3 \sum_{k\le j} A^{(1)}_{kj}\delta_{s_i s_k}
\nonumber\\
&\quad + \beta^{(2)}_4 \sum_{k\le i,\;k'\le j} A^{(1)}_{ki}A^{(1)}_{k'j}\delta_{s_k s_{k'}} + P^{(2)}_{j-i} - \log Z^{(2)},
\nonumber
\end{align}
with $w_A,w_B,w_C,w_D\ge 0$, $w_A+w_B+w_C+w_D=1$, $A_{ji}^{(\ell)}$ is the attention paid by state $i$ to state $j$ at layer $\ell$, and the $Z^{(\ell)}$s are normalization constants. The weighted sum is interpreted as a mixture-of-experts operation, that is, each expert corresponds to a distinct path and its contribution to the final output is weighted by its ability (namely, the four $w$ terms) to accurately predict the next state. 

The SA-transformer is trained non-autoregressively on transition matrices drawn from $\mathcal{D}_T$ and sequences of length $N$. Figure~\ref{fig:Loss}b shows that the reduced model recapitulates the same phenomenology: it first reaches $\pgo$, remains there for an extended time, and then undergoes a sharp drop to $\pgt$. The parameter trajectories in Figure~\ref{fig:Loss}c show that this transition is structured. Specifically, the first-layer content matrix $M^{(1)}$, the second-layer positional bias $P^{(2)}$, and three of the four blocks of $M^{(2)}$ remain at zero (Figure~\ref{fig:gen_parameters}). The dynamics are dominated by two parameters: the previous-state attention in the first layer, $\delta \equiv P^{(1)}_{-1}$, and the parameter that reads the content that was written to by the first attention head, $\beta \equiv \beta^{(2)}_3$ (which correspond precisely to the readouts $\phi_{\delta}^{(1)}$ and $\phi_{\beta}^{(2)}$ in Figure~\ref{fig:Phases}c). The attention heads therefore further reduce to 
\begin{align}\label{eq:approx_main}
\log A^{(1)}_{ji} &= \delta\,\delta_{j,i-1} - \log Z^{(1)}, \qquad \nonumber \\
\log A^{(2)}_{ji} &= \beta \sum_{k\le j} A^{(1)}_{kj}\delta_{s_i s_k} - \log Z^{(2)}.
\end{align}
This form delineates the mechanism of the statistical induction head presented schematically in Figure~\ref{fig:Loss}d and described in detail in Appendix~\ref{sec:Induction_head}. When $\delta\to\infty$, the first layer's operation can be interpreted as a write operation that stores a state $s_i$'s information in a `buffer' subspace of the next state $s_{i+1}$. When $\beta\to\infty$, the second attention head attends only to positions whose buffered state matches the most recent state $\mu=s_N$. This head then retrieves an average of the states that immediately followed previous occurrences of $\mu$ in the sequence. With $w_C=1$, the readout reduces to the empirical 2-point estimator, i.e., the 2-Gen predictor. 

\begin{figure*}[ht!]
\begin{nolinenumbers}
\centering
\includegraphics[width=0.95\linewidth]{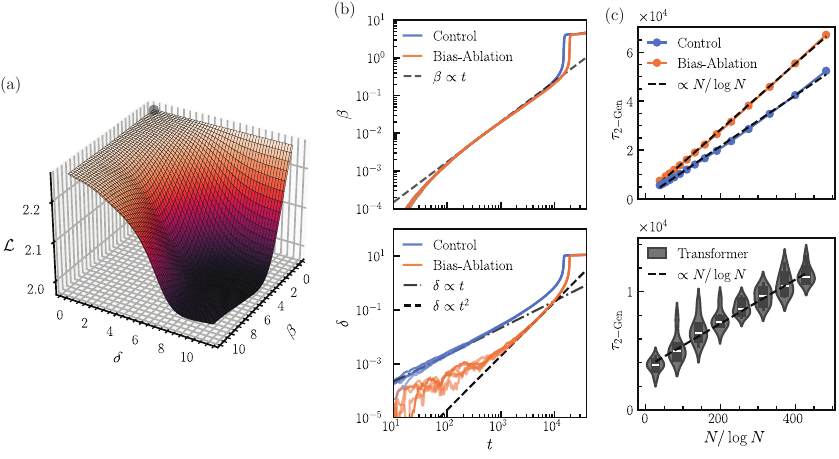}
     \caption{(a) Loss landscape in the $(\beta, \delta)$ parameter space given by equation \ref{eq:sa_main} and \ref{eq:approx_main} (see equation \ref{eq:minimal_simple} in Appendix) where $w_{A-D}$ are obtained by optimization over the full landscape. (b) Top: Scaling of the effective parameter $\beta$ with training iterations $t$. 
     Bottom: Scaling of the effective parameter $\delta$ with training iterations $t$. 
     Different lines in each plot with the same color represent numerics with different seeds. In the ``Control'' setting, the direct-\(\delta\) bias predicts a linear growth of both \(\delta,\beta\). In the ``Bias-Ablation'' setting, the second-to-last sequence state has been resampled to remove the direct-\(\delta\) bias which results in \(\delta\sim t^2\) instead.
    (c) Scaling of the transition time $\tau_{2\text{-Gen}}$ with sequence length \(N\). The time \(\tau_{2\text{-Gen}}\) is measured as the number of iterations of SGD required until the sharp drop in loss indicating the development of the induction head. Top: Measurements for the SA-transformer (described in eq. \ref{eq:base_model}) in the two settings considered in panel (b). Bottom: Measurements for a non-autoregressive 2-layer transformer with MLP1 removed and additive relative positional biases. Optimization uses SGD with momentum. Kernel density estimated from 8 seeds.}
    \label{fig:Landscape}
\end{nolinenumbers}
\end{figure*}

With the SA-transformer further simplified, we turn to the dynamics of the 1-Gen solution. At initialization, both attention layers are uniform over the context, so the experts corresponding to $w_B$, $w_C$, and $w_D$ all reduce to the same 1-point estimator. The only term that does something different is $w_A x_N$, which predicts by repeating the final state. This repeat bias is disadvantageous on average for the Dirichlet ensemble, and we show that gradient descent rapidly drives $w_A$ to zero (Appendix~\ref{sec:x_N}). The $\pgo$ plateau is therefore well-approximated by
$
w_A=0,\;
w_B=w_C=w_D=1/3,\;
\beta=\delta=0.
$
Expanding the loss around this point to first order, we have
\begin{align}
\mathcal{L}(\beta,\delta) = \mathcal{L}^{\text{1-Gen}} - c_\beta \beta - c_\delta \delta \cdots .
\label{eq:sa_loss_expand}
\end{align}
A naive self-averaging argument that replaces sums with expectations shows that the first nontrivial term is the second-order mixed contribution $\beta\delta$, so that the origin is a saddle point. This argument, however, does not explain why the network flows along $\beta >0, \delta >0$ rather than $\beta  < 0, \delta  < 0$. A careful calculation of the first-order terms shows that there are two important biases that tilt the landscape towards $\beta >0, \delta >0$. These biases imply that the network consistently converges to $\pgt$, and that these terms determine how long it takes for the network to transition to $\pgt$, i.e., $\tau_{\text{2-Gen}}$. We summarize key results below, and present the full details in Appendix~\ref{sec:SA_dynamics}.

The coefficient $c_\delta$ has contributions from the $w_B$ and $w_D$ experts. Intuitively, these contributions arise because, given a sequence $s_1,s_2,\dots, s_n$, it is slightly more likely for the next state $s_{n+1}$ to return to $s_{n-1}$ when averaged over the ensemble of stationary transition matrices. This leads the network to develop a slight bias towards paying attention to the previous state ($\delta > 0$). We calculate this bias by expanding $A^{(1)}$ to first order, which gives terms proportional to
$P(s_{n-1}=\tau \mid s_N=\mu)-p_\tau$, where $p$ is the stationary distribution of a particular chain. For most values of $n$, this difference is negligible because the chain has mixed, but it cannot be neglected when $n \approx N$. Averaging first over sequences conditioned on the final state and then over the transition matrix ensemble gives
\begin{align}
c_\delta &\approx \frac{w_B F_1}{N} + \frac{w_D}{N}\sum_{i\le N}\frac{F_{N-i+1}}{i}, \text{  where} \nonumber \\
F_d &\equiv \left\langle \sum_\mu (T^{d+1})_{\mu\mu} \right\rangle_{\mathcal{D}_T} -1.
\label{eq:sa_cdelta}
\end{align}
For a symmetric Dirichlet ensemble with parameter $\alpha$, $F_1 = (C-1)/(C^2\alpha + C)$, so for the parameters used here ($C=10$, $\alpha=1$) one has $F_1 \approx 0.08$. Numerically, $F_d$ decays rapidly with $d$ (see Figure \ref{fig:markov_bias}), so the second term in \eqref{eq:sa_cdelta} is sub-leading and $c_\delta \simeq w_BF_1/N$. 

The bias in coefficient $c_\beta$ comes mainly from the $w_C$ expert. Here, the bias can be understood by considering a state (say, $s$) that consistently follows the current state. The current state is slightly over-represented in the output of the first attention layer at previous occurrences of $s$ (recall that the first attention layer pools when $\delta = 0$). This weak correlation between the current state and the over-representation of the current state at occurrences of $s$ is sufficient to increase $\beta$. Each position $i$ contributes only $\sim 1/i$, but the sum over the whole sequence produces a harmonic sum $H_N = \sum_{i=1}^N 1/i \sim \log N$. The resulting coefficient is
\begin{align}
c_\beta &\approx w_C\frac{H_N}{N} I, \text{  where} \quad
I \equiv \left\langle \sum_\mu p_\mu^2 \left( \sum_\tau \frac{T_{\tau\mu}^2}{p_\tau} - 1 \right)\right\rangle.
\label{eq:sa_cbeta}
\end{align}
It is possible to show (using the Cauchy-Schwarz inequality) that $I>0$ for every transition matrix, so this term contributes a second systematic bias ($\beta > 0$) toward induction head formation. Together,~\eqref{eq:sa_cdelta} and \eqref{eq:sa_cbeta} show that the transition is not governed by a rare fluctuation that allows it to escape $\pgo$, but rather by weak statistical biases that are fortuitously oriented towards the induction head solution.  Notably, the $w_B$ expert does not contribute to the final 2-Gen solution, yet it provides the leading contribution to $\delta$. Keeping only the leading terms near the plateau gives,
\begin{align}
\frac{d\beta}{dt}
&\approx
\frac{I H_N}{3N} ,
\qquad
\frac{d\delta}{dt} \approx \frac{F_1}{3N} ,
\label{eq:sa_dynamics}
\end{align}
with $w_B=w_C=1/3$ to lowest order. Figure~\ref{fig:Landscape}a shows the corresponding reduced loss landscape after optimizing $w_A$-$w_D$ at fixed $(\beta,\delta)$. The origin is a shallow 1-Gen basin. The two linear biases tilt the surface towards $\beta > 0, \delta > 0$, and once $\beta$ and $\delta$ are order one, the model encounters a sharp cliff (due to the nonlinear softmax operation in an attention head) and rapidly converges to the 2-Gen basin. This picture explains why the drop in Figure~\ref{fig:Loss}b appears discontinuous. 

\begin{figure*}[ht!]
\begin{nolinenumbers}
\centering
\includegraphics[width=0.9\linewidth]{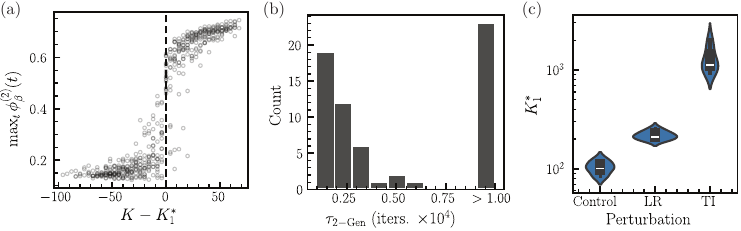}
    \caption{(a) The maximum value of the induction head parameter \(\phi_\beta^{(2)}\) through training is sigmoidal as the data diversity crosses over \(K_1^*\). This indicates that the model either reaches the threshold and proceeds to \(\pgt\) (high \(\phi_\beta^{(2)}\)) or does not reach the threshold and proceeds to \(\pmo\), with few intermediates. Evaluated over 16 different model seeds trained on the same task distribution. Each seed has been shifted along the horizontal axis such that \(\phi_{\beta}^{(2)}\) first crosses the threshold value \(\phi_{\beta}^{*}\) at \(K = K_1^*\). See Appendix \ref{sec:mem_to_gen} for details. (b) Bimodality of the model dynamics near \(K_1^*\), shown for \(K = 94 \approx K_1^\ast\) and 64 seeds trained on the same \(\mathcal{D}_T\). Different model seeds show either small \(\tau_\text{2-Gen}\) or never achieve \(\pgt\). All model seeds in the rightmost bin did not enter \(\pgt\) after \(5 \times 10^4\) iterations. (c) Perturbations to the learning dynamics of the competing sub-circuits have effects on the memorization-generalization transition that are explainable by a kinetic competition. $K_1^*$ increases when the effective learning rate of the 2-Gen circuit \(\gamma_\text{2-Gen}\) is reduced (LR perturbation). Increasing the rate of memorization by providing information about the ground truth task identity also increases $K_1^*$ (TI perturbation). Kernel density estimated from 4 seeds.}
    \label{fig:K1}
\end{nolinenumbers}
\end{figure*}

The theory makes two further predictions that can be tested numerically. First, if the direct $\delta$-bias is removed while leaving the long-range statistics intact, then $\beta$ should still grow linearly but $\delta$ should grow only through its coupling to $\beta$. We test this by resampling the second-to-last state after sequence generation, which destroys the short-range correlation responsible for $F_1$ while preserving the longer-range structure of the task.  We refer to this perturbed setting as the “bias-ablation” case, and to the original setting as the control case. The result of this perturbation matches  predictions, as shown in Figure~\ref{fig:Landscape}b,c: $\beta \sim t$ in both the original and perturbed settings, whereas $\delta \sim t$ in the original task but $\delta \sim t^2$ in the bias-ablation setting. Second, because $\beta$ is the faster variable and reaches the nonlinear regime at a rate set by $H_N/N$, the transition time for a non-autoregressive transformer obeys $\tau_{2\text{-Gen}} \sim {N}/{H_N} \sim {N}/{\log N}$.
Figures~\ref{fig:Landscape}c verify this scaling both in the SA-transformer and for the transformer architecture (albeit, without MLP1) trained non-autoregressively. For autoregressive training, one should average  gradients over all possible context lengths $n$ uniformly weighted until $N$, which leads to 
\begin{align}
\tau_{2\text{-Gen}} \sim \left(\frac{1}{N} \sum_{n = 1}^N \frac{H_n}{n} \right)^{-1} \approx 2N/(\log N)^2,
\end{align}
for $N \gg 1$ ($N=1024$ in our setting). 

\rev{
To determine whether these theoretical predictions extend beyond the SA-transformer, we also evaluated a standard architecture with learnable embeddings, query, key, and value matrices, and a MLP readout (detailed in Appendix~\ref{sec:beyond_sa}).  While the derived scaling is valid for autoregressive training with SGD and RPB, more sophisticated optimizer and positional encoding schemes introduce elements that are not captured by the theory.}

\subsection{$K_1^*$ is determined by a kinetic competition}
\label{sec:k1_kinetics}

The first transition from memorization to generalization appears at $K = K_1^*$ (Figure~\ref{fig:Background}\rev{d} and Figure~\ref{fig:Phases}). When $K < K_1^*$, the network transitions from $\pgo$ to $\pmo$ whereas for  $K > K_1^*$, the network transitions from $\pgo$ to $\pgt$. The induction head parameter $\phi_\beta^{(2)}$ is nonzero only in $\pgt$ and thus we measure $K_1^*$ based on whether $\phi_\beta^{(2)}$ crosses a threshold (0.45) during training. This measurement leads to an estimate of $K_1^* \approx 94$. Figure~\ref{fig:K1}a shows that the transition is sharp and effectively bimodal: models either cross the threshold and develop an induction head or transition into $\pmo$, with few intermediates. Bimodality close to the transition is recapitulated in the distribution of $\tau_{\text{2-Gen}}$ close to $K_1^*$ across seeds  (Figure~\ref{fig:K1}b).

The transition at $K_1^*$ can be intuitively understood from a mixture-of-experts viewpoint. Suppose a network contains distinct subcircuits indexed by $a$ and $b$ (for e.g., 2-Gen and 1-Mem) and the next-state prediction is 
\begin{align}
    \hat{\pi} \approx  \vartheta \hat{\pi}_a + (1-  \vartheta)\hat{\pi}_b,
\end{align}
where $0 \le \vartheta \le 1$ gates the importance given to each subcircuit. This gating parameter is set by the relative reliability of each subcircuit in predicting the next state: the gradient update in $\vartheta$ due to a given sequence that incurs a loss $\mathcal{L} = -\log \hat{\pi}$ is $d\vartheta/dt \propto -\partial \mathcal{L}/\partial \vartheta \propto (\hat{\pi}_a - \hat{\pi}_b)$.
That is, $\vartheta$ grows if $\hat{\pi}_a$ is a better predictor than $\hat{\pi}_b$. Further, if distinct sets of parameters contribute to $\hat{\pi}_a$ and $\hat{\pi}_b$, then their respective gradients have prefactors $\vartheta$ and $(1-\vartheta)$. In short, the subcircuit that already assigns a larger probability to the correct next state is upweighted by the gate, while the gradient flowing into the losing subcircuit is suppressed. If the gate is learned on a faster timescale than the subcircuits, a small early performance difference in the two subcircuits favors the leading subcircuit regardless of which subcircuit is a better predictor after convergence. This kinetic competition between subcircuits can lead to a sharp transition (Figure~\ref{fig:K1}a) if one of the subcircuits (here, 2-Gen) forms abruptly. 

We directly test this kinetic picture with two perturbations that shift the two timescales in opposite directions\rev{, an effect which we verify in Figure~\ref{fig:kinetic_demo}} (further details in Appendix~\ref{sec:mem_to_gen}). In the first perturbation, we slow the learning dynamics of only the $2$-Gen circuit using gradient reweighting. This method reweights the gradient on previous-state attention in the first layer while leaving the forward pass unchanged. The perturbation shifts $K_1^*$ upward approximately by a factor of two (Figure~\ref{fig:K1}c). Importantly, this perturbation has no impact on the model architecture, thus highlighting that the shift in $K_1^*$ is due to a shift in training dynamics. In the second experiment, we apply a complementary intervention that accelerates memorization by injecting the ground-truth task identity into \rev{MLP2}. This perturbation shifts $K_1^*$ upward by a factor of ten (Figure~\ref{fig:K1}c). Gradient reweighting and task injection therefore move the boundary in the same direction by increasing $\tau_{\text{2-Gen}}$ and decreasing $\tau_{\text{1-Mem}}$, respectively.

\begin{figure*}[ht!]
\begin{nolinenumbers}
\centering
\includegraphics[width=0.9\linewidth]{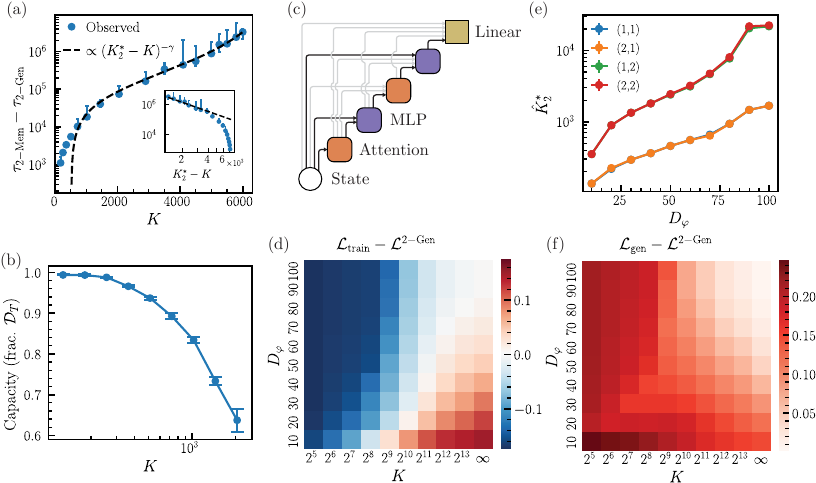}
    \caption{
    (a) The time between the onset of \(\pgt\) and its disappearance in favor of \(\pmt\) diverges as \((K_2^\ast - K)^{-\gamma}\) with \(\gamma \approx 2\) and \(K_2^\ast \approx 7000\). The model is assumed to enter \(\pmt\) when it achieves a training loss \(\mathcal{L}_\text{train} < \mathcal{L}^{\text{2-Gen}}\). Median value over 4 different seeds trained on the same task distribution \(\mathcal{D}_T\) shown, with errorbars indicating the 95\% confidence interval. Dashed line indicates a divergence fit to the medians with a linear offset. Inset: Replot of the data highlighting asymptotic behavior. The fit value for \(K_2^\ast\) was used to compute the offset \(K_2^\ast - K\). Dashed line with slope \(-\gamma\) provided for reference. (b) The mean fraction of memorized tasks extrapolated to convergence decreases linearly past \(K \approx 2^8\). See Appendix \ref{sec:partial_mem} for details. A linear extrapolation of these data suggests that the model will not memorize any chain for \(K \approx 5000\), which is consistent with the estimate for \(K_2^\ast\) from the data in panel (a). Errorbars show standard deviation over the 4 seeds. (c) Schematic of the  minimal network inspired from the circuit-tracing results in Figure~\ref{fig:Circuits}, used to reproduce the 2-Mem behavior shown in the remaining panels. (d,\rev{f}) Heat maps of $\mathcal{L}_{\mathrm{train}}-\mathcal{L}^{\mathrm{2\text{-}Gen}}$ and 
$\mathcal{L}_{\mathrm{gen}}-\mathcal{L}^{\mathrm{2\text{-}Gen}}$ as functions of the task-vector embedding dimension $D_\varphi$ and data diversity $K$. 
Blue (red) indicates that the model is over (under) performing relative to the 2-Gen predictor; white denotes the crossover.
(\rev{e}) Average critical data diversity $\hat{K}_2^*$ over three independent trials. The labels $(\ell_1,\ell_2)$ indicate the number of hidden layers in MLP1 and MLP2, respectively. The standard deviation error bar is estimated with three independent trials.} \label{fig:mem}
\end{nolinenumbers}
\end{figure*}

\subsection{$K_2^*$ and the representational capacity of the task recognition head}
\label{sec:minimal_2mem}

\rev{
We anticipate a threshold $K_2^*$ above which $\pmt$ is never learned (Figure~\ref{fig:Background}d). However, this is challenging to detect as the network's loss continues to gradually decline during training and may eventually transition into $\pmt$ if trained long enough. To establish an estimate for $K_2^*$, we measure the time $\Delta \tau_K$ from the onset of $\pgt$ until the transformers achieve a training loss $\mathcal{L}_{\text{train}} < \mathcal{L}^{\text{2-Gen}}$ (which is not possible without memorization) for different values of $K$. $\Delta \tau_K$ diverges as $(K_2^* - K)^{-\gamma}$, with an exponent $\gamma \approx 2$ and yields an estimate $K_2^* \approx 7000$ (Figure~\ref{fig:mem}a). This estimate is consistent with an alternative measurement obtained from the fraction of memorized chains versus $K$ (Figure~\ref{fig:mem}b).
}


Since implementing 2-Mem requires the network to encode the transition matrices in its weights, we consider the hypothesis that the primary factor that determines $K_2^*$ is tied to a representational bottleneck in one of the elements of the $\pmt$ circuit, namely, the task recognition head. Recall that this circuit includes three core elements (equation~\ref{eq:2memcirc}): MLP1 in the encoder ($\lambda$), the task vector ($\varphi$) and MLP2 in the decoder ($\psi$). To systematically quantify how $K_2^*$ changes with variations in these elements, we turn to a minimal autoregressive model of the task recognition head that isolates its essential features. As shown in Figure~\ref{fig:mem}c and derived in Appendix~\ref{app:minimal_network_mem}, we replace the first attention head by providing MLP1 direct access to neighboring pairs and the second attention head by averaging over the pair embeddings written by MLP1 (Figure~\ref{fig:mem}c). The resulting network retains only the operations that are essential for $\pmt$. Concretely,  the minimal model performs
\begin{equation}
\varphi_n = \Big\langle \mathrm{MLP1} \left(x_{i+1} \oplus x_i \right) \Big\rangle_{i<n}, 
\quad z_n = \mathrm{MLP2}\!\left(x_n \oplus \varphi_n\right),
\label{eq:minimal_task_head}
\end{equation}
where $x_i$ is the one-hot representation of $s_i$ and the logits $z_n$ determine the predicted probability of state $s_{n+1}$. The only control parameter relative to the full transformer is the task vector dimension $D_{\varphi}$. Note that when $D_{\varphi} \gtrsim C^2$, the network has sufficient expressivity to encode the full transition matrix in the task vector \rev{(whether training makes use of this capacity depends on $K$, Section~\ref{sec:tv_2gen}).}

For a reference architecture (one hidden layer in MLP1 and two hidden layers in MLP2),  we reproduce the defining signature of $\pmt$ (Figures~\ref{fig:mem}d, f). When $K$ is small, the model achieves a training loss $\mathcal{L}_{\mathrm{train}}<\mathcal{L}^{\mathrm{2\text{-}Gen}}$ while displaying suboptimal generalization $\mathcal{L}_{\mathrm{gen}}>\mathcal{L}^{\mathrm{2\text{-}Gen}}$. As $K$ increases, there is a crossover and 2-Gen is optimal. Because the minimal architecture is not expressive enough to form an induction head, the loss of this advantage isolates the point at which the task vector mechanism can no longer implement a 2-Mem circuit that outperforms 2-Gen. We therefore interpret this crossover as the analogue of the $K_2^\ast$ boundary in the full transformer\rev{, and denote the corresponding threshold by $\hat{K}_2^\ast$}.

We obtain an estimate $\hat{K}_2^\ast$ in the minimal model via binary search while varying $D_{\varphi}$ and the depths of MLP1 and MLP2. Figure~\ref{fig:mem}e shows that $\hat{K}_2^*$ is only weakly sensitive to the encoder. That is, for fixed $D_{\varphi}$ and MLP2 depth, increasing MLP1 from one to two layers leaves $\hat{K}_2^*$ nearly unchanged. The primary determinants are $D_{\varphi}$, such that $\hat{K}_2^*$ scales approximately exponentially with $D_{\varphi}$ when $D_{\varphi} < C^2$, and MLP2, such that removing a hidden layer from this block decreases $\hat{K}_2^*$ by a factor of ten. Thus, encoding a pair embedding is relatively simple, but representing and retrieving the corresponding collection of transition matrices in $\mathcal{S}$ is constrained by the finite-dimensional residual stream (set by the model dimension) and the expressivity of the decoder (set by MLP2). \rev{Specifically, the bottleneck at $K_2^\ast$ primarily reflects a constraint on the decoder's ability to map task vectors to transition probabilities, rather than an information bottleneck in the encoder's capacity to isolate tasks and encode bigram distributions.}

\subsection{Task vectors as an alternative mechanism for in-context generalization}
\label{sec:tv_2gen}


\begin{revblock}
Importantly, the task recognition head is not intrinsically a memorizing mechanism. As $K \to \infty$, both $\mathcal{L}_{\mathrm{train}}$ and $\mathcal{L}_{\mathrm{gen}}$ approach $\mathcal{L}^{\mathrm{2\text{-}Gen}}$ provided $D_{\varphi}\gtrsim C^2$, which means MLP2 reads transition statistics from $\varphi$ and approximates the 2-Gen predictor without an induction head. In contrast, at small $K$, the model achieves $\mathcal{L}_{\mathrm{train}}<\mathcal{L}^{\mathrm{2\text{-}Gen}}$ but generalizes poorly (Figure~\ref{fig:mem}d,f). This is true even when the task vector could in principle hold all $C^2$ transition probabilities (i.e., $D_{\varphi}\approx C^2$), indicating the network favors distinguishing among stored chains and retrieving their transition matrices.

These observations also demonstrate that memorization and generalization by the task recognition head are governed by both representational capacity ($D_{\varphi}/C^2$) and task diversity ($K$). These dependencies are recapitulated in the transformer, where ablating the $\pgt$ circuit by restricting the second attention layer to uniform attention yields the usual four phase diagram when $D \approx C^2$ (Figure~\ref{fig:pool_phases}).

Finally, the ability of the task recognition head to generalize suggests two possible encoding schemes for the task vector, namely directly encoding the transition statistics or using a statistics-independent code such as representing the task posterior. We distinguish between these in the $\pmt$ transformer by training linear decoders to identify each quantity in the task vectors (Figure~\ref{fig:lin_decoders}), the results of which suggest that the $\pmt$ task vectors encode the task posterior rather than the transition statistics.
\end{revblock}

\section{Discussion}
Through numerical experiments and phenomenology-driven theory, we characterize four distinct multi-layer circuits that implement distinct in-context learning algorithms in transformers. These circuits show that sequence context is used by the model in two qualitatively different ways: to estimate $n$-point statistics solely from the sequence, or to identify and retrieve a latent representation of a data generating process seen during training (both motifs shown schematically in Figure~\ref{fig:Circuits}b,c). While the former allows the model to generalize outside the training set, the latter requires fewer examples in the context to achieve equivalent predictive accuracy if the data is sampled from a known process. 

\begin{figure}[t!]
\begin{nolinenumbers}
\centering
\includegraphics[]{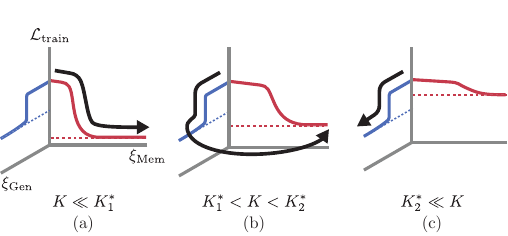}
\caption{A schematic summarizing the factors that govern memorization-generalization transitions. Suppose the transformer's training dynamics proceeds along a generalized coordinate for memorization (\(\xi_\text{Mem}\)) and  generalization (\(\xi_\text{Gen}\)). (a) When the data diversity $K$ is small relative to \(K_1^*\), memorization is both faster than generalization and reaches a lower loss, so the model proceeds along \(\xi_\text{Mem}\) alone. (b) For $K$ between \(K_1^*\) and \(K_2^*\), generalization develops before memorization and the model begins progressing along \(\xi_\text{Gen}\). However, the model eventually switches over to memorization, which can achieve a lower loss. (c) When $K$ is well above \(K_2^*\), constraints on model expressivity render memorization unfavorable relative to generalization, and dynamics are constrained to \(\xi_\text{Gen}\) as a result.}
\label{fig:discussion}
\end{nolinenumbers}
\end{figure}

In the generalizing regime, 2-point statistics are extracted by a statistical induction head~\cite{edelman2024evolutionstatisticalinductionheads}, a generalized version of an induction head first identified in large language models~\cite{olsson2022incontextlearninginductionheads}. This circuit primarily relies on a multi-layer interaction between two attention heads. That is, the first attention layer Att1 writes local pair information into the residual stream, and the second attention layer Att2 performs a matching operation that estimates the relevant transition statistics~\cite{edelman2024evolutionstatisticalinductionheads,reddy2023mechanistic,bietti2023birthtransformermemoryviewpoint, olsson2022incontextlearninginductionheads}. In the memorizing regime, we show that the model instead uses a distinct encoder-pool-decoder circuit, which we term the task recognition head. Here, the feedforward block MLP1 in the first layer constructs nonlinear embeddings of neighboring state pairs created by the first attention layer Att1, the second attention layer Att2 pools these pair embeddings across the sequence, and the second feedforward block MLP2 decodes the resulting sequence-level representation together with local state information. This pooled representation is naturally interpreted as a task vector, a compact latent representation of the process that generated the contextual sequence \rev{which} can subsequently gate the operation performed on new inputs.

Two plausible mechanisms for in-context learning in large language models  have been proposed: induction heads~\cite{olsson2022incontextlearninginductionheads}, and retrieval-oriented latent task representations~\cite{hendel2023context,todd2023function}, though a circuit that implements the latter has not yet been found. Both these mechanisms are reproduced in our setting, and our proposed minimal models isolate the key features of both motifs. Importantly, we show that the task recognition head can also be a mechanism for generalization, though the constraints are much stronger than those required for a statistical induction head. Specifically, a task vector can support perfect generalization when a network with a sufficiently large residual stream and expressive feedforward block (here, MLP2) is trained on sufficiently diverse data (as quantified in Figure~\ref{fig:mem}f). \rev{The task recognition head thus represents an expressive and seemingly general-purpose predictive mechanism, requiring only an encoder-pool-decoder structure, which could be implemented in a variety of substrates including chemical reaction networks~\cite{floyd2026context} and RNNs~\cite{zheng2024rapid}.}

Our analysis highlights the importance of the feedforward block, which is absent in previous analyses of in-context learning. Here, MLP1 is required to construct a nonlinear embedding of state pairs and MLP2 serves as a decoder that maps task vectors to the next-state distribution. In short, both attention and MLP blocks play distinct and algorithmically essential roles. More generally, this picture suggests a functional separation between layers in larger transformer models: early layers first compile evidence from subsequences in the context, these pieces of evidence are then pooled together to form a vector representation of a latent variable (task vector), and later layers act as decoders that efficiently process the new input in a context-dependent manner. This interpretation is consistent with empirical work in large language models which suggest the same general pattern of specialization~\cite{press2020improvingtransformermodelsreordering, lad2025remarkablerobustnessllmsstages, dong2025randomattentionsufficientsequence}.

The effective model for the 2-Gen circuit, based on our SA-transformer architecture, reveals the importance of statistical biases in the early formation of statistical induction heads. Although these biases do not reflect the 2-point structure of the data and weakly align with the final predictor in parameter space, they provide fortuitously oriented drift terms that enable the system to reliably escape flat regions of the loss landscape. This perspective could explain why pre-training with even nonlinguistic synthetic data can accelerate the performance of large language models~\cite{lee2026training}. Our data are limited to 2-point correlations; in the limit $K \rightarrow \infty$, we expect ladder-like generalization transitions to emerge as higher-order correlations and deeper transformer architectures are incorporated~\cite{rende2024distributional}. For instance, transformers also pick up on 3-point correlations~\cite{edelman2024evolutionstatisticalinductionheads}. It is straightforward to extend the SA-transformer architecture to more than two layers and multiple heads per layer, offering an opportunity for theory in settings that involve more complex statistical structure. 

This work reconciles competing views of memorization-generalization transitions~\cite{nguyen2024differential,lu2024asymptotic,park2024competition}. The first threshold, $K_1^\ast$, is kinetic: as data diversity increases, memorization slows, whereas the 2-Gen circuit forms on a largely $K$-independent timescale. The resulting kinetic competition between circuits determines whether the network transitions from $\pgo$ to $\pmo$ or from $\pgo$ to $\pgt$. The second threshold, $K_2^\ast$, is likely due to an expressivity constraint: even once the model can infer task identity from context, the finite capacity of the residual stream to represent task vectors and the limited decoding capacity of MLP2 limit how many distinct tasks can be encoded and reliably retrieved. The intuition behind these two mechanisms is presented schematically in Figure~\ref{fig:discussion}. This two-threshold picture explains why memorization and generalization can look like a competition of timescales in some settings, but like a capacity constraint in others~\cite{nguyen2024differential,lu2024asymptotic}.  More broadly, this picture suggests a common language for related phenomena in diffusion models, where recent work likewise points to an early generalization time, a later dataset-size-dependent memorization time, and a separate model-dependent threshold beyond which long-time overfitting disappears~\cite{bonnaire2025diffusion,favero2025bigger}.


\rev{Our results are not limited to transformers and contribute} to the emerging viewpoint that complex networks, once trained on large and diverse datasets, acquire remarkably rich mechanisms to represent and infer structure from a small number of examples. Understanding how and when such systems implement such computations should sharpen hypotheses about context-dependent learning in biological systems \cite{heald2023computational, zheng2024rapid} and offer design principles for physical learning systems in which direct parameter tuning is often difficult or impossible \cite{floyd2026context}. We therefore view the present results as fundamental mechanisms for how rapid learning can arise from a small set of reusable computational motifs.

\section{Code Availability}
The code used to produce the results of this article is openly available at \cite{code}.

\begin{acknowledgments}
We thank Colin Scheibner and members of the Reddy lab for insightful comments. GR was partially supported by a joint research agreement between NTT Research Inc. and Princeton University, and by a seed grant from the Princeton AI Lab. WC was supported by a grant from Coefficient Giving.
This material is based upon work supported by the National Science Foundation Graduate
Research Fellowship Program under Grant No. DGE-2444107. Any opinions,
findings, and conclusions or recommendations expressed in this material are those of the
author(s) and do not necessarily reflect the views of the National Science Foundation.
A portion of the simulations presented in this article were performed on computational resources managed and supported by Princeton Research Computing, a consortium of groups including the Princeton Institute for Computational Science and Engineering (PICSciE) and Research Computing at Princeton University.
\end{acknowledgments}

\bibliography{refs.bib}

\clearpage
\renewcommand{\thefigure}{S\arabic{figure}}
\setcounter{figure}{0} 
\onecolumngrid
\begin{center}
\textbf{\large Supplementary Information}
\end{center}
\appendix
\renewcommand\appendixname{}
\renewcommand{\thesection}{\Roman{section}} 
\renewcommand\appendixname{}

\renewcommand{\thefigure}{A\arabic{figure}}
\setcounter{figure}{0}

\tableofcontents

\section{Summary of  notation}\label{sec:summaryvariables}
We provide a brief description of each quantity and its value (unless otherwise specified). \rev{The subscripted symbols $\phi_\delta^{(1)}$ and $\phi_\beta^{(2)}$ denote first- and second-layer attention readouts, respectively; $\varphi$ denotes the task vector.}

\subsection{Data generating process}
\makebox[1.5cm]{$C$}        number of states (set to 10) \par


\makebox[1.5cm]{$T$}   $C \times C$ column-stochastic transition matrix, element \(T_{\tau\mu}\) is transition probability $\mu \rightarrow \tau$ \par

\makebox[1.5cm]{$\mathcal{D}_T$}    distribution over \(T\) imposed by drawing each column i.i.d. from \text{Dir}($\alpha, C$) $\alpha$ set to $1$   \par

\makebox[1.5cm]{$\mathcal{S}$}  set of training transition matrices (or `tasks') $\{T^{(1)}\dots T^{(K)}\}$ drawn from $\mathcal{D}_T$  \par

\makebox[1.5cm]{$K$}  data diversity $|\mathcal{S}|$  \par

\makebox[1.5cm]{$N$}        maximum number of states in a sequence (typical value $2^8$ for autoregressive, $2^{10}$ for fixed-length training) \par

\makebox[1.5cm]{$p_{\mu}$}   $C$-dimensional stationary distribution, which satisfies $\sum_{\mu} p_{\mu}T_{\tau \mu} = p_{\tau}$ for corresponding transition matrix $T$. \par

\makebox[1.5cm]{$s_i$}  state at position $i$ \par

\makebox[1.5cm]{$S_n$}  length \(n\) ordered sequence of states $\{s_i\}_{1:n}$ \par


\subsection{Reference transformer architecture}

\makebox[1.5cm]{$x_\tau$}  embedding vector for the state $\tau$ (interchangeably one-hot or learned embedding) \par

\makebox[1.5cm]{$D$} state embedding dimension (typical value \(64\)) \par

\makebox[1.5cm]{$\hat{\pi}^\theta$}     predictive distribution output by a network \par

\makebox[1.5cm]{$A_{ni}^{(\ell)}$}  attention weight given by sequence position \(n\) to position \(i\) \par

\makebox[1.5cm]{$\mathcal{L}_\text{train}$}  training loss evaluated on sequences generated from chains in \(\mathcal{S}\) \par

\makebox[1.5cm]{$\mathcal{L}_\text{gen}$}  generalization loss evaluated on sequences generated from chains in \(\mathcal{D}_T\) \par

\subsection{Four algorithmic phases}

\makebox[1.5cm]{$\hat{\pi}^{\text{1-Gen}}$}
Bayes optimal predictor which uses one-point statistics to infer the chain from $\mathcal{D}_T$ that generated a sequence and predict the next state  \par

\makebox[1.5cm]{$\pgo$} algorithmic phase in which the transformer implements \(1\)-Gen \par 

\makebox[1.5cm]{$\hat{\pi}^{\text{2-Gen}}$}
Bayes optimal predictor which uses two-point statistics to infer the chain from $\mathcal{D}_T$ that generated a sequence and predict the next state \par 

\makebox[1.5cm]{$\pgt$} algorithmic phase in which the transformer implements \(2\)-Gen \par 

\makebox[1.5cm]{$\hat{\pi}^{\text{1-Mem}}$}
Bayes optimal predictor which uses one-point statistics to infer the chain from $\mathcal{S}$ that generated a sequence and predict the next state \par

\makebox[1.5cm]{$\pmo$} algorithmic phase in which the transformer implements \(1\)-Mem \par 

\makebox[1.5cm]{$\hat{\pi}^{\text{2-Mem}}$}
Bayes optimal predictor which uses two-point statistics to infer the chain from $\mathcal{S}$ that generated a sequence and predict the next state \par

\makebox[1.5cm]{$\pmt$} algorithmic phase in which the transformer implements \(2\)-Mem \par 

\makebox[1.5cm]{$K_1^\ast$} first data diversity threshold, above which the model enters \(\pgt\) at some point during training \par 

\makebox[1.5cm]{$K_2^\ast$} second data diversity threshold, above which the model remains in \(\pgt\) indefinitely \par 

\makebox[1.5cm]{$n_\tau$}  number of occurrences of state $\tau$ in a given sequence \par

\makebox[1.5cm]{$m_{\tau\mu}$}  number of occurrences of the transition \(\mu \rightarrow \tau\) in a given sequence \par

\subsection{Behavioral and mechanistic readouts}

\makebox[1.5cm]{$\phi_\delta^{(\ell)}$}
order parameter indicating selective attention to the previous sequence position in the layer \(\ell\) attention block \par

\makebox[1.5cm]{$\phi_{\beta}^{(\ell)}$}
order parameter indicating selective induction-head-like attention in the layer \(\ell\) attention block \par

\makebox[1.5cm]{$n_A^{(\ell)}$} entropy-based measure of the effective number of sequence positions attended to by the layer \(\ell\) attention block

\subsection{Minimal model for 2-Gen}
\makebox[1.5cm]{$\delta$}    effective previous state positional bias parameter in the first attention layer \par

\makebox[1.5cm]{$\beta$}     effective mean of the diagonal entries in the dot product between key and query matrices in the second attention layer
\par

\makebox[1.5cm]{$w_{\{A,B,C,D\}}$}   mixture weight of each expert in minimal model \par

\makebox[1.5cm]{$k_{\tau n}$}  number of occurrences of state $\tau$ in the first $n$ states of a given sequence

\subsection{Memorization and task vectors}
\makebox[1.5cm]{$\varphi$}  Task vector, representation of the generating chain produced internally by the \(\pmt\) transformer \par

\makebox[1.5cm]{$D_{\varphi}$}  Dimension of the task vector in the minimal model \par

\newpage

\section{Transformer architecture}


\subsection{Data generation}\label{sec:data_gen}
We follow a meta-training setup, where the parameters of a transformer are optimized on data generated from multiple distinct stochastic processes (`tasks'). Each process is specified by a Markovian transition matrix over $C$ states. The transformer is trained on a fixed set of chains $\mathcal{S} = \{T^{(1)},T^{(2)},\dots, T^{(K)}\}$. These transition matrices are sampled from a distribution over $C \times C$ transition matrices, which we denote $\mathcal{D}_T$. In our model, each column of each transition matrix is independently and identically drawn from a symmetric Dirichlet distribution, $\text{Dir}(\alpha=1, C=10)$, where $\alpha$ is the Dirichlet concentration hyperparameter. The number of transition matrices ($K$) in $\mathcal{S}$ is a hyperparameter fixed before training.

During training, we first sample a transition matrix $T$ from among the $K$ transition matrices in $\mathcal{S}$. A sequence of $N$ states is then sampled from $T$, beginning from a state drawn from the stationary distribution of $T$. We denote a sequence generated in this manner as $S_N\sim T$. To probe generalization, we sample \emph{out-of-distribution} sequences of $N$ states from a new transition matrix drawn from $\mathcal{D}_T$.  
Each state is implicitly represented as a \(C\)-dimensional one-hot vector (\(x_\tau\)) before it is passed into the transformer.





\subsection{Network architecture}\label{sec:transformerinfo}
Here, we describe our primary architecture, which is the two-layer transformer illustrated in Figure~\ref{fig:Background}c. A transformer takes a sequence of vector-embedded states as input and produces a probability distribution over the next state as output~\cite{vaswani2023attentionneed}. Since the data generating process generates a sequence over $C$ states, each sequence is first mapped to a sequence of \(D\)-dimensional vectors via a learnable embedding matrix, $W_E \in \mathbb{R}^{D \times C}$ which maps each state's one-hot representation to $D$ dimensions. We typically refer to the D-dimensional representation of a state in a sequence as a \emph{state}.

Each layer \(\ell\) of the transformer consists of a self-attention block followed by a position-wise feedforward (MLP) block. Both blocks first normalize their input states via \textit{layer-normalization}:
\begin{gather}
    \text{LN}(x; \mu, \sigma^2) = \frac{x - \mu}{\sqrt{\sigma^2+\epsilon}}, \nonumber\\
    \mu(x) = \frac{1}{D}\sum_{i=1}^D x_i, \nonumber\\
    \sigma^2(x) = \frac{1}{D}\sum_{i=1}^D (x_i - \mu)^2, \label{eq:layernorm}
\end{gather}
where \(\epsilon=10^{-5}\) is a small constant added for numerical stability. The layer normalization acts state-wise when applied to a state sequence. For the typical case when the statistics are computed from the input, we write \(\text{LN}(x) \equiv \text{LN}(x; \mu(x), \sigma^2(x))\). We write \(\overline{x} \equiv \text{LN}(x)\) to denote a vector normalized in this manner.

Self-attention computes a scalar score for a pair of states which depends on the query state \(x_q\), key state \(x_k\), and their relative distance \(r\) in the sequence
\begin{equation}
    s^{(\ell)}(x_q, x_k, r) \equiv \frac{1}{\sqrt{D}}\overline{x}_q^\top W_Q^{(\ell)\top} W_r W_K^{(\ell)} \overline{x}_k,
\end{equation}
where \(W_Q^{(\ell)},\,W_K^{(\ell)} \in \mathbb R^{D\times D}\) are learnable \emph{query} and \emph{key matrices}, \(W_Qx_q\) is the \emph{query vector}, and $W_Kx_k$ is the \emph{key vector}. The fixed matrix \(W_r \in \mathbb R^{D \times D}\) implements the rotary positional encoding~\cite{su2024roformer}, and is suppressed from the notation going forward. These scores are normalized to form \emph{attention weights} between the current state and each sequence state
\begin{equation}
    A_{ni}^{(\ell)} \equiv \frac{e^{s^{(\ell)}\left(x_n^{(\ell-1)}, x_i^{(\ell-1)}, n-i\right)}}{\sum_{j \leq n} e^{s^{(\ell)}\left(x_n^{(\ell-1)},x_j^{(\ell-1)}, n-j\right)}}\quad \text{for } i\leq n.
\end{equation}
The output of the attention block is a sum over the sequence weighted by the attention weights
\begin{equation}\label{eq:full_att_func}
    \text{Att}^{(\ell)}_n\left(x_{i\leq n}^{(\ell-1)}\right) \equiv\sum_{i\leq n}A_{ni}^{(\ell)} W^{(\ell)}_V \overline{x}_i^{(\ell-1)},
\end{equation}
where \(W^{(\ell)}_V\in\mathbb R^{D\times D}\) is a learnable \emph{value matrix} and \(W^{(\ell)}_V \bar{x}_i^{(\ell-1)}\) is the \emph{value vector} for state \(i\).

Each attention block is followed by a position-wise feedforward (MLP) block. The MLP block consists of a single hidden layer with hidden dimension \(d_{\mathrm{hidden}} = 4D\) and Gaussian Error Linear Unit (GELU) activation function~\cite{hendrycks2023gaussianerrorlinearunits}, and is computed
\begin{revblock}
\begin{equation}
\text{MLP}^{(\ell)}\left(y_n^{(\ell-1)}\right) \equiv W_{\text{MLP},U}^{(\ell)}\operatorname{GELU}\left(W_{\text{MLP},E}^{(\ell)}\overline{y}_n^{(\ell-1)}\right),
\end{equation}
\end{revblock}
where \(W_{\text{MLP},E}^{\top(\ell)}, \, W_{\text{MLP},U}^{(\ell)} \in \mathbb R^{D \times 4D}\) and the activation acts point-wise \(\rev{\operatorname{GELU}(x)}\equiv x\cdot\frac{1}{2}\left[1 + \text{erf}(x/\sqrt2)\right]\).

Since each block adds its output to its input, we can recursively construct the current state embedding after the operation of each block. This quantity is commonly referred to as the \emph{residual stream}, which tracks the evolution of the state representations over the transformer layers. Beginning from the state embedding, the residual stream after each block operation is

\begin{align}
    x_n^{(0)} &= W_Ex_n \label{eq:model_start} \\
    y^{(1)}_n &= x^{(0)}_n + \text{Att}^{(1)}\left(x^{(0)}_{\leq n}\right) \\
    x^{(1)}_n &= x^{(0)}_n + \text{Att}^{(1)}\left(x^{(0)}_{\leq n}\right) + \text{MLP}^{(1)}\left(y_n^{(1)}\right) \\
    y^{(2)}_n &= x^{(0)}_n + \text{Att}^{(1)}\left(x^{(0)}_{\leq n}\right) + \text{MLP}^{(1)}\left(y_n^{(1)}\right) + \text{Att}^{(2)}\left(x^{(1)}_{\leq n}\right)\label{res3} \\
    x^{(2)}_n 
    &= x^{(0)}_n + \text{Att}^{(1)}\left(x^{(0)}_{\leq n}\right) + \text{MLP}^{(1)}\left(y_n^{(1)}\right) + \text{Att}^{(2)}\left(x^{(1)}_{\leq n}\right) + \text{MLP}^{(2)}\left(y_n^{(2)}\right) \label{eq:model_end}
\end{align}

Note that we denote the residual stream after the layer \(\ell\) attention block as \(y_n^{(\ell)}\) and the residual stream after the MLP block as \(x_n^{(\ell)}\). The remaining operation of the transformer is a linear output projection $W_U \in \mathbb{R}^{D \times C}$ which is applied to the residual stream after the final MLP block to produce the \(C\)-dimensional logit vector \(x_n^{(3)}\)
\begin{revblock}
\begin{gather}
\text{Linear}(x) \equiv W_U^\top \bar{x} \\
x_n^{(3)} = \text{Linear}(x_n^{(2)})
\end{gather}
\end{revblock}
to which a softmax is applied to obtain a predictive distribution over the next state \(\hat\pi_n^\theta(\tau) = e^{x_{n,\tau}^{(3)}}/\sum_{\tau'}e^{x_{n,\tau'}^{(3)}}\). Going forward, we write the output predictive distribution of the transformer given an input sequence \(S_n\) as \(\hat{\pi}^\theta(\tau \mid S_n)\).

The model weights are initialized by independently sampling from a normal distribution with zero mean and variance $1/D$ except for weights that project to the residual stream, which are sampled from a normal distribution with zero mean and variance $1/4D$. This follows the GPT-2 initialization scheme~\cite{radford2019language}.

\subsection{Training process}\label{sec:training_AR}
The parameters of the model are optimized by minimizing the auto-regressive cross-entropy sequence prediction loss,
\begin{equation}
    \mathcal L_\text{train}(\hat\pi^\theta) = \expval{-\frac{1}{N}\sum_{n=1}^N \log \hat \pi^\theta(s_{n+1}\mid S_n)}_{\substack{T\sim\mathcal{S} \\ S_{N+1}\sim T}},
\end{equation}
where $S_n = (s_1,s_2,\dots,s_n)$ for all $n$. We drop the argument and write \(\mathcal{L}_\text{train}(\hat\pi) = \mathcal{L}_\text{train}\) when the predictor the loss is evaluated for is clear from context. A transition matrix $T$ is sampled independently and uniformly at random from $\mathcal{S}$ for each sequence in the batch.

Minimization of this objective is accomplished through minibatch stochastic gradient descent. At each training step, the model's loss is computed on a batch of $B = 128$ sequences and the model's weights are updated via the AdamW optimizer~\cite{loshchilov2019decoupledweightdecayregularization} with learning rate $\gamma = 10^{-3}$, beta parameters $(\beta_1, \beta_2) = (0.9, 0.95)$, and weight decay $\lambda = 10^{-3}$. 

\newpage

\section{Defining the four algorithmic phases}\label{app:4phases}
\subsection{Form of the Bayes predictor}
Across task diversities and over the course of training, we find that a transformer's behavior can be well-characterized by four algorithms (illustrated in Figure \ref{fig:Background}\rev{d}). These algorithms are specific implementations of the Bayes-optimal predictor, which first infers the underlying transition matrix from an observed sequence \(S_N\) and then predicts the distribution \(\hat \pi(s_{n+1} = \tau |s_n = \mu, S_n)\) over next states \(\tau\) conditioned on the current state \(\mu\) accordingly. Specifically, 
\begin{align}
   \hat \pi_n(\tau |\mu) \equiv \hat \pi(s_{n+1} = \tau |s_n = \mu, S_n)  &= 
     \int dT\, P(T\mid S_n) T_{\tau \mu}, \\
    &= \int dT\,\frac{P(S_n\mid T)P(T)}{P(S_n)} T_{\tau \mu }.
    \label{Tinference}
\end{align}
The four algorithms vary in the choice of prior $P(T)$ and the choice of sequence statistic (1-point or 2-point, see Figure \ref{fig:Background}b) when computing the likelihood $P(S_n\mid T)$.

\subsection{Memorizing  predictors}
We model memorization by ideal predictors that have complete information about the transition matrices in $\mathcal{S}$. In this case, memorizing predictors correspond to the general Bayes predictor (Eq.~\ref{Tinference}) when the prior $P(T)$ matches $\mathcal{S}$. Making the substitution $P(T)=\frac{1}{K}\sum_{k=1}^K\delta(T-T^{(k)})$, we have
\begin{align}
    \hat \pi_n^{\text{Mem}}(\tau |\mu)
    &= \frac{1}{K}\sum_{k=1}^K \frac{P(S_n\mid T^{(k)})}{P(S_n)} T^{(k)}_{\tau\mu}.
\end{align}
Such memorizing predictors can be further classified into two types based on the choice of sequential statistics in $P(S_n\mid T^{(k)})$.

\paragraph{1-point memorization.}
Computing the likelihood using 1-point sequence statistics,
\begin{align}
    P(S_n\mid T^{(k)}) = \prod_{i=1}^n P(s_i\mid T^{(k)}),
\end{align}
yields the \textit{1-Mem predictor},
\begin{align}
    \hat \pi^\text{1-Mem}_n(\tau | \mu)
    &= \frac{1}{K}\sum_{k=1}^K \frac{\prod_{i=1}^n P_k(s_i)}{P(S_n)}T^{(k)}_{\tau\mu}, 
\end{align}
where $P_k(s_i) = P(s_i\mid T^{(k)}) $ is the stationary distribution over states for transition matrix $T^{(k)}$.

\paragraph{2-point memorization.}
Computing the likelihood from 2-point sequence statistics,
\begin{align}
    P(S_n\mid T^{(k)}) = P_k(s_1)\prod_{i=2}^n P(s_i \mid s_{i-1}, T^{(k)})
\end{align}
in turn yields the \textit{2-Mem predictor},
\begin{align}
     \hat \pi^\text{2-Mem}_n(\tau | \mu)
    &= \frac{1}{K}\sum_{k=1}^K \frac{P_k(s_1)\prod_{i=2}^n T^{(k)}_{s_is_{i-1}}}{P(S_n)} T^{(k)}_{\tau\mu}
\end{align}
where $T^{(k)}_{s_is_{i-1}} = P(s_i\mid s_{i-1}, T^{(k)})$.

If the posterior over the transition matrices collapses to the true transition matrix $T^{(k)}$ from which the sequence is drawn (which is possible when \(K\) is small and \(n\) is large), we expect both predictors to recover the oracle solution, namely,
\[
\hat \pi^\text{2-Mem}_n(\tau | \mu) =
\hat \pi^\text{1-Mem}_n(\tau | \mu)
=
T^{(k)}_{\tau\mu}.
\]
More generally, however, \(\hat \pi^\text{2-Mem}_n(\tau | \mu)\) outperforms
\(\hat \pi^\text{1-Mem}_n(\tau | \mu)\), as the 2-point statistics enables more accurate discrimination when the data diversity is high.

\subsection{Generalizing predictors}
We model generalization by ideal predictors that predict the next state given complete knowledge of the data distribution \(\mathcal{D}_T\) from which the transition matrices in \(\mathcal{S}\) are sampled. Generalizing predictors thus correspond to the Bayes-optimal predictors (Eq.~\ref{Tinference}) when the assumed prior \(P(T)\) matches \(\mathcal{D}_T\).

\paragraph{1-point generalization.}
For 1-point statistics, the relevant prior corresponds to the distribution over stationary distributions induced by transition matrices \(T \sim \mathcal{D}_T\). Since the exact form of this distribution is unknown, we approximate it using a Dirichlet prior \(\mathrm{Dir}(\alpha', C)\). Under this assumption, a standard Bayesian calculation yields the 1-Gen predictor, which predicts by counting the frequency of the target state in the finite sequence of length $n$:
\begin{align}
 \hat \pi^\text{1-Gen}_n(\tau | \mu)  = \frac{n_\tau + \alpha'}{n + C\alpha'},
\end{align}
where \(n_\tau\) denotes the number of occurrences of state \(\tau\) in the length-$n$ sequence. The role of $\alpha'$ is to act as a regularizer when $n$ is small. Throughout our analysis, we fix \(\alpha' = 1\).


In the limit \(n \to \infty\), this estimate converges to the stationary distribution \(p_\tau\) of $T$. The minimum cross-entropy loss in this regime is therefore given by
\begin{align}
\mathcal{L}^{\text{1-Gen}}
=
-
\left\langle
\sum_\tau p_\tau \log p_\tau
\right\rangle,
\end{align}
which corresponds to the optimal generalizing loss when no information about the preceding state \(\mu\) is available.


\paragraph{2-point generalization.}
For 2-point statistics, the prior corresponds to the distribution over next-state frequencies conditioned on the current state. For \(\mathcal{D}_T\), this prior is exactly a Dirichlet distribution \(\mathrm{Dir}(\alpha=1, C)\) applied independently to each row of the transition matrix. Consequently, the posterior factorizes over the rows and a standard Bayesian calculation yields the \textit{2-Gen predictor}
\begin{align}\label{eq:2-Gen}
 \hat \pi^\text{2-Gen}_n(\tau | \mu) 
=
\frac{m_{\tau\mu} + 1}{n_\mu + C},
\end{align}
where $m_{\tau\mu}$ denotes the number of occurrences of state $\tau$ given that the preceding state is the current state $\mu$ (note $\sum_\tau m_{\tau\mu} = n_\mu $).

In the large-$n$ limit, we can show $ \hat \pi^\text{2-Gen}_n(\tau | \mu)  \to T_{\tau\mu}$  through $m_{\tau\mu} = np_\mu T_{\tau\mu} = n_\mu T_{\tau\mu}$. 
The optimal 2-point cross-entropy loss is
\begin{align}
\mathcal{L}^{\text{2-Gen}}
=
-
\left\langle
\sum_{\mu,\tau}
p_\mu T_{\tau\mu}
\log T_{\tau\mu}
\right\rangle.
\end{align}
Moreover, in the large-$n$ limit, 2-Gen has at most the loss incurred by 1-Gen:
\begin{align}
\mathcal{L}^{\text{2-Gen}} - \mathcal{L}^{\text{1-Gen}}
=
\left\langle
\sum_{\mu,\tau}
p_\mu T_{\tau\mu}
\log\frac{p_\tau}{T_{\tau\mu}}
\right\rangle
\le 0,
\end{align}
where we have used the inequality \(\log x \le x - 1\).
Equality holds if and only if \(T_{\tau\mu} = p_\tau\), i.e., when transitions are independent of the current state.

\paragraph{Optimal generalizing solutions for models with $n$-point dependencies.}
Here, we provide a mathematical argument that formalizes the intuition that an $n$-point model is sufficient to describe the optimal general solution when the next state depends at most on the preceding $n$ states. Because the data-generating process in our experiments is first-order Markovian, the 2-point model emerges as the optimal general solution, even when longer histories are available. For simplicity, we only provide the argument for large and fixed (i.e., not autoregressive) $n$.

We begin by considering a predictor $\pi$ that does not condition on any preceding state. In this case, the expected cross-entropy objective can be written as
\begin{align}
\mathcal{C}(\pi)
=
-
\left\langle
\sum_{\mu} p_\mu
\sum_\tau T_{\tau\mu} \log \pi_\tau
+
\lambda\left(\sum_\tau \pi_\tau - 1\right)
\right\rangle,
\end{align}
where \(\lambda\) is a Lagrange multiplier enforcing normalization of \(\pi\).
Minimizing \(\mathcal{C}(\pi)\) yields the optimal solution \(\pi_\tau = p_\tau\), corresponding to the 1-Gen predictor.

Next, suppose that the predictor conditions only on the immediately preceding state \(\mu\).
The objective becomes
\begin{align}
\mathcal{C}(\pi)
=
-
\left\langle
\sum_{\mu}
\left[
p_\mu
\sum_\tau T_{\tau\mu} \log \pi_{\tau|\mu}
+
\lambda_\mu
\left(\sum_\tau \pi_{\tau|\mu} - 1\right)
\right]
\right\rangle,
\end{align}
where \(\lambda_\mu\) enforces normalization of \(\pi_{\tau|\mu}\) for each \(\mu\).
Optimizing with respect to \(\pi_{\tau|\mu}\) yields
\begin{align}
\pi_{\tau|\mu}
=
\frac{p_\mu T_{\tau\mu}}{\lambda_\mu}.
\end{align}
Imposing the normalization constraint gives \(\lambda_\mu = p_\mu\), and therefore
\(\pi_{\tau|\mu} = T_{\tau\mu}\), which is precisely the 2-Gen predictor. Finally, if the predictor is allowed to condition on two preceding states \((\nu, \mu)\), the objective can be written as
\begin{align}
\mathcal{C}(\pi)
=
-
\left\langle
\sum_{\nu,\mu}
\left[
p_\nu T_{\mu\nu}
\sum_\tau T_{\tau\mu} \log \pi_{\tau|\mu,\nu}
+
\lambda_{\mu\nu}
\left(\sum_\tau \pi_{\tau|\mu,\nu} - 1\right)
\right]
\right\rangle.
\end{align}
Optimizing this objective again yields
\(\pi_{\tau|\mu,\nu} = T_{\tau\mu}\),
which coincides with the 2-Gen predictor. Thus, although a longer history is available, the Markov property of the data-generating process implies that the optimal general solution is always the 2-Gen predictor.

\subsection{Scaling of the four Bayes predictors with $K$ and $N$}
\begin{figure}[H]
\begin{nolinenumbers}
\centering
\includegraphics[]{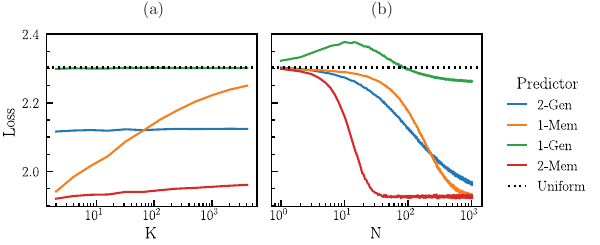}
    \caption{
    Scaling of the four predictor losses with data diversity \(K\) and sequence length \(N\) computed over a large batch of sequences and averaged over 8 task distributions \(\mathcal{D}_T\). The \(N=256\) autoregressive loss is shown for \(K\) scaling. \(N\) scaling of the non-autoregressive loss is shown at data diversity \(K=128\).
    } 
\label{fig:predictor_scaling}
\end{nolinenumbers}
\end{figure}

Figure~\ref{fig:predictor_scaling} compares the loss incurred by the four Bayesian predictors as a function of data diversity $K$ and sequence length $N$ on training sequences.

\paragraph{Dependence on data diversity ($K$).}
In Figure~\ref{fig:predictor_scaling}a, we plot the autoregressive loss at fixed context length $N=256$ while varying $K$. The 1-Mem predictor degrades as $K$ increases. This behavior is expected: when many processes may have generated a sequence, short sequences provide insufficient information to reliably identify which transition matrix the sequence is sampled from. A notable feature is a cross-over between the 1-Mem and 2-Gen curves, suggesting that the loss landscape favors a different solution with increasing $K$. Thus, the preferred predictor for training sequences switches from 1-Mem to 2-Gen.

By contrast, the other three predictors remain comparatively stable across a broad range of $K$. In the idealized limit of exact Bayesian inference with unbounded memory, we therefore expect the ordering
\[
\mathcal{L}^{\text{2-Mem}} < \mathcal{L}^{\text{2-Gen}} < \mathcal{L}^{\text{1-Gen}},
\]
for our experimental setup. While this ordering applies for ideal Bayes predictors, architectural constraints could reverse the comparison between 2-Gen and 2-Mem at large $K$, that is, $\mathcal{L}^{\text{2-Gen}} < \mathcal{L}^{\text{2-Mem}}$.

\paragraph{Dependence on sequence length ($N$).}
In Figure~\ref{fig:predictor_scaling}b, we fix $K=128$ and vary the sequence length. All four predictors improve with increasing $N$, but for different reasons.

For memorization, the dominant bottleneck is task inference: the estimator needs enough evidence in the sequence to identify the latent task. Consistent with this, the 2-Mem predictor approaches its asymptotic performance at $N \sim 10^2$, whereas the 1-Mem predictor requires substantially longer sequences, $N \sim 10^3$.

For generalization, the limiting factor is finite-sample noise in estimating transition statistics from the sequence. Consequently, the loss decreases more gradually with $N$ than for the memorization predictors. The 1-Gen predictor performs worst throughout, as it ignores sequential structure and predicts only the stationary distribution. The 2-Gen predictor steadily improves and becomes comparable to the optimal predictor (2-Mem) only at much longer sequence lengths, $N \sim 10^4$.

As in the $K$-scaling, we again observe a crossover between the 1-Mem and 2-Gen predictors, consistent with a transition between a task-identification regime (small $N$/small $K$) and a statistics-estimation regime (large $N$/large $K$).

\section{Phase identification}\label{app:phase_idenfication}
We trained the transformer model (Section \ref{sec:transformerinfo}) in settings where the data diversity $K$ was varied over powers of 2 from $2^3$ to $2^{10}$, corresponding to logarithmically spaced diversities between 8 and 1024 transition matrices. A total of $10^6$ iterations of stochastic gradient descent were performed in each setting, with checkpoints of the model state saved at logarithmically spaced intermediate iterations to allow for post-training evaluation. The training task distributions for a single data diversity sweep were constructed such that \(\mathcal{S}\) at data diversity \(k\) is a subset of all \(\mathcal{S}^\prime\) for task diversities \(k^\prime\geq k\).

Across data diversities and throughout training, we find that transformer behavior is well-characterized by one of the four predictors discussed in the previous section. These observations reproduce those found in \cite{park2024competition}. We denote these regimes $\pgo$, $\pmo$, $\pgt$ and $\pmt$ which correspond to the transformer implementing \(1\)-Gen, \(1\)-Mem, \(2\)-Gen, and \(2\)-Mem respectively. This characterization is supported by two complementary observations:
\begin{enumerate}
    \item  \textbf{Behavioral readouts}: Close agreement between the predictions of the transformer and a Bayesian predictor.
    \item \textbf{Mechanistic readouts}: Attention patterns with clear interpretations.
\end{enumerate}

\subsection{Behavioral readouts}\label{sec:behave_readouts}

\begin{figure}[H]
\begin{nolinenumbers}
\centering
\includegraphics[width=0.5\linewidth]{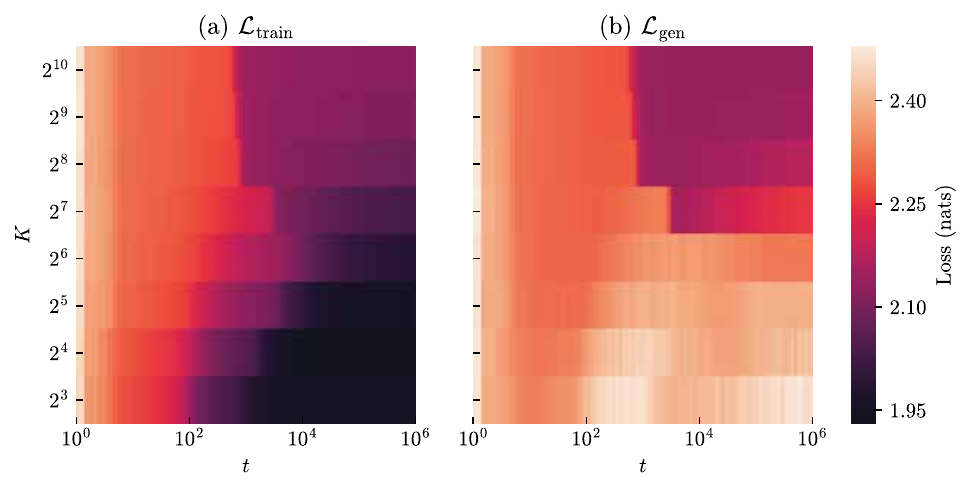}
    \caption{
    \rev{
    Training and generalization losses for the same models shown in Figure~\ref{fig:Phases}.
    }}
\label{fig:loss_heatmap}
\end{nolinenumbers}
\end{figure}

We evaluated the training and generalization loss of each model checkpoint on a sample of $8 \times K$ and 2048 sequences respectively, where the generalization loss is given by
\begin{equation}
    \mathcal L_\text{gen} = \expval{-\frac{1}{N}\sum_{n=1}^N \log \hat \pi_\theta(s_{n+1}\mid S_n)}_{\substack{T\sim\mathcal{D}_T \\ S_{N+1}\sim T}}.
\end{equation}
This allowed us to compare the loss values through training to the loss of each predictor on the same sequences. We find plateaus in both the training and generalization losses, which correspond closely with loss values of the predictors (Figure~\ref{fig:Phases}a \rev{, Figure~\ref{fig:loss_heatmap} visualizes these trajectories for all $K$.}). To construct a quantitative metric, we compute the divergence between a predictor $\hat \pi\in \{\hat{\pi}^{\text{1-Mem}}, \hat{\pi}^{\text{2-Mem}}, \hat{\pi}^{\text{1-Gen}},\hat{\pi}^{\text{2-Gen}} \}$ and transformer model $\hat \pi_\theta(S_n)$ as
\begin{align}
    \text{D}(\hat\pi, \hat\pi_\theta)\equiv \frac{1}{2}\expval{D_\text{KL}^{S_N}(\hat\pi\,\|\, \hat\pi_\theta)}_{\substack{T\sim\mathcal{S} \\ S_N\sim T}} + \frac{1}{2}\expval{D_\text{KL}^{S_N}(\hat\pi \,\|\, \hat\pi_\theta)}_{\substack{T\sim \mathcal{D}_T \\ S_N\sim T}}, 
\end{align}
where $D_\text{KL}^{S_N}(\hat\pi\,\|\,\hat \pi_\theta) \equiv \frac{1}{N}\sum_{n=1}^N D_\text{KL}\left(\hat\pi(S_n) \,\|\, \hat\pi_\theta(S_n)\right)$.
In practice, for a model trained on $K$ tasks, the expectation values over $\mathcal{S}$ and $\mathcal \mathcal \mathcal{D}_T$ are approximated by averaging over $8 \times K$ and $2048$ sequences respectively.

In Figure \ref{fig:Phases}b, the plots of the divergence to each predictor in the $(K,t)$ parameter space reveal that the transformer's response almost always approximates one of the four predictors, and that these regimes form contiguous regions of parameter space. We refer to these regions as algorithmic phases of the transformer.

\subsection{Mechanistic readouts}\label{sec:mech_readouts}
Only the attention layers of the transformer architecture can mix information along the sequence dimension. Consequently, for the model to infer nearest-neighbor 2-point correlations (i.e., bigrams) in a sequence, at least one attention layer must attend to the previous state. To identify when an attention layer exhibits this behavior, we compute the expected average attention weight assigned to the previous state across a sequence:
\begin{align}\label{pta}
\phi_\delta^{(\ell)} = \expval{\frac{1}{N-1}\sum_{i=2}^N A_{i,i-1}^{(\ell)}}_{\substack{T\sim\mathcal{S} \\ S_N\sim T}} .
\end{align}
When the layer does not preferentially attend to the previous state, $\phi_\delta^{(\ell)} = \mathcal{O}(1/N)$, whereas $\phi_\delta^{(\ell)} = 1$ when the layer attends exclusively to the previous state. We therefore refer to $\phi_\delta^{(\ell)}$ as the \textit{previous-state order parameter} of the attention layer. Empirically, we find that only the first attention layer attains a non-negligible value of $\phi_\delta^{(\ell)}$, reflecting the necessity for 2-point information to be extracted in the first layer so that it can be aggregated across the sequence by the second attention layer.

The $\pgt$ phase of the transformer is implemented through the composition of the first and second attention layers, forming a `statistical induction head' which we discuss at length in Appendix~\ref{sec:Induction_head} ~\cite{reddy2023mechanistic, bietti2023birthtransformermemoryviewpoint, edelman2024evolutionstatisticalinductionheads}. A defining feature of an induction head circuit is that the second-layer attention concentrates on sequence positions corresponding to states that follow the current state in the sequence. To detect this behavior, we compute the expected total attention weight that the final state assigns to all states that immediately follow occurrences of itself in the sequence:
\begin{align}\label{iha}
\phi_\beta^{(\ell)} \equiv \expval{\frac{1}{N}\sum_{n=2}^N\sum_{i=2}^n \delta_{s_n, s_{i-1}}A_{ni}^{(\ell)}}_{\substack{T\sim\mathcal{S} \\ S_N\sim T}}
\end{align}
Since $\phi_\beta^{(2)}$ becomes non-zero when an induction head circuit is formed, we refer to $\phi_\beta^{(2)}$ as the \textit{induction-head order parameter} of the second attention layer.

In Figure \ref{fig:Phases}c, the upper two plots of the order parameters in the $K, t$ parameter space reveal regions of nontrivial $\phi_\delta^{(1)}$ and $\phi_\beta^{(2)}$ which are in strong correspondence with the behavioral readouts indicating general 2-point utilization and $\pgt$, respectively.

To complement the above order parameters, which indicate the presence of highly specific attention patterns, we compute the effective number of sequence states an attention block attends to at the final position:
\begin{gather}
n_A^{(\ell)} \equiv e^{H}, \\
H \equiv
\expval{
-\sum_{i\le n} A_{ni}^{(\ell)}\log A_{ni}^{(\ell)}
}_{\substack{T\sim\mathcal{S} \\ S_N\sim T}}
\label{eq:na}
\end{gather}
where \(H\) is the expected entropy of the attention map for the largest sequence length \(N\). When an attention layer has \(n_A^{(\ell)} \approx N\), the layer can be well approximated by an average of the states over the sequence.

In practice, the expectation value over sequences in equations \ref{pta},~\ref{iha}, and~\ref{eq:na} is approximated by averaging over a batch of $2048$ sequences (with the exception of the results in Figure~\ref{fig:K1}a, which were evaluated on $4096$ sequences).


\subsection{Hyperparameter variations}
\begin{revblock}
To better understand the generality of the identified algorithmic phases, we trained models with variations of architectural and dataset parameters relative to the baseline configuration. The behavioral readouts from these models reveal that the phase structure is maintained across a wide variety of settings(Figure~\ref{fig:hyperparameter_variatiions}).

\begin{figure}[H]
\begin{nolinenumbers}
\centering
\includegraphics[width=0.9\linewidth]{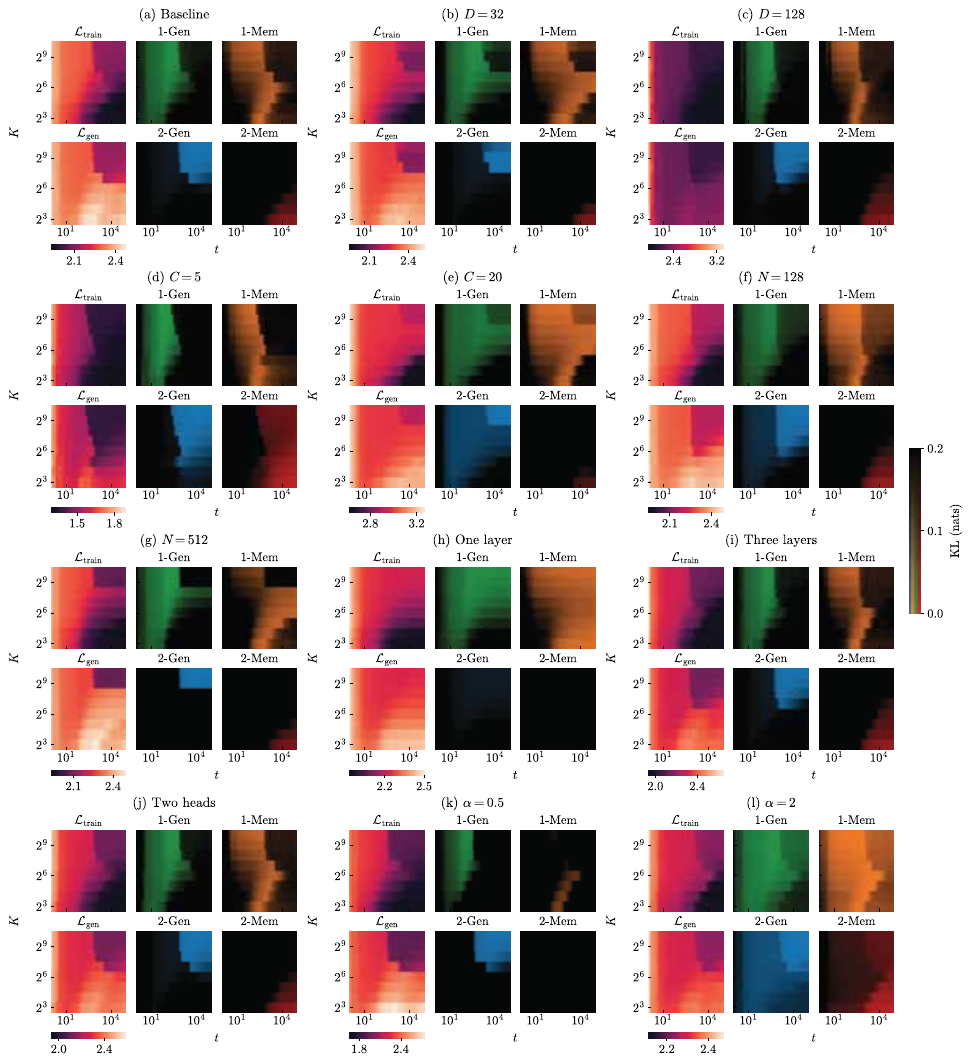}
    \caption{
    \rev{
    Training and generalization loss dynamics, and behavioral readouts as $K$ is varied for a variety of architectural and dataset modifications. The baseline phase diagram is reproduced in (a), and is truncated to $t \leq 10^5$ training iterations to ease comparison. The colorbar below each subplot corresponds to loss values. The four phases are present for all models except the one-layer model (h), in which case the absence of both 2-point phases can be understood by the circuit analyses for these phases (Section~\ref{sec:tracing}). For sparse transition distributions (k), the $\pmt$ phase  is indicated by the training loss but not the similarity, likely due to  extreme predictions made out-of-distribution by the Bayes predictor.
    }
    } 
\label{fig:hyperparameter_variatiions}
\end{nolinenumbers}
\end{figure}
\end{revblock}

\section{Transformer circuits}\label{sec:tracing}
We find strong evidence that each of the four algorithmic phases of the model is governed by a sparse circuit. This sparsity enables us to decompose the full, complex network into reduced and interpretable sub-networks, and to study the mechanisms by which different Bayesian algorithms are implemented.

\begin{table}
\begin{nolinenumbers}
\centering
\renewcommand{\arraystretch}{2.5}

\begin{subtable}{\textwidth}
\centering
\begin{tabular}{||c|c||}
\hline
\textbf{Source Name} & \textbf{Vector} \\ [0.5ex]
\hline\hline
State & \(x^{(0)}_n\) \\
\hline
A1(State) &
\(\sum_{i\leq n}A_{ni}^{(1)}
W^{(0)}_V \overline{x}_i^{(0)}\) \\
\hline
MLP1 & \(\text{MLP}^{(1)}\left(y_n^{(1)}\right)\) \\
\hline
A2(State) &
\(\sum_{i\leq n}A_{ni}^{(2)}
W^{(2)}_V \text{LN}\left(x_i^{(0)}; \mu\left(x_i^{(1)}\right), \sigma^2\left(x_i^{(1)}\right)\right)\) \\
\hline
A2(A1) &
\(\sum_{i\leq n}\text{A}_{ni}^{(2)}
W^{(2)}_V \text{LN}\left(\text{Att}^{(1)}\left(x^{(0)}_{\le i}\right); \mu\left(x_i^{(1)}\right), \sigma^2\left(x_i^{(1)}\right)\right)\) \\
\hline
A2(MLP1) &
\(\sum_{i\leq n}\text{A}_{ni}^{(2)}
W^{(2)}_V \text{LN}\left(\text{MLP}^{(1)}\left(y_i^{(1)}\right); \mu\left(x_i^{(1)}\right), \sigma^2\left(x_i^{(1)}\right)\right)\) \\
\hline
MLP2 & \(\text{MLP}^{(2)}\left(y_n^{(2)}\right)\) \\
\hline
\end{tabular}
\caption{Source vector contributions to the residual stream.}
\label{tab:a}
\end{subtable}

\vspace{1em}

\begin{subtable}{\textwidth}
\centering
\begin{tabular}{||c|c||}
\hline
\textbf{Target Name} & \textbf{Perturbation} \\ [0.5ex]
\hline\hline
A[\(\ell\)]Q &
\(\text{A}_n^{(\ell)}(i)
= \frac{e^{s^{(\ell)}\left(\tilde{x}_n^{(\ell-1)}, x_i^{(\ell-1)}, n-i\right)}}{\sum_{j \leq n} e^{s^{(\ell)}\left(\tilde{x}_n^{(\ell-1)},x_j^{(\ell-1)}, n-j\right)}}\) \\
\hline
A[\(\ell\)]K &
\(\text{A}^{(\ell)}(i)
= \frac{e^{s^{(\ell)}\left(x_n^{(\ell-1)}, \tilde{x}_i^{(\ell-1)}, n-i\right)}}{\sum_{j \leq n} e^{s^{(\ell)}\left(x_n^{(\ell-1)}, \tilde{x}_j^{(\ell-1)}, n-j\right)}}\) \\
\hline
MLP[\(\ell\)] & \(\text{MLP}^{(\ell)}\left(\tilde y_n^{(\ell)}\right)\) \\
\hline
Linear & \(\text{Linear}\left(\tilde{x}_n^{(2)}\right)\) \\
\hline
\end{tabular}
\caption{Perturbation targets used in analysis. The perturbed input is marked with a tilde. For the query and key perturbations, all attention weights are perturbed following this prescription.}
\label{tab:b}
\end{subtable}

\caption{Explicit definition of source vectors contributing to the residual stream, and the target perturbations. To ablate the edge from source SRC to a target TGT at layer $\ell$, we compute the perturbed input as \(\tilde x_n^{(\ell-1)} \equiv x_n^{(\ell-1)} - \left(x_{\text{SRC},n} - \expval{x}_{\text{SRC},n}\right)\) where $x_{\text{SRC},n}$ is the source vector corresponding to SRC (Subtable (\subref{tab:a})) and \(\expval{\cdot}\) denotes the batch average. The target is then perturbed during the forward pass by injection of \(\tilde x_n^{(\ell-1)}\) according to the prescription for target TGT (Subtable (\subref{tab:b})). \rev{The layer-normalization each block applies to the perturbed input recomputes the mean $\mu$ from the perturbed input and reuses the variance $\sigma^2$ from the unperturbed input. This allows, in particular, the attention output Eq.~\ref{app:expandedatt2} to be decomposed into additive contributions from earlier blocks.}}
\label{tab:model_decomposition}
\end{nolinenumbers}
\end{table}

\subsection{Path expansion}
We first explain the nature of the circuit edges we consider. Recall the construction of the residual stream after each block
\begin{align}
    x_n^{(0)} &= W_Ex_n \\
    y^{(1)}_n &= x^{(0)}_n + \text{Att}^{(1)}\left(x^{(0)}_{\leq n}\right) \\
    x^{(1)}_n &= x^{(0)}_n + \text{Att}^{(1)}\left(x^{(0)}_{\leq n}\right) + \text{MLP}^{(1)}\left(y_n^{(1)}\right) \\
    y^{(2)}_n &= x^{(0)}_n + \text{Att}^{(1)}\left(x^{(0)}_{\leq n}\right) + \text{MLP}^{(1)}\left(y_n^{(1)}\right) + \text{Att}^{(2)}\left(x^{(1)}_{\leq n}\right)\label{app:res3} \\
    x^{(2)}_n 
    &= x^{(0)}_n + \text{Att}^{(1)}\left(x^{(0)}_{\leq n}\right) + \text{MLP}^{(1)}\left(y_n^{(1)}\right) + \text{Att}^{(2)}\left(x^{(1)}_{\leq n}\right) + \text{MLP}^{(2)}\left(y_n^{(2)}\right) \label{hatpi} \\
    x_n^{(3)} &= \text{Linear}(x_n^{(2)})
\end{align}
Each term in the input to a \rev{block} corresponds to the output of previous \rev{blocks}\footnote{We do not explicitly count edges between \rev{blocks} at different sequence positions}, so we consider these terms as connections from prior \rev{blocks} to the input \rev{block}. This produces $1+2+3+4+5=15$ edges in the transformer graph. This graph is depicted in Figure \ref{fig:Circuits}a.

\paragraph{Attention subcircuit.}\label{sec:attn_subcircuits}
We decompose the attention \rev{block} into three operations and consider each as a node in an attention subcircuit: the computation of the queries, the computation of the keys, and the linear weighted sum along the sequence. Each of these operations read from the full residual stream and thus have 1 and 3 incoming edges in the first and second attention \rev{blocks} respectively. The query and key computations are internal to the attention circuit and do not produce outgoing edges. On the other hand, the sum over the sequence is linear and acts on inputs independently, producing an equal number of outgoing edges as incoming. This observation is trivial for the first attention \rev{block} since it has one input and output regardless. For the second attention \rev{block}, the three outputs correspond to a linear transformation by \(W_V^{(2)}\) of each layer-normalized term of the residual stream summed over the sequence,

\begin{align}
    \text{Att}^{(2)}\left(x_{\leq n}^{(1)}\right) &= \sum_{i\leq n}{A}_{ni}^{(2)} W^{(2)}_V\text{LN}\left(x^{(0)}_i+\text{Att}^{(1)}\left(x^{(0)}_{\leq i}\right) + \text{MLP}^{(1)}\left(y_i^{(1)}\right)\right) \\
    &= \sum_{i\leq n}{A}_{ni}^{(2)} W^{(2)}_V\text{LN}\left(x^{(0)}_i+\text{Att}^{(1)}\left(x^{(0)}_{\leq i}\right) + \text{MLP}^{(1)}\left(y_i^{(1)}\right); \mu\left(x_i^{(1)}\right), \sigma^2\left(x_i^{(1)}\right)\right)\label{app:expandedatt2}
\end{align}
where we have written the layer-normalization explicitly following the definition in Eq.~\ref{eq:layernorm}.

Substituting the attention subcircuit \eqref{app:expandedatt2} into \eqref{app:res3}, $y_n^{(2)}$ has 2 additional terms which result in a total of 6 inputs to MLP2 and 7 inputs to the linear unembedding. Including the key and query computations of each attention \rev{block} as nodes, there are now $2\cdot1+2+2\cdot3 + 6 + 7 = 23$ edges on the graph.

\subsection{Circuit tracing}

\begin{figure}[H]
\begin{nolinenumbers}
\centering
\includegraphics[]{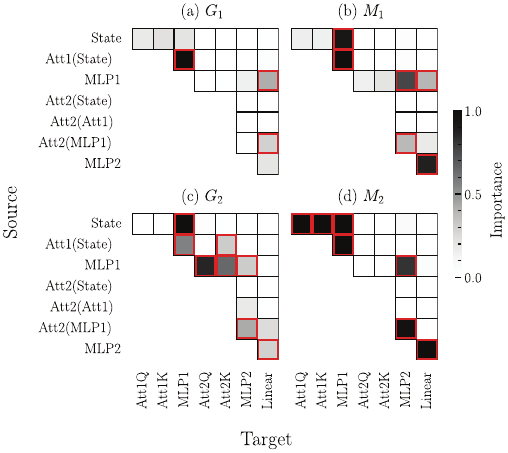}
    \caption{
    Connectivity matrices between the circuit components \rev{as inferred from single-edge ablations}. See Table \ref{tab:model_decomposition} for the definitions of sources and targets. Dark squares indicate edges that demonstrate a large change in model behavior when they are ablated. Generalizing (memorizing) prototype circuits are traced from the \(K=1024\) \rev{(\(K=32\))} models at the training checkpoint with lowest divergence to the corresponding predictor, \rev{with the exception of $\pmt$ which is evaluated at the final checkpoint. Red boxes highlight edges which exceed the threshold importance of $0.25$ for inclusion in the prototype circuits.}
    }
    \label{fig:circuits_raw}
\end{nolinenumbers}
\end{figure}

\rev{Circuit tracing followed a two-stage process in which prototype circuits were identified from single-edge ablations and then refined according to the performance of circuit-constrained models.}

\paragraph{Circuit prototypes} We first measured the importance of each connection in producing the observed transformer behavior. To do so, we developed a custom Python implementation of the model forward pass that explicitly exposes the vector passed along each block connection. This allows the passed vectors along all edges to be cached during a forward pass of the unperturbed model on a batch of $512$ sequences. We then ablated each connection during additional forward passes on the same sequences by replacing the passed vector with the mean passed vector over the batch \cite{wang2022interpretabilitywildcircuitindirect}, and measured for each sequence the KL-divergence of the predictions from the unperturbed model \(D_\text{KL}^{S_N}(\hat\pi^\theta\,\|\,\tilde\pi^\theta)\), where \(\tilde\pi^\theta\) are the perturbed model predictions. The batch average of this divergence forms the edge importance weights, so connections are given small weight if they can be ablated without significantly disrupting the behavior. This process is similar to the path patching method proposed in \cite{goldowskydill2023localizingmodelbehaviorpath}. \rev{Prototype circuits for each phase were then formed by including only connections with (normalized) importance above a threshold of $0.25$. The results of these tracing experiments are visualized as connectivity matrices of the transformer (Figure~\ref{fig:circuits_raw})}.

\begin{figure}[H]
\begin{nolinenumbers}
\centering
\includegraphics[]{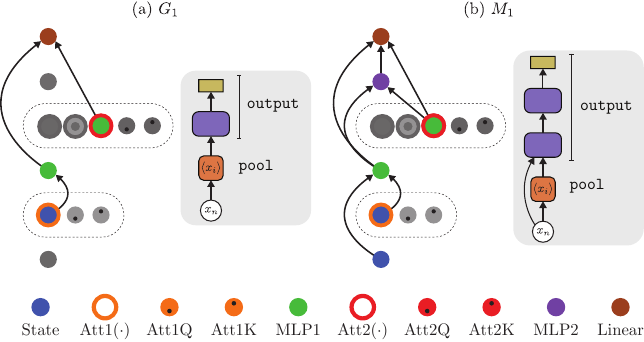}
    \caption{Complete visualization and interpretive schematic of the traced circuits as in Figure \ref{fig:Circuits} for the remaining two phases (a) \(\pgo\) and (b) \(\pmo\).} 
\label{fig:add_circuits}
\end{nolinenumbers}
\end{figure}


\begin{revblock}
\paragraph{Circuit refinement}
The prototype circuits were then tested by simultaneously ablating all omitted edges (following the same batch-mean replacement scheme as earlier) and measuring the same importance score. In addition, perturbations to these prototypes were applied by adding/removing an edge to/from the circuit to verify that the proposed circuit was both sufficient and necessary. For the $\pmt$ circuit, the State $\rightarrow$ Linear edge was found to be unnecessary and we removed it from the identified circuit. For the $\pgt$ circuit, the MLP1 $\rightarrow$ MLP2 edge was identified as unnecessary and removed, and the Att2(MLP1) $\rightarrow$ Linear edge produced significant improvement and was added as a result. For the $\pmo$ circuit, the Att2(MLP1) $\rightarrow$ Linear edge was added. No modifications to the prototype circuit for $\pgo$ was necessary.

We interpret these final, refined circuits as follows.
\end{revblock}

\begin{figure}[H]
\begin{nolinenumbers}
\centering
\includegraphics[]{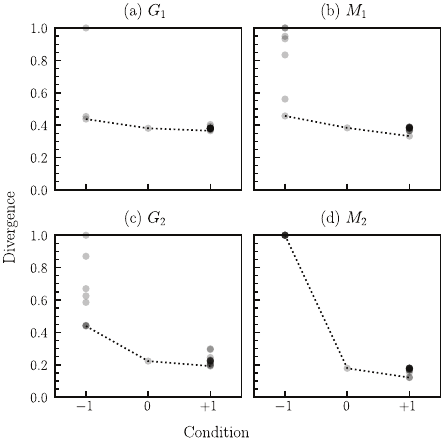}
    \caption{
    \rev{
    Effect of perturbing the circuit-constrained model about the identified circuits. The divergence is normalized by the divergence expected for uniform prediction. Condition values indicate the number of edges either removed or added to the identified circuit. Dashed lines connect the minimum divergence for each condition, and values above $1$ have been clipped. Edges in the identified circuits are necessary (divergence increases when any are removed) and sufficient (divergence is reasonably low, and does not improve much when any edge is added). The representative models correspond to those used for Figure~\ref{fig:circuits_raw}.
    }
    }
    \label{fig:circuit_variations}
\end{nolinenumbers}
\end{figure}

\begin{figure}[H]
\begin{nolinenumbers}
\centering
\includegraphics[]{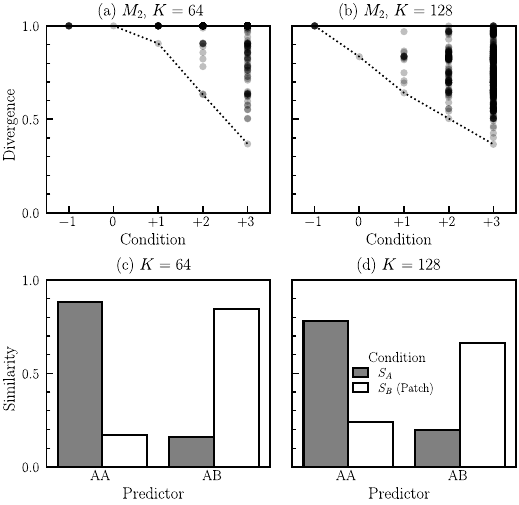}
    \caption{
    \rev{
    (a, b) Results from the circuit perturbation analysis as in Figure~\ref{fig:circuit_variations} for $\pmt$ at larger values of $K$. Models for $K = 64, 128$ do not show strong circuit-constrained performance, and adding edges produces only incremental improvement. (c, d) The task vector patching experiment as in Figure~\ref{fig:task_vector}b for $K = 64$ and $K = 128$. The effect of task vector patching is retained for these $\pmt$ models, despite the indicated circuit complications.}
    }
    \label{fig:m2_triplets}
\end{nolinenumbers}
\end{figure}

\subsubsection{1-Gen}
The 1-Gen predictor requires an estimate of the empirical marginal distribution over states and produces predictions independent of the current state.

In \(\pgo\), the large value of \(n_A^{(1)}\) (Figure~\ref{fig:Phases}\rev{c, bottom left}) indicates that Att1 attends approximately uniformly over all positions in the sequence. Att1 thus computes an average over the sequence states, which effectively encodes the empirical 1-point statistics. A state-wise nonlinear operation which maps this representation to the output logits is sufficient to predict according to the stationary distribution.

The traced circuit (Figure \ref{fig:circuits_raw}(a)) indicates that MLP1 computes the logits for the prediction, with the 1-point statistics read directly from Att1. Since pathways downstream of the current state all have negligible importance, the logit is produced independent of the current state.

In summary, the transformer in \(\pgo\) functions as a simple pool-and-readout circuit implemented by the first layer, with pooling accomplished by Att1 and subsequent readout by MLP1 (Figure \ref{fig:add_circuits}a).

\subsubsection{2-Gen}
The 2-Gen predictor requires an estimate of the empirical frequencies of states which follow the current state \(\mu\), i.e. the quantities \(\frac{m_{\tau\mu}}{n_\mu}\) for each \(\tau\) as shown in \eqref{eq:2-Gen}.

In \(\pgt\), both \(\phi_\delta^{(1)}\) and \(\phi_\beta^{(2)}\) take on large values (Figure~\ref{fig:Phases}\rev{c, upper row}). These order parameters indicate the selective attention of Att1 to the previous state and the selective attention of Att2 to positions in the context which follow the current state $\mu$ respectively. Together, this strongly signals the presence of a statistical induction head, a canonical transformer circuit motif that composes two transformer layers and offers a natural method to compute the empirical distribution needed by 2-Gen. We provide a detailed introduction to the induction head motif in Section \ref{sec:Induction_head}. For now, we interpret how the traced circuit \rev{(Figure \ref{fig:Circuits}b)} maps onto the two basic operations of a statistical induction head:

\paragraph{Pair construction.}
The selective attention of Att1 produces a residual-stream representation of the current and previous states. This representation alone could be sufficient to support the statistical induction head behavior. However, the traced circuit reveals the importance of State \(\rightarrow\) MLP1 and Att1 \(\rightarrow\) MLP 1 pathways. Thus, MLP1 may first denoise or otherwise map the 2-point representation to an embedding space.

\paragraph{Match and readout.}

The match and readout operation of a statistical induction head is often presumed to depend on the composition of the Att2 query with the current state and the Att2 key with the output of Att1, along with Att2 writing the filtered states to the residual stream \cite{olsson2022incontextlearninginductionheads, reddy2023mechanistic}. We would thus expect the pathways State \(\rightarrow\) Att2Q, Att1 \(\rightarrow\) Att2K, and Att2(State) \(\rightarrow\) MLP2 to be of high importance. The traced circuit instead implicates the pathways MLP1 \(\rightarrow\) Att2Q, MLP1 \(\rightarrow\) Att2K, and Att2(MLP1) \(\rightarrow\) MLP2 \(\rightarrow\) Linear. Both constructions appear functionally degenerate, with the resulting operation aggregating the empirical transition frequencies conditioned on $\mu$ and MLP2 reading out this representation to produce the next-state logits.

In summary, the transformer in \(\pgt\) forms a statistical induction head through selective attention over embedded state pairs to produce the empirical distribution of transitions from the current state; the logit is then produced according to this distribution (Figure \ref{fig:Circuits}b).


\subsubsection{1-Mem}

The 1-Mem predictor requires estimating the empirical 1-point statistics from the sequence and using those statistics to infer the latent task identity. Given the inferred task identity and the current state, the next-state probabilities are retrieved and used as predictions.

In \(\pmo\), the large value of \(n_A^{(1)}\) (Figure \ref{fig:Phases}\rev{c, bottom left}) again indicates uniform attention of Att1; the output of Att1 thus encodes the empirical 1-point statistics. Given the current state and this encoding of the 1-point statistics, a state-wise nonlinear operation may be sufficient to predict according to the 1-Mem prescription.

The traced circuit \rev{(Figure \ref{fig:add_circuits}b)} indicates that MLP1 and MLP2 compose to produce the prediction, with the 1-point statistics entering via the Att1 \(\rightarrow\) MLP1 path, and the current state information via the State \(\rightarrow\) MLP1 path.

In summary, the transformer functions as a similar pool-and-readout circuit as in \(\pgo\). However, in \(\pmo\) both MLP blocks are recruited to produce the logit, and the prediction is made conditioned on the current state (Figure \ref{fig:add_circuits}b).

\subsubsection{2-Mem}

The 2-Mem predictor requires estimating the empirical 2-point statistics from the sequence and using those statistics to infer the latent task identity. Given the inferred task identity and the current state, the next-state probabilities are retrieved and used as predictions.

In \(\pmt\), the large value of \(\phi_\delta^{(1)}\) (Figure~\ref{fig:Phases}\rev{c, top left}) indicates the selective attention of Att1 to the previous state. This produces a representation of each sequence transition in the residual stream. Likewise, the large value of \(n_A^{(2)}\) (Figure \ref{fig:Phases}\rev{c, bottom right}) indicates that Att2 produces a representation of information aggregated over the sequence. It is reasonable to expect that the logit may be produced by a state-wise operation given the current state and the aggregated sequence information.

The traced circuit \rev{(Figure \ref{fig:Circuits}c)} reveals the exact nature of this computation. Each transition in the sequence is embedded nonlinearly by MLP1, which operates on the transition representation formed by State and Att1 in the residual-stream. These embeddings are aggregated by Att2 to produce a compact representation of the current task (see Section~\ref{sec:taskvector} for a discussion of this representation). MLP2 receives the current state via the MLP1 \(\rightarrow\) MLP2 path, the inferred task identity via the Att2(MLP1) \(\rightarrow\) MLP2 path, and produces the logits accordingly.

In summary, the \(\pmt\) transformer functions in two stages (Figure \ref{fig:Circuits}c). In the first stage, a compact representation of the inferred task is constructed by the pooling of nonlinear embedding of state pairs across the sequence. In the second stage, the logit is produced given the representation of the inferred task and the current state.

\rev{Note that significantly more edges must be included to recover performance for the $K = 64$ and $128$ circuit-constrained models. (Figure~\ref{fig:m2_triplets}a,b). Despite this, the general features of our $\pmt$ circuit interpretation appear to hold in this regime, most importantly the task vector construction (Figure~\ref{fig:task_vector}b,c). Overall, these results suggest that the $\pmt$ circuit learned for $K_1^\ast < K < K_2^\ast$ continues to leverage the encoder-pool-decoder structure while exhibiting inter-block dependencies not observed for $K < K_1^\ast$.}

\section{Statistical induction head}
\label{sec:Induction_head}

The only sequence information necessary to compute the 2-Gen prediction is the count \(m_{\mu\tau}\) of occurrences of state \(\tau\) that follow the current state \(\mu\) (since the remaining quantity which normalizes the prediction is \(n_\mu = \sum_\tau m_{\mu\tau}\)). The \emph{statistical induction head} is a simple circuit motif commonly observed in transformers that is capable of computing \(m_{\mu\tau}\). Specifically, the statistical induction head computes the empirical conditional distribution \(\hat{P}_n(x_{i+1} = \tau \mid x_i = \mu)\) where \(i\) ranges over the sequence length \(n\). The sufficient circuit for this computation comprises two consecutive attention operations detailed below.

\paragraph{Pair construction.}
The first attention block attends preferentially to the previous state at each position to produce a representation of the current and previous state in the residual stream. The state embeddings for the input sequence lie in a subspace \(B_\text{state}\) of the residual stream, which is guaranteed to have dimension \(\dim{B_\text{state}} \leq C\). Thus, the value matrices may readily rotate the state embeddings of attended positions to an orthogonal subspace when the model dimension \(D\) is large compared to the number of states \(C\). This condition holds for our setting, and we consider the previous states as rotated into an orthogonal subspace \(B_1\). The formation of state-pair representations via the \(B_1\) subspace is depicted in Figure~\ref{fig:Loss}(d, ii), where the strength of the previous-state selectivity is summarized by the scalar \(\delta\). 

\paragraph{Match and readout.}
The second attention block uses the contents of both \(B_\text{state}\) and \(B_1\) to attend such that its output is precisely a vector representation of the empirical conditional distribution. To accomplish this, the attention block matches the current state (in \(B_\text{state}\)) to the previous state (in \(B_1\)) at each position. As a result, the attention weights concentrate on positions where the previous state matches the current state. The value matrix rotates the state in \(B_\text{state}\) to a third subspace \(B_2\), and these new representations are then averaged over the matching positions to form a representation of the empirical conditional distribution in \(B_2\). This representation is of the form \(\sum_{\tau} \hat{P}_n(x_{i+1} = \tau \mid x_i = \mu) \cdot x^\prime_\tau\) where the \(x^\prime_\tau\) are the rotated state embeddings residing in \(B_2\). The operation of the second attention block is depicted in Figure~\ref{fig:Loss}(d, iii), where the strength of the match operation is summarized by the scalar \(\beta\).

In principle, a linear transformation of the empirical conditional distribution is insufficient to implement the 2-Gen predictor, since the predictions must be in the form of log-probabilities. However, an MLP block which follows the two attention blocks may read the contents of \(B_2\) and compute log-probabilities for the next-state prediction according to the 2-Gen prescription.

\section{Minimal network for generalization in the limit $K \to \infty$}
\label{sec:MFT_model}

A direct theoretical analysis of the full circuit in 
equations~\ref{eq:model_start}--\ref{eq:model_end} in Sec.~\ref{sec:transformerinfo} is analytically intractable. To retain the essential computational mechanism revealed by the circuit-tracing results while achieving analytical tractability, we construct a symmetry-constrained, attention-only transformer model 
that preserves the generalization features of the full architecture. We henceforth call this model the \emph{SA-transformer}. The SA-transformer is related to the disentangled transformer \cite{friedman2023learning}. As detailed below, three key assumptions alongside symmetry arguments lead to a substantially reduced description. Although the argument is presented for a two-layer transformer with single-headed attention, the model can be extended naturally to architectures with multiple attention heads per layer and greater depth.

\subsection{Reduction of the network and circuits}
We first introduce simplifications at the architectural level.

\paragraph*{(i) Fixed one-hot embeddings.}
Since sequence states are embedded as orthogonal one-hot vectors, 
we eliminate the embedding matrix $W_E$ and directly work with the one-hot state representations.

\paragraph*{(ii) Disentangled value subspaces.}
We assume that the value matrices $W_V^{(1)}$ and $W_V^{(2)}$ write their respective contributions into distinct, non-overlapping subspaces (either directly, or via MLP1 as seen in the traced \(\pgt\) circuit). 
This separation can be represented explicitly through concatenation. 
Under this assumption, MLP1 becomes redundant and can be removed from the model.

The resulting simplified network is
\begin{align}
x_n^{(0)} &= x_n, \\
y_n^{(1)} &= \mathrm{Att}^{(1)}\!\left(x^{(0)}_{\le n}\right), \\
x_n^{(1)} &= x_n^{(0)} \oplus y_n^{(1)}, \\
y_n^{(2)} &= \mathrm{Att}^{(2)}\!\left(x^{(1)}_{\le n}\right), \\
x_n^{(2)} &= x_n^{(1)} \oplus y_n^{(2)},
\label{eq:gen_model}
\end{align}
where the final prediction is obtained by applying an MLP readout to $x_n^{(2)}$. We further introduce two assumptions at the level of the attention-mechanism.

\paragraph*{(i) Relative positional bias.}
We replace rotary embeddings with a relative positional bias, 
which disentangles positional and content-based attention and facilitates analytical treatment.

\paragraph*{(ii) Value-matrix elimination.}
Under the assumption of non-overlapping value subspaces, 
the value matrices $W_V^{(1)}$ and $W_V^{(2)}$ can be absorbed into the concatenated representation. As a result, they can be removed from \eqref{eq:full_att_func} without altering the effective computation. 

With the above considerations, we can write the output of the network as 
\begin{align}
    \hat{\pi}_{\tau} =  \frac{e^{z_{\tau}}}{\sum_{\tau'} e^{z_{\tau'}}}, \quad z = g\left(x_N \oplus \sum_{i \le N} A_{iN}^{(1)} x_i \oplus \sum_{i \le N} A_{iN}^{(2)} x_i \oplus \sum_{i \le N} \sum_{j\le i} A_{iN}^{(2)}A_{ji}^{(1)} x_j\right),\label{eq:minimal_pre}
\end{align}
with
\begin{align}
A_{ji}^{(1)} = \frac{e^{x_j^TM^{(1)}x_i + P^{(1)}_{j-i}}}{\sum_{j' \le i} e^{x_{j'}^TM^{(1)}x_{i} +P^{(1)}_{j'-i}}}, \quad A_{ji}^{(2)} = \frac{e^{\left(x^{(1)}_j\right)^TM^{(2)}x^{(1)}_i + P^{(2)}_{j-i}}}{\sum_{j' \le i} e^{\left(x^{(1)}_{j'}\right)^TM^{(2)}x^{(1)}_{i} +P^{(2)}_{j'-i}}}. \label{eq:minimal_As}
\end{align}
where we introduce $M^{(\ell)}=(K^TQ)^{(\ell)}$ for convenience, $P^{(l)}_{j-i}$ for relative positional bias and $\oplus$ denotes concatenation. Note that $A^{(2)}$ takes the output of the first layer $x^{(1)}_{1:N}$ as its input. With our simplifications, $x^{(1)}_i = x_i \oplus \sum_{j \le i} A_{ji}^{(1)} x_j$. Since $x^{(1)}_i$ is $2C$-dimensional, the query-key product in the second layer, $M^{(2)}$, has dimensions $2C\times 2C$. The nonlinearity $g$ maps the $4C$-dimensional input to a $C$-dimensional logit vector, which is then passed through a softmax layer to produce the final output. 

\paragraph*{Replacement of MLP2.}
Finally, since the $x_i$ are one-hot vectors, each of the four contributions in \eqref{eq:minimal_pre} are a $C$-dimensional vector with non-negative entries that sum to one. Each term can therefore be interpreted as a probability distribution over state classes, and thus represents a fully normalized predictive distribution for the next state. Under this observation, we can replace the nonlinear MLP readout $g$ by a convex combination of these four distributions. This approximation amounts to restricting $g$ to act linearly on the simplex spanned by the four  contributions, 
thereby preserving normalization and interpretability while substantially reducing parameter complexity. With this approximation, the predicted probability distribution over next states becomes
\begin{align}
\hat{\pi} 
&= 
W_A x_N 
+ W_B \sum_{i \le N} A_{iN}^{(1)} x_i 
+ W_C \sum_{i \le N} A_{iN}^{(2)} x_i 
+ W_D \sum_{i \le N} \sum_{j \le i} A_{iN}^{(2)} A_{ji}^{(1)} x_j ,
\label{eq:minimal_W}
\end{align}
where $W_A, W_B, W_C,$ and $W_D$ are matrices which are restricted such that the output still lies on the probability simplex (these are further simplified below).

\subsection{Symmetry reduction}

In our case, since the generative model for transition matrices $T$ is symmetric over the $C$ state classes, all $C$ state classes are statistically identical. This symmetry implies that the four $W$ matrices will have equal diagonal terms and equal off-diagonal terms. The off-diagonal terms can be set to zero as they contribute a constant offset; for example, if $W_A$ has the form
\begin{align}
    \begin{bmatrix}
w_{\text{d}} & w_{\text{od}} & \cdots & w_{\text{od}} \\
w_{\text{od}} & w_{\text{d}} & \ddots & \vdots \\
\vdots & \ddots & \ddots & w_{\text{od}} \\
w_{\text{od}} & \cdots & w_{\text{od}} & w_{\text{d}}
\end{bmatrix},
\end{align}
then the first term on the right hand side of \eqref{eq:minimal_W} evaluates to $(w_{\text{d}} - w_{\text{od}})x_N + w_{\text{od}}\mathbf{1}$, where $\mathbf{1}$ is a $C$-dimensional vector with all ones. This term can be ignored as it introduces a constant offset. We therefore set $w_{\text{od}}= 0$, without loss in generality. Repeating this argument for $W_B, W_C, W_D$, we observe that the four $W$ matrices can be replaced with non-negative scalar weights $w_A, w_B, w_C, w_D$ that obey $w_A + w_B + w_C + w_D = 1$. In other words, we interpret the MLP $g$ as implementing a mixture-of-experts operation as shown in \eqref{eq:mix_expert_pred}, where the four `experts' correspond to the four terms in \eqref{eq:minimal_W}.

A similar argument also applies to the matrices $M^{(1)}$ and $M^{(2)}$ in the attention head. The exponents involved in the attention weights $A^{(\ell)}$ contain vector-matrix-vector products of the form $x_i M x_j$ for an arbitrary $D$-dimensional matrix $M$. This feature is apparent for $A^{(1)}$ from \eqref{eq:minimal_As}. For $A^{(2)}$, since $u_i = x_i \oplus \sum_{j\le i} A_{ji}^{(1)} x_j$, the exponent will have four vector-matrix-vector products of the form $x_i M x_j$ and our symmetry argument below will apply independently to the corresponding four block matrices in $M^{(2)}$. Now, proceeding with the argument, since the $x_i$ are one-hot vectors with non-zero elements in the first $C$ components, only the top left $C \times C$ submatrix of $M$ will influence this product. This implies that $M^{(1)}$ and the four block matrices in $M^{(2)}$ can be reduced to $C \times C$ matrices. The problem is then entirely independent of $D$. 

Next, as argued for the $W$ matrices, since the diagonal and off-diagonal elements respectively have identical gradients, all diagonal/off-diagonal elements that grow significantly beyond their initial values will be approximately equal. To be concrete, say these values are $\beta_{\text{d}}$ and $\beta_{\text{od}}$. A product of the form $x_i M x_j$ can be written as $\beta_{\text{d}} \delta_{ij} + \beta_{\text{od}}(1-\delta_{ij}) = \left(\beta_{\text{d}} -  \beta_{\text{od}}\right)\delta_{ij} + \beta_{\text{od}}$. The constant term $\beta_{\text{od}}$ appears both in the numerator and denominator of the softmax operation. The off-diagonal terms therefore cancel out. Thus, only the difference between the diagonal and non-diagonal element matters, and any such matrix $M$ can be replaced with a diagonal matrix. 

Combining these arguments leads to an SA-transformer with two layers and one head per layer:
\begin{align}
\hat{\pi} &= w_A x_N + w_B\sum_{i \le N} A_{iN}^{(1)} x_i + w_C\sum_{i \le N} A_{iN}^{(2)} x_i + w_D \sum_{i \le N} \sum_{j\le i} A_{iN}^{(2)}A_{ji}^{(1)} x_j\label{eq:mix_expert_pred}, \text{where}\\
   \log A_{ji}^{(1)} &= \beta^{(1)}_1 \delta_{s_is_j} + P^{(1)}_{j-i} - \log Z^{(1)}, \nonumber \\
   \log A_{ji}^{(2)} &= \beta^{(2)}_1 \delta_{s_is_j} +  \beta^{(2)}_2 \sum_{k \le i} A_{ki}^{(1)} \delta_{s_ks_j} +  \beta^{(2)}_3 \sum_{k \le j} A_{kj}^{(1)} \delta_{s_is_k} +  \beta^{(2)}_4 \sum_{k \le i, k' \le j}  A_{k'j}^{(1)} A_{ki}^{(1)} \delta_{s_ks_{k'}}  + P^{(2)}_{j-i} - \log Z^{(2)}, \nonumber
\end{align}
where $Z^{(1)}, Z^{(2)}$ are normalization constants, $\beta^{(1)}_1$ is the scalar that parameterizes $M^{(1)}$, i.e., $M^{(1)} = \beta^{(1)}_1\mathbf{I}_C$. And $\beta^{(2)}_1, \beta^{(2)}_2, \beta^{(2)}_3, \beta^{(2)}_4$ are four scalars that parameterize the matrix $M^{(2)}$, i.e.,
\be
M^{(2)}=\begin{pmatrix}
M^{(2)}_{11} & M^{(2)}_{12} \\
M^{(2)}_{21} & M^{(2)}_{22}
\end{pmatrix} = \begin{pmatrix}
\beta^{(2)}_1\mathbf{I}_C & \beta^{(2)}_2\mathbf{I}_C \\
\beta^{(2)}_3\mathbf{I}_C & \beta^{(2)}_4\mathbf{I}_C \label{eq:blockM}
\end{pmatrix}.
\ee
where we partition $M^{(2)}$ into 4 $C\times C$ blocks. 
These modifications remove redundancy in the parameter space and substantially simplify the analysis by focusing on the irreducible parameters.

\subsection{Numerical validation}

\subsubsection{Training details}\label{sec:training}
Denote the last \rev{state}  in the sequence $x_N$  as $\mu$. The cross-entropy loss averaged over input sequences given transition matrix $T$ is
\begin{align}
  \mathcal{L} = -\left\langle \sum_{\mu,\tau} p_{\mu} T_{\tau \mu}  \log \pi_{\tau}(x_{1:N}) \right\rangle. \label{eq:loss_markov}
\end{align}
where $T_{\tau \mu}$ is the probability that the next \rev{state} is $\tau$ given that the current \rev{state} is $\mu$ and $p_{\mu}$ is the stationary probability of \rev{state} $\mu$ for transition matrix $T$. Unless otherwise specified, the expectation $\langle . \rangle$ is over all possible sequences that could be drawn given $T$ and the current \rev{state} $\mu$, averaged over transition matrices $T$. For the SA-transformer, we fix the sequence length $N$ during optimization instead of using autoregressive training.

Without loss of generality, we choose the following prediction function to investigate the generalization phase: 
\begin{align}
\hat{\pi} &= w_A x_N + w_B\sum_{i \le N} A_{iN}^{(1)} x_i + w_C\sum_{i \le N} A_{iN}^{(2)} x_i + w_D \sum_{i \le N} \sum_{j\le i} A_{iN}^{(2)}A_{ji}^{(1)} x_j\label{eq:base_model}, \quad \text{where}\\
A_{ji}^{(1)} &= \frac{e^{x_j^TM^{(1)}x_i + P^{(1)}_{j-i}}}{\sum_{j' \le i} e^{x_{j'}^TM^{(1)}x_{i} +P^{(1)}_{j'-i}}}, \quad A_{ji}^{(2)} = \frac{e^{\left(x^{(1)}_j\right)^TM^{(2)}x^{(1)}_i + P^{(2)}_{j-i}}}{\sum_{j' \le i} e^{\left(x^{(1)}_{j'}\right)^TM^{(2)}x^{(1)}_{i} +P^{(2)}_{j'-i}}}. \nonumber
\end{align}

The relative positional bias terms $P^{(\ell)}$ and the elements of the $M^{(\ell)}$ matrices (with $\ell=1, 2$) are initialized at zero. To test our symmetry argument, we \emph{do not} impose the constraint that the $M^{(\ell)}$ matrices have block-identity structure as in \eqref{eq:blockM}. The constraints $w_A,w_B,w_C,w_D \ge 0, w_A + w_B + w_C + w_D = 1$ are imposed by expressing them using a softmax with three learnable parameters. We initialize $w_A = w_B = w_C = w_D = \frac{1}{4}$. The SA-transformer is trained using stochastic gradient descent (SGD) with learning rate 1 and a batch size of 256.

\subsubsection{Training results}
The training loss is shown in Figure~\ref{fig:Loss}b, and shows that the SA-transformer reproduces the abrupt learning of the full network (compare with Figure \ref{fig:Phases}\rev{a} for $K=1024$, where the transformer rapidly learns the 1-Gen solution and subsequently transitions to $\pgt$ after a plateau).

\begin{figure}[h]
\begin{nolinenumbers}
    \centering
\includegraphics[]{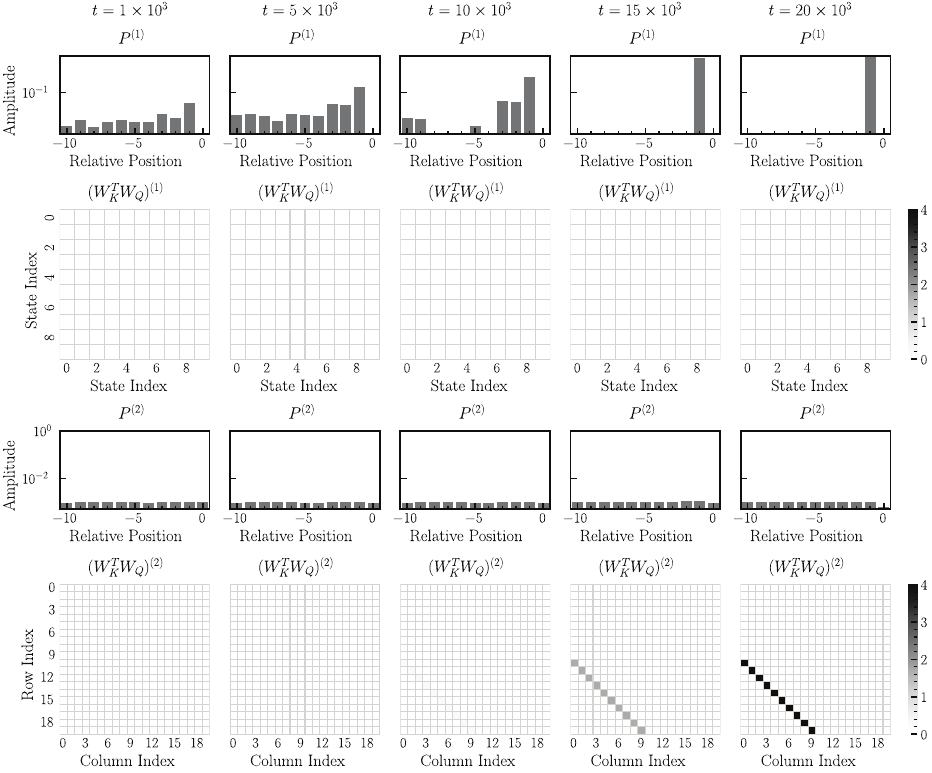}
    \caption{Parameters of the attention circuits defined in \eqref{eq:base_model} at different iterations. The corresponding loss dynamics is given in Figure \ref{fig:Loss} (b). }
    \label{fig:gen_parameters}
\end{nolinenumbers}
\end{figure}
In Figure~\ref{fig:gen_parameters}, we further observe that $M^{(1)}$ in $A^{(1)}_{ij}$, $P^{(2)}$, and the upper-triangular blocks of $M^{(2)}$ in $A^{(2)}_{ij}$ remain unchanged when the parameters are initialized to zero. The equal increase of the diagonal elements of $M^{(2)}_{21}$ also demonstrates that our symmetry-reduction hypothesis holds well. Therefore, we can simplify the model described by \eqref{eq:mix_expert_pred} to
\begin{align}
\hat{\pi} &= w_A x_N + w_B\sum_{i \le N} A_{iN}^{(1)} x_i + w_C\sum_{i \le N} A_{iN}^{(2)} x_i 
+ w_D \sum_{i \le N} \sum_{j\le i} A_{iN}^{(2)}A_{ji}^{(1)} x_j, \quad \text{where}
\label{eq:mix_expert_pred1}\\
\log A_{ji}^{(1)} &= \delta \delta_{j(i-1)}- \log Z^{(1)}, \quad
\log A_{ji}^{(2)} = \beta \sum_{k \le j} A_{kj}^{(1)} \delta_{s_i s_k} - \log Z^{(2)}.
\end{align}
To simplify notation, we define $\delta = P^{(1)}_{-1}$ and $\beta = \beta^{(2)}_3$.
With these reductions, the minimal model is characterized by two parameters, $\beta$ and $\delta$, together with three independent weight parameters $w_B, w_C, w_{D}$.


\section{Dynamics of the SA-transformer}\label{sec:SA_dynamics}
We now examine in detail how the SA-transformer described with \eqref{eq:base_model} can implement the 1-Gen and 2-Gen solutions through the organization of functional circuits across different layers. The analysis is organized as follows: 
\begin{enumerate}
    \item The network first enters $\pgo$ before entering $\pgt$. We show that $w_A \to 0$ and $w_B = w_C = w_D = 1/3$ in $\pgo$, whereas the rest of the parameters remain at zero.
    \item Next, we show that $\pgt$ corresponds to $w_C = 1, \beta \to \infty, \delta \to \infty$ in the SA-transformer. The other terms involving $w_A,w_B,w_D$ do not contribute to the solution after convergence. 
    \item Third, we show that accurately capturing the training dynamics of $\beta, \delta$ requires a careful computation of expectations when expanding the loss in a Taylor series near the 1-Gen solution. While the 2-Gen solution involves a second-order term in the loss of the form $\beta \delta$ (which would imply a saddle-point at the origin $\beta,\delta = 0$), there are subtle first-order contributions that have an important influence on the dynamics of $\beta,\delta$. One of these contributions arises from the term involving $w_B$, even though this term does not matter after convergence. Importantly, these subtle first-order contributions determine the duration of the loss plateau while the network transitions from $\pgo$ to $\pgt$ (Figure~\ref{fig:Loss}b). 
\end{enumerate} 

\subsection{Competition between $x_N$ and the 1-Gen solution}\label{sec:x_N}
\begin{figure}[h]
\begin{nolinenumbers}
    \centering
\includegraphics[width=0.45\linewidth]{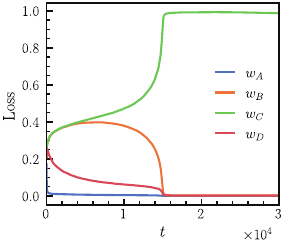}
\caption{Training dynamics of the mixture-of-experts weights in \eqref{eq:base_model}.}\label{fig:gen_weights}
\end{nolinenumbers}
\end{figure}

The network nearly implements the 1-Gen solution at initialization except for a contribution due to the first term involving $w_A$ in \eqref{eq:mix_expert_pred1}. At initialization, we have $A_{ji}^{(\ell)} = 1/i$ for $\ell = 1,2$. Plugging this into \eqref{eq:mix_expert_pred1}, we observe that the terms involving $w_B,w_C,w_D$ compute the 1-point statistics. Since these three terms have identical contributions, we have $w_B = w_C = w_D = (1-w_A)/3$. The term $w_A x_N$ leads to a bias towards repeating the final state class $\mu$. We can write the output of the network as
\be
\hat{\pi}_{\tau} = w_{A} \delta_{\mu\tau} + (1-w_{A})p_\tau.
\ee
The gradient descent dynamics of $w_A$ is
\be
\frac{dw_A}{dt} &= \left< \sum_{\mu,\tau}\left( \frac{p_\mu T_{\tau\mu}}{w_A\delta_{\mu\tau} + (1-w_A)p_\tau}\delta_{\mu\tau}-\frac{p_\mu p_\tau T_{\tau\mu}}{w_A\delta_{\mu\tau} + (1-w_A)p_\tau}\right)\right> \\
&= \left< \sum_{\mu}\frac{p_\mu T_{\mu\mu}(1-p_\mu )}{p_\mu + w_A(1-p_\mu )}-\sum_{\mu, \tau; \mu \neq \tau}\frac{p_\mu T_{\tau\mu}}{(1-w_A)}\right> \equiv \mathcal{F}(w_A) .
\ee
It is obvious that when $w_A\to 1$, the $-\frac{1}{1-w_A}$ term is negative and diverges, driving the decrease of $w_A$.  Since $\mathcal{F}(0)=0$ and $\mathcal{F}(w_A)$  monotonically decreases with $w_A$ when $p_\mu<1$ for $0<w_A<1$,  $\mathcal{F}(w_A)$ is strictly negative throughout his interval $0<w_A<1$. We can estimate the decay time as
\be
s_{d} \approx - \int_0^1\frac{1}{\mathcal{F}(w_A)}dw_A.
\ee
Since $\mathcal{F}(w_A)$ decreases monotonically with $w_A$, we use the dynamics at $w_A\ll 1$ to obtain an upper bound on $s_d$. In this regime,
\be
\frac{dw_A}{dt}&\approx - \left<\sum_\mu \frac{(1-p_\mu)^2T_{\mu\mu}}{p_\mu} +\sum_{\tau\neq \mu }p_\mu T_{\tau\mu}\right>w_A
\\
&= - \left<1+\sum_\mu \frac{(1-p_\mu)^2T_{\mu\mu}}{p_\mu}-p_\mu T_{\mu\mu} \right>w_A\\
&= -\left< 1+\sum_\mu \left(\frac{T_{\mu\mu}}{p_\mu} - 2T_{\mu\mu}\right)\right>w_A\\
&= -\left< -1 + \sum_\mu \frac{T_{\mu\mu}}{p_\mu} \right>w_A
\ee
Typically, $T_{\mu\mu}$ and $p_\mu$ are of the same order, so that $dw_A/dt \approx -C w_A$, i.e., $w_A$ decays rapidly to zero as shown in Figure \ref{fig:gen_weights}. The network thus predicts the 1-point statistics, $\hat{\pi}_{\tau} = p_{\tau}$.

\subsection{The 2-Gen solution after convergence}
Now, we show that 2-Gen can be implemented by setting all other parameters except $w_C, \delta, \beta$ to zero. Specifically, $w_C = 1, \delta \to \infty, \beta \to \infty$ corresponds to the 2-Gen solution. Recall that, by definition, $\delta = P^{(1)}_{-1}$ and $\beta = \beta_3^{(2)}$. When all other parameters are set to zero, from \eqref{eq:mix_expert_pred1}, we have
\begin{align}
    \hat{\pi} &= \sum_{i\le N}A_{iN}^{(2)}x_i, \quad \text{where} \\
    A_{ji}^{(1)} &= \frac{\left(e^{\delta}-1\right)\delta_{j(i-1)} + 1}{i + e^{\delta} - 1}, \quad A_{iN}^{(2)} = \frac{e^{\beta \sum_{j \le i} A_{ji}^{(1)} \delta_{\mu s_j}}}
{\sum_{i' \le N} e^{\beta \sum_{j' \le i'} A_{j'i'}^{(1)} \delta_{\mu s_{j'}}}}.
\end{align}
$\hat{\pi}_{\tau}$ can be written in terms of $\beta$ and $\delta$ as
\begin{align}
    \hat{\pi}_{\tau} = \sum_{i \le N} \frac{ \exp\left(\beta\frac{(e^{\delta}-1)\delta_{s_{i-1}\mu } + k_{\mu i}}{i + e^{\delta} - 1} \right)}{\sum_{i' \le N} \exp\left(\beta\frac{(e^{\delta}-1)\delta_{s_{i'-1}\mu } + k_{\mu i'}}{i' + e^{\delta} - 1} \right)}\delta_{s_i \tau },
\end{align}
where $k_{\mu i} = \sum_{j \le i} \delta_{s_j \mu}$ is the number of occurrences of $\mu$ in the first $i$ states. In the limit $\delta \to \infty$, $A_{ji}^{(1)} = \delta_{j (i-1)}$, i.e., the first attention head pays attention to the previous state, so that
\begin{align}
    \hat{\pi}_{\tau}  =  \sum_{i \le N}\frac{e^{\beta \delta_{s_{i-1}\mu }}}
{\sum_{i' \le N} e^{\beta \delta_{s_{i'-1}\mu }}} \delta_{s_i \tau},
\end{align}
When $\beta \to \infty$, the denominator in the above expression counts the number of occurrences of $\mu$, whereas the numerator counts the number of co-occurrences of $\mu$ and $\tau$ (in that order). Thus, $\hat{\pi}_{\tau} = m_{\tau \mu}/n_{\mu}$, which is the 2-Gen prediction with the regularizing \(\alpha\) term omitted.

Figure~\ref{fig:gen_weights} shows that $w_A$ decays rapidly, since the $w_A$ term corresponds to predicting by simply repeating the last state. The remaining three terms initially yield the one-point prediction. The contribution associated with $w_D$ becomes progressively worse than the one-point prediction, and consequently $w_D$ decays. In contrast, the contribution associated with $w_B$ remains equivalent to the one-point prediction and stays competitive with $w_C$ until the $w_C$ term converges to the two-point solution.
\subsection{Acquisition of the 2-Gen solution}
We now examine the kinetics of acquisition of the 2-Gen solution $w_C = 1, \beta \to \infty, \delta \to \infty$. To do this, we expand the loss to first order in $\beta$ and $\delta$ around the unigram solution, 
\begin{align}
    \mathcal{L}(\beta,\delta) = \mathcal{L}^{\text{1-Gen}} - c_{\beta}\beta - c_{\delta}\delta + \dots. 
\end{align}
Recall from Section \ref{sec:x_N} that the 1-Gen solution corresponds to $w_A = 0, w_B = w_C = w_D = 1/3$ and the rest of the parameters equal to zero. 

\subsubsection{First-order contributions in $\delta$} 
First-order contributions to the loss in $\delta$ appear in the second and fourth terms in \eqref{eq:mix_expert_pred1} ($w_B$ and $w_D$). When computing $\hat{\pi}_{\tau}$, both terms have sums of the form $\sum_{j \le i} A_{ji}^{(1)}\delta_{s_j \tau}$. To first order, this sum evaluates to
\begin{align}
    \sum_{j \le i} A_{ji}^{(1)}\delta_{s_j \tau} = \frac{k_{\tau i}}{i} + \frac{\delta}{i}\left(\delta_{s_{i-1}\tau} - \frac{k_{\tau i}}{i}\right). 
\end{align}
For large $i$, $k_{\tau i}/{i} \to p_{\tau}$. However, the expectation of $\langle \delta_{s_{i-1}\tau} \rangle - p_{\tau}$ over all sequences \emph{given} the transition matrix $T$ and $s_N = \mu$ is not zero. Specifically, $\langle \delta_{s_{i-1}\tau} \rangle = P(s_{i-1} = \tau | s_N = \mu) = (T^{N-i+1})_{\mu \tau}p_{\tau}/p_{\mu}$, where Bayes' rule has been applied to obtain the latter expression. When $N-i$ is large compared to the mixing time of the Markov chain, $(T^{N-i+1})_{\mu \tau} \to p_{\mu}$, and the expression $(T^{N-i+1})_{\mu \tau}p_{\tau}/p_{\mu} - p_{\tau}$ can be neglected. However, this difference cannot be ignored for small values of $N-i+1$. 

\begin{figure}[h]
\begin{nolinenumbers}
\centering
\includegraphics[width=0.45\linewidth]{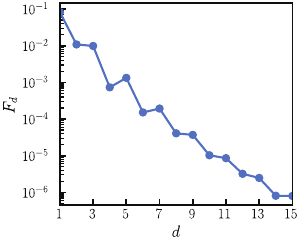}
    \caption{Scaling of $F_d$ with different $d$ for $C= 10, \alpha = 1$.  }
    \label{fig:markov_bias}
\end{nolinenumbers}
\end{figure}

Taking the derivative of the loss w.r.t $\delta$ and setting $A_{iN}^{(2)} = 1/N$ in \eqref{eq:mix_expert_pred1}, we get
\begin{align}
    c_{\delta} &= \left\langle \sum_{\mu, \tau} \frac{p_{\mu}T_{\tau \mu}}{p_{\tau}}\left( \frac{w_B}{N} \left( \frac{T_{\mu \tau}p_{\tau}}{p_{\mu}} - p_{\tau} \right) + \frac{w_D}{N} \sum_{i\le N}\frac{1}{i}\left( \frac{T_{\mu \tau}^{N-i+1}p_{\tau}}{p_{\mu}} - p_{\tau} \right) \right) \right\rangle,\nonumber \\
    &= \left\langle \sum_{\mu, \tau} T_{\tau \mu}\left(\frac{w_B}{N} \left( T_{\mu \tau} - p_{\mu} \right) + \frac{w_D}{N} \sum_{i\le N}\frac{1}{i}\left( T_{\mu \tau}^{N-i+1} - p_{\mu} \right) \right)\right\rangle, \nonumber\\
    &= \left\langle \sum_{\mu} \left(\frac{w_B}{N} \left( T^2_{\mu \mu} - 1 \right) + \frac{w_D}{N} \sum_{i\le N}\frac{1}{i}\left( T^{N-i+2}_{\mu \mu} - 1 \right)\right) \right\rangle, \nonumber\\
    &= \frac{w_B F_1}{N} + \frac{w_D}{N} \sum_{i\le N} \frac{F_{N-i+1}}{i}, \label{eq:c_delta}
\end{align}
where the angled brackets now represent an expectation over the Dirichlet ensemble of transition matrices, and we have defined $F_d = \left< \sum_{\mu}T^{d+1}_{\mu\mu}\right> -1$. In Figure \ref{fig:markov_bias}, we show numerically obtained values of $F_d$ for $C=10, \alpha = 1$. $F_d$ is positive and decays approximately exponentially with $d$, which implies that the second term on the right hand side of \eqref{eq:c_delta} can be ignored when $i$ is closed to 1, while the second term can also be ignored as it becomes $\sim \mathcal{O}(1/N^2)$ when $i$ is closed to $N$. For $d=1$, we obtain an analytical expression:
\begin{align}
    F_1 = \left\langle \sum_{\tau, \mu \ne \tau} T_{\tau \mu} T_{\mu \tau} \right\rangle + \left\langle T_{\tau \tau} T_{\tau \tau} \right\rangle - 1 = \frac{C-1}{C^2\alpha + C}. 
\end{align}
Therefore, we have $c_{\delta} \approx w_B F_1/N$.

\subsubsection{First-order contributions in $\beta$}
First-order contributions to the loss in $\beta$ appear in the third and fourth terms ($w_C$ and $w_D$) in \eqref{eq:mix_expert_pred1}. Both terms involve $A_{iN}^{(2)}$, which to first order in $\beta$ can be written as
\begin{align}
    A_{iN}^{(2)} = \frac{e^{\beta k_{\mu i}/i}}{\sum_{j\le N} e^{\beta k_{\mu j}/j}} \approx \frac{1}{N} + \frac{\beta}{N}\left(\frac{k_{\mu i}}{i} - \frac{1}{N}\sum_{j\le N}\frac{k_{\mu j}}{j}\right).
\end{align}
It is tempting to assume that $k_{\mu i}/i \to p_{\mu}$ for large $i$ and neglect the term in the parenthesis above. However, this term has a non-negligible contribution once the expectation given a transition matrix $T$ and $s_N = \mu$ is computed. In particular, focusing on the expected contribution (given a transition matrix $T$ and $s_N = \mu$) to $\hat{\pi}_{\tau}$ from the $w_C$ term in \eqref{eq:mix_expert_pred1}, we get an expression
\begin{align}
    \frac{k_{\tau N}}{N} + \frac{\beta}{N}\sum_{i\le N} \left\langle \left( \frac{\sum_{j \le i} \delta_{s_{j}\mu}}{i} - p_{\mu} \right)\delta_{s_i \tau} \right\rangle &\approx \frac{k_{\tau N}}{N} +  \frac{\beta}{N} \sum_{i\le N, j\le i} \left( \frac{\langle \delta_{s_{j}\mu}\delta_{s_i \tau} \rangle}{i}  - p_{\mu}p_{\tau} \right) \\
    &\approx \frac{k_{\tau N}}{N} + \frac{\beta}{N} \sum_{i\le N, j\le i} \left( \frac{p_{\mu}T^{i-j}_{\tau \mu}}{i}  - p_{\mu}p_{\tau} \right),
\end{align}
where we have approximated $P(s_i = \tau | s_N = \mu) \approx p_{\tau}$, $P(s_i = \tau, s_j = \mu | s_N = \mu) \approx p_{\mu} T^{i-j}_{\tau \mu}$. When $i \gg j$, $T^{i-j}_{\tau \mu} \approx p_{\tau}$. Approximating $T^{i-j}_{\tau \mu} \approx p_{\tau}$ for $i \ge j + 2$, we get
\begin{align}
    \frac{k_{\tau N}}{N} +  \frac{\beta}{N} \sum_{i\le N, j\le i} \left( \frac{p_{\mu}T^{i-j}_{\tau \mu}}{i}  - p_{\mu}p_{\tau} \right) \approx \frac{k_{\tau N}}{N} +\frac{\beta}{N} H_N \left( p_{\mu} T_{\tau \mu} + p_{\mu} \delta_{\tau \mu}  - 2p_{\mu}p_{\tau} \right),
\end{align}
where the harmonic number $H_N = \sum_{i\le N} 1/i \approx \log N + \gamma$ for large $N$ ($\gamma \approx 0.577$ is the Euler-Mascheroni constant). 

We omit the calculation for the $w_D$ term noting in brief that we obtain an expression of the form
\begin{align}
    \frac{k_{\tau N}}{N} +  \frac{\beta}{N} \sum_{i\le N} \left( \frac{k_{\mu i}}{i} - p_{\mu}\right) \left(\frac{k_{\tau i}}{i}\right). 
\end{align}
The latter expressions self-average, and though the expectation is non-zero, its contribution is small relative to the contribution from the $w_C$ term. 

Taking the derivative of the loss w.r.t $\beta$ and using the above arguments, we get
\begin{align}
    c_{\beta} &\approx w_C\frac{H_N}{N}\left\langle \sum_{\mu, \tau} \frac{p_{\mu}T_{\tau \mu}}{p_{\tau}} \left( p_{\mu}T_{\tau \mu} + p_{\mu} \delta_{\tau \mu} - 2p_{\mu}p_{\tau} \right) \right\rangle, \\
    &\approx w_C\frac{H_N}{N}\left( \left\langle \sum_{\mu} \left( p_{\mu}T_{\mu \mu} - p_{\mu}^2\right)  \right\rangle + \left\langle \sum_{\mu, \tau} \left( p_{\mu}^2T_{\tau \mu}^2/p_{\tau} - p_{\mu}^2 T_{\tau \mu}\right)  \right\rangle\right).
\end{align}
The second expectation above (denoted $I$) is positive for any $T$, i.e., 
\begin{align}
   I=\left\langle \sum_{\mu} p_\mu^{2}\!\left(\sum_{\tau}\frac{(T_{\tau\mu})^{2}}{p_{\tau}}-1\right)\right\rangle \ge 0. 
\end{align}
To show this, we use the Cauchy-Schwarz inequality $(x.y)^2 \le ||x||^2 ||y||^2$ for two vectors $x$ and $y$ to prove  $\sum_{\tau}(T_{\tau \mu})^2/p_{\tau} \ge 1$:
\begin{align}
1 = \left(\sum_{\tau}\sqrt{p_\tau}\,\frac{T_{\tau\mu}}{\sqrt{p_\tau}}\right)^2
\le \left(\sum_{\tau} p_\tau\right)\!\left(\frac{(T_{\tau\mu})^{2}}{p_\tau}\right) = \sum_{\tau}\frac{(T_{\tau\mu})^{2}}{p_\tau}.
\end{align} 
We were unable to show that the first expectation $\left\langle \sum_{\mu} \left( p_{\mu}T_{\mu \mu} - p_{\mu}^2\right)  \right\rangle$ in $c_{\beta}$ is always non-negative, though we find numerically that this term is negligible compared to $I$ for the Dirichlet ensemble. To summarize,
\begin{align}
    c_{\beta} \approx w_C\frac{H_N}{N} I \approx \frac{\log N}{N} I, \quad \text{where } I = \left\langle \sum_{\mu} p_\mu^{2}\!\left(\sum_{\tau}\frac{(T_{\tau\mu})^{2}}{p_{\tau}}-1\right)\right\rangle. 
\end{align}

\subsubsection{Dynamics near $\beta, \delta=0$}
In our numerical experiments (described further below), we perform a perturbation that effectively sets $c_{\delta} = 0$. In this case, $\beta$ increases due to the contribution from $c_{\beta}$ and $\delta$ increases at a slower rate due to its coupling with $\beta$. To derive this scaling, we calculate the loss to the lowest-order coupling term between $\beta$ and $\delta$ (that is, $O(\beta \delta)$). This contribution arises from the $w_C$ term. We omit the calculation here; we find
\begin{align}
    \mathcal{L}(\beta,\delta) = \mathcal{L}^{\text{1-Gen}} - \beta (1 + \delta) w_C \frac{\log N}{N} I - \delta w_B \frac{F_1}{N}.
\end{align}
For small values of $\beta, \delta$, changes in $w_B, w_C$ remain very small. Consequently, $w_B, w_C$ increase gradually during training until $\beta$ and $\delta$ rapidly increase in the nonlinear regime and $w_c$ eventually dominates. 

In summary, the approximate gradient descent dynamics of $\beta$ and $\delta$ in our theory is given by
\begin{align}
\frac{d\beta}{dt} &\approx \frac{\log N}{N}  w_C  I, \quad 
\frac{d\delta}{dt}  
\approx \frac{\log N}{N}  w_C  \beta I + \frac{1}{N}w_BF_1, \nonumber\\ 
I = &\left\langle \sum_{\mu} p_\mu^{2}\!\left(\sum_{\tau}\frac{(T_{\tau\mu})^{2}}{p_{\tau}}-1\right)\right\rangle , \quad F_1 = \frac{C-1}{C^2\alpha + C}, \label{eq:parameter_dynamics_simplest}
\end{align}
and $w_B = w_C = 1/3$ (to lowest order).

\subsection{Testing predictions}\label{sec:Predit_from_minimal_gen_model}

\subsubsection{Loss landscape}
Through above analysis in previous sections, we could express the model as:
\begin{align}
   \hat{\pi}_\tau &= w_A \delta_{\mu\tau} + w_B\sum_{i \le N} A_{iN}^{(1)} \delta_{\tau s_i} + w_C\sum_{i \le N} A_{iN}^{(2)} \delta_{\tau s_i} + w_D \sum_{i \le N} \sum_{j\le i} A_{iN}^{(2)}A_{ji}^{(1)}\delta_{\tau s_j}, \quad \text{where} \label{eq:minimal_simple} \\ 
   A_{ji}^{(1)} &=\frac{\left(e^{\delta}-1\right)\delta_{j(i-1)} + 1}{i + e^{\delta} - 1} \quad \text{and} \quad
  A_{ji}^{(2)} = \frac{\exp\left(\beta \sum_{k \le j} A_{kj}^{(1)} \delta_{s_is_k}\right)}{
\sum_{j'}\exp\left(\beta \sum_{k' \le j'} A_{k'j'}^{(1)} \delta_{s_is_{k'}}\right)}.
\end{align}
Thus the loss landscape can be parameterized by five variables. In order to reduce it into the two parameters, $\delta$ and $\beta$, we assume that the MLP rapidly optimize over the weights of four components. Therefore, for each fixed pair $(\delta, \beta)$, the weights $w_{A\text{–}D}$ are determined by minimizing the loss. The resulting loss landscape predicted by our minimal model is shown in  Figure \ref{fig:Landscape}(a).  The flat region at the origin corresponds to the first plateau in the training loss curve in Figure \ref{fig:Loss}(b). Our analytical model predicts that, when $\beta, \delta \sim 1$, there is a sharp cliff in the landscape that triggers abrupt learning. After the rapid transition, it reaches another flat region that corresponds to the second, final plateau in the training loss curve.

\subsection{Ablating the first-order contribution in $\delta$}
In \eqref{eq:parameter_dynamics_simplest}, the dynamics of $\delta$ is governed by two terms of which only the first depends on $\beta$. If we set $F_1 = 0$, the dynamics change qualitatively. $F_1$ is non-zero due to subtle correlations between $s_{N-1}$ and $s_{N+1}$. We remove these correlations by resampling the state at position $N-1$ after generating the sequence, thereby ablating the first order term in $\delta$. Resampling $s_{N-1}$ breaks the short-range correlation $P(s_{N-1} = \tau, s_{N+1} = \tau | s_N = \mu)$ while preserving long-range statistics. 

\begin{figure}[h]
\begin{nolinenumbers}
\centering
\includegraphics[width=0.65\linewidth]{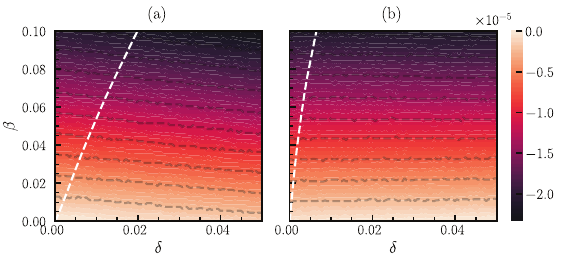}
    \caption{Loss landscapes near $\delta,\beta \to 0$ for (a) original data and (b)  data with the second-to-last state bias removed. The white dashed line is the corresponding dynamics of $\beta$ and $\delta$.  }
    \label{fig:gen_loss_near_origin}
\end{nolinenumbers}
\end{figure}

Figure \ref{fig:gen_loss_near_origin}a and \ref{fig:gen_loss_near_origin}b compares the loss landscape near the origin ($\beta, \delta \sim 0$) for the original and ablated cases, with optimal $w_{A-D}$ for given $\beta, \delta$. Our theory predicts the following behavior: in the original scenario, the loss decreases along both $\beta$ and $\delta$. In contrast, after ablating the first order contribution in $\delta$, the landscape remains flat when $\beta = 0$. The loss decreases appreciably along the positive $\delta$ direction only once $\beta$ becomes non-negligible. This accounts for the scaling of $\delta$ observed in the minimal neural network (Figure \ref{fig:Landscape} b), where $\delta$ grows on average but still fluctuates until $\beta$ becomes non-negligible.

Moreover, we can derive scaling relations directly from \eqref{eq:parameter_dynamics_simplest}. In the original scenario, the $\beta$ term in the $\delta$ dynamics can be neglected. In this case, we expect $\beta \sim t, \delta \sim t$. In contrast, when the first-order contribution in $\delta$ is ablated, we expect $\beta \sim t, \delta \sim t^2$. Thus, while $\beta$ scales linearly with time in both cases, $\delta$ grows linearly in the original scenario but quadratically in the perturbed scenario. This distinct scaling behavior is consistent with the dynamics observed in numerical simulations using the SA-transformer described by equation \ref{eq:base_model} (see Figure \ref{fig:Landscape}b).

\subsection{The time to transition from the 1-Gen to the 2-Gen solution}
Finally, a key result of our theory is an estimate for the number of iterations required to transition from the 1-Gen to the 2-Gen solution, which we call $\tau_{\text{2-Gen}}$. From \eqref{eq:parameter_dynamics_simplest}, $\beta$ grows at a constant rate until it reaches the nonlinear regime, after which it increases rapidly and saturates shortly thereafter. Since $\beta$ grows more rapidly than $\delta$, the dynamics of $\beta$ predominantly determine $\tau_{\text{2-Gen}}$. This leads to the scaling law,
\be\label{eq:2-Gen_time}
\tau_{\text{2-Gen}} \sim \frac{N}{\log N},
\ee
which is in excellent agreement with the behavior of the minimal network described by equation \ref{eq:base_model} and the full transformer, as shown in Figure \ref{fig:Landscape}c.

\subsection{Beyond the SA-Transformer}
\label{sec:beyond_sa}
\begin{revblock}
We next examine whether the predictions derived for the
SA-transformer extend to a less constrained architecture.
We restore learned state embeddings, query, key, and value
matrices, and a final nonlinear MLP readout, while retaining
the separation of the original state representation and
the first-layer output into distinct subspaces.
We compare relative positional bias (RPB) and rotary
positional embedding (RoPE)\cite{su2024roformer}, fixed-context and autoregressive
(AR) training, and SGD and AdamW optimization.
These comparisons test both the context-length dependence
of the transition time and the drift dynamics underlying
the transition.

\subsubsection{Architecture}
Let $x_n\in\mathbb{R}^{C}$ denote the one-hot representation
of the state at position $n$.
We define the embedding matrix as
\begin{equation}
    W_E=\frac{1}{\sqrt{D}}E
    \in\mathbb{R}^{D\times C},
    \label{eq:general_embedding}
\end{equation}
where $E$ is trainable and its entries are initialized
independently from $\mathcal{N}(0,1)$.
Generalizing \eqref{eq:gen_model}, the network is
\begin{align}
    x_n^{(0)}
    &=W_E x_n,
    \nonumber\\
    y_n^{(1)}
    &=\mathrm{Att}^{(1)}
      \!\left(x_{\leq n}^{(0)}\right),
    \nonumber\\
    x_n^{(1)}
    &=x_n^{(0)}\oplus y_n^{(1)},
    \nonumber\\
    y_n^{(2)}
    &=\mathrm{Att}^{(2)}
      \!\left(x_{\leq n}^{(1)}\right),
    \nonumber\\
    x_n^{(2)}
    &=x_n^{(1)}\oplus y_n^{(2)},
    \nonumber\\
    \hat{\pi}_n
    &=\operatorname{softmax}
      \!\left[\varphi\!\left(x_n^{(2)}\right)\right].
    \label{eq:general_model}
\end{align}
Here $\oplus$ denotes concatenation, so that
$x_n^{(0)}\in\mathbb{R}^{D}$,
$x_n^{(1)}\in\mathbb{R}^{2D}$, and
$x_n^{(2)}\in\mathbb{R}^{4D}$.
The readout $\varphi:\mathbb{R}^{4D}\to\mathbb{R}^{C}$
is an MLP with two hidden layers of widths $512$ and $256$,
GELU activations, and a final linear map to the logits.
The architecture applies no layer normalization and has
no MLP between the two attention layers.

Both attention layers are single-head.
Let $h_i\in\mathbb{R}^{D_\ell}$ denote the input to layer
$\ell$ at position $i$, with $h_i=x_i^{(0)}$ for $\ell=1$
and $h_i=x_i^{(1)}$ for $\ell=2$.
The query, key, and value vectors are obtained through
learned linear maps without biases:
\begin{equation}
    q_i^{(\ell)}=W_Q^{(\ell)}h_i,
    \qquad
    k_i^{(\ell)}=W_K^{(\ell)}h_i,
    \qquad
    v_i^{(\ell)}=W_V^{(\ell)}h_i.
    \label{eq:general_qkv}
\end{equation}
Following Sec.~\ref{sec:transformerinfo}, the first index
of $A_{ij}^{(\ell)}$ denotes the query position $i$ and
the second denotes the attended (key) position $j$.
The attention weights and output are
\begin{align}
   \mathrm{Att}^{(\ell)}(h_{\leq i})
    =
    \sum_{j\leq i}A_{ij}^{(\ell)}v_j^{(\ell)}, \quad \text{with } A_{ij}^{(\ell)}
    =
    \frac{\exp\!\left(s_{ij}^{(\ell)}\right)}
         {\sum_{j'\leq i}
          \exp\!\left(s_{ij'}^{(\ell)}\right)}.
    \label{eq:general_attention}
\end{align}
The matrix dimensions are
\begin{equation}
\begin{gathered}
    W_Q^{(\ell)},\,W_K^{(\ell)}
        \in\mathbb{R}^{d_\ell\times D_\ell},
    \qquad
    W_V^{(\ell)}
        \in\mathbb{R}^{D_\ell\times D_\ell},
\end{gathered}
    \label{eq:general_matrix_dimensions}
\end{equation}
where $D_2 = 2D_1=2D, d_2 = 2d_1=d$,with $D=d=64$ in our experiments.

For RPB, the attention score is
\begin{equation}
    s_{ij}^{(\ell)}
    =
    \left(q_i^{(\ell)}\right)^{\mathsf T}k_j^{(\ell)}
    +P_{j-i}^{(\ell)},
    \label{eq:general_rpb}
\end{equation}
where $P_r^{(\ell)}$ is a learned bias indexed by the signed
displacement $r=j-i$ and initialized to zero.
The previous-state offset therefore corresponds to $r=-1$.
Keeping the definition
$M^{(\ell)}=(W_K^{(\ell)})^{\mathsf T}W_Q^{(\ell)}$
from \eqref{eq:minimal_As}, the attention score
can equivalently be written as
$h_j^{\mathsf T}M^{(\ell)}h_i$.

For RoPE \cite{su2024roformer}, position-dependent rotations replace the
additive positional bias:
\begin{align}
    s_{ij}^{(\ell)}
    &=
    \left(R_i^{(\ell)}q_i^{(\ell)}\right)^{\mathsf T}
    \left(R_j^{(\ell)}k_j^{(\ell)}\right)
    \nonumber\\
    &=
    \left(q_i^{(\ell)}\right)^{\mathsf T}
    R_{j-i}^{(\ell)}k_j^{(\ell)}.
    \label{eq:general_rope}
\end{align}
Here $R_i^{(\ell)}\in\mathbb{R}^{d_\ell\times d_\ell}$
is block diagonal, with each $2\times2$ block rotating
each pair of elements by the angle
$i\omega_a^{(\ell)}$, where
\begin{equation}
    \omega_a^{(\ell)}=10000^{-2a/d_\ell},
    \qquad
    a=0,\ldots,\frac{d_\ell}{2}-1.
    \label{eq:general_rope_frequencies}
\end{equation}
The second equality in \eqref{eq:general_rope} follows
from
$(R_i^{(\ell)})^{\mathsf T}R_j^{(\ell)}
=R_{j-i}^{(\ell)}$.
Thus, the explicit positional dependence enters through
the relative displacement $j-i$.
Rotations are applied to queries and keys in both layers,
while the value vectors remain unrotated.
The frequency sets are determined separately for each layer
through $d_\ell$.

At initialization, the attention matrices have independent
Gaussian entries:
\begin{equation}
    W_\nu^{(\ell)}
    \overset{\mathrm{iid}}{\sim}
    \mathcal{N}\!\left(0,\frac{\sigma^2}{2D_\ell}\right),
    \qquad \nu\in\{Q,K,V\}.
    \label{eq:general_attention_initialization}
\end{equation}
The parameter $\sigma$ controls the initialization scale
of the attention matrices in both layers.

\begin{figure}[h]
\begin{nolinenumbers}
    \centering
    \includegraphics[width=0.85\linewidth]{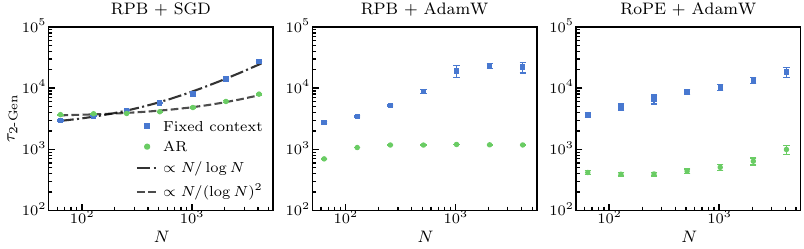}
    \caption{
    \rev{
    Context-length dependence of the transition time beyond the
SA-transformer. The network has two attention layers with learned
embeddings, query, key, and value matrices, and a final MLP readout.
Left: relative positional bias (RPB) with SGD.
Middle: RPB with AdamW.
Right: rotary positional embedding (RoPE) with AdamW.
Blue squares and green circles denote fixed-context and autoregressive
(AR) training, respectively.
Points show the mean transition time $\tau_{\mathrm{2\text{-}Gen}}$
over three random seeds; error bars indicate the standard error of
the mean.
The dash-dotted and dashed curves in the left panel are fits of
the form $\tau_0+ a f(N)$, with
$f(N)=N/\log N$ and $f(N)=N/(\log N)^2$, respectively.
We use batch size $B=64$ throughout.
Learning rates are $0.5$ for SGD and $10^{-3}$ for AdamW.
}
}
    \label{fig:scaling_N_general}
\end{nolinenumbers}
\end{figure}

\subsubsection{Context-length dependence of the transition time}

\paragraph*{The invariant scaling of $\beta$ under the new architecture with RPB and SGD optimizer}
In the architecture of \eqref{eq:general_model},
the order parameter $\beta$ is represented by
\begin{equation}
   \beta=\frac{1}{C}\operatorname{tr}G, \text{ with } G =
    W_E^{\mathsf T}
    \left(W_V^{(1)}\right)^{\mathsf T}
    M_{21}^{(2)}W_E.
    \label{eq:beyond_sa_beta}
\end{equation}
where $M^{(2)}=(W_K^{(2)})^{\mathsf T}W_Q^{(2)}$. Then we want to prove that 
the $N$ scaling relation in $\beta$ dynamics shall not change as a product of
trainable matrices. 

Let $\mathcal W$ denote all parameters entering
\eqref{eq:beyond_sa_beta} and $W\in\mathcal W$    indicates each weight matrix.
Retaining the matching contribution to their
gradient-flow dynamics gives
\begin{align}
    \frac{dW_{ab}}{dt}
    &\propto - \frac{\partial \mathcal{L}}{\partial\beta}\nabla_{W_{ab}}\beta,
    \qquad W\in\mathcal W.
    \label{eq:general_factor_dynamics}
\end{align}
Then 
\begin{equation} \label{eq:general_beta_drift}
    \frac{d\beta}{dt}
=
\sum_{W\in\mathcal W}\sum_{a,b}
\frac{\partial\beta}{\partial W_{ab}} \frac{dW_{ab}}{dt}\propto - \sum_{W\in\mathcal W}\sum_{a,b}\frac{\partial \mathcal{L}}{\partial\beta}(\nabla_{W_{ab}}\beta)^2.
\end{equation}
The response coefficient $\nabla_{W_{ab}}\beta$ depends
on the matrix factors and their initialization, but
has no explicit $N$-dependence. So we expect that our predictions from SA-transformer still hold for the larger architecture under the same setting (relative positional bias, SGD, fixed-context training), which is verified in Fig.~\ref{fig:scaling_N_general}, left panel. 

\paragraph*{Effects of autoregressive training}
The transition time $\tau_{2-Gen}$ for fixed-context training has been given in \eqref{eq:2-Gen_time}. 
AR training averages the loss over different positions.
Applying the fixed-context matching bias in
\eqref{eq:sa_cbeta} at each length gives
\begin{align}
    \frac{d\beta}{dt}
    \propto
    \frac{1}{N}\sum_{n=1}^{N}\frac{H_n}{n}
    \approx  \frac{1}{N}\int_1^N\frac{\log n}{n} dn = \frac{(\log N)^2}{N}.
    \label{eq:beyond_sa_ar_drift}
\end{align}
The corresponding transition-time scale is given by
\begin{equation}
    \tau_{\mathrm{2\text{-}Gen}}^{\mathrm{AR}}
    \propto  \frac{N}{(\log N)^2}.
    \label{eq:beyond_sa_tau_ar}
\end{equation}
The RPB architecture autoregressively trained with SGD exhibits trends
consistent 
with \eqref{eq:beyond_sa_tau_ar} as shown in Fig.~\ref{fig:scaling_N_general}, left panel.

\paragraph*{Effects of AdamW optimizer}
AdamW normalizes the first gradient moment by the square
root of the second moment.
For AR training, the matching gradient scales as
$(\log N)^2/N$.
If this factor is the only dependence of the gradient
history on $N$, it also multiplies the first moment,
while its square multiplies the second moment.
Neglecting the numerical stabilizer $\varepsilon$, these $N$-dependent factors cancel.
Adaptive normalization can therefore remove the
context-length dependence arising from the gradient
amplitude.
Consistent with this mechanism, AR training with RPB
and AdamW exhibits an approximately constant transition
time  as shown in Fig.~\ref{fig:scaling_N_general} in the middle panel. The fixed-context case still exhibits an N-dependence due to the finite sampling noise introduced by a small batch size of 64. 

\paragraph*{Effects of RoPE}
Our predicted scaling fails for RoPE (see Fig.~\ref{fig:scaling_N_general}, right panel). Because RoPE tunes weights underlying different frequency channels directly within the embedding space (see \eqref{eq:general_rope}) —rather than operating in a separate vector space like RPB (see \eqref{eq:general_rpb})—it introduces non-trivial dynamics not captured by our present theory.

\subsubsection{Initialization-dependent drift dynamics}
\begin{figure}[h]
\begin{nolinenumbers}
    \centering
    \includegraphics[width=0.55\linewidth]{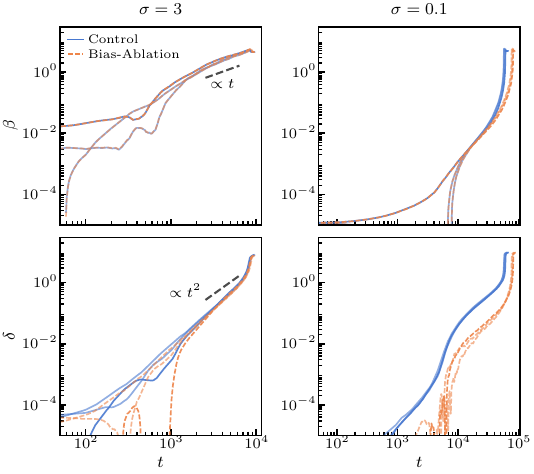}
    \caption{
    \rev{
    Initialization-dependent drift dynamics beyond the SA-transformer. The two-layer attention architecture is the same as in Fig. \ref{fig:scaling_N_general} but trained only with RPB and fixed-context SGD at
$N=1024$, batch size $B=256$, and learning rate $0.5$.
Columns correspond to attention-matrix and embedding matrix initialization scales
$\sigma=3$ (left) and $\sigma=0.1$ (right).
Rows show the effective matching parameter $\beta$ (top) and
the previous-state positional bias $\delta$
(bottom).
Blue solid curves denote control training; orange dashed curves
denote bias ablation.
Each condition contains three trajectories from distinct
initializations, paired between control and ablation.}
}
    \label{fig:scaling_drift_general}
\end{nolinenumbers}
\end{figure}

Equation~\ref{eq:beyond_sa_beta} expresses $\beta$ as a product
of learned matrices.
Its learning dynamics therefore depend on the
initialization of these constituent factors, as
illustrated by Saxe et al.~\cite{saxe2013exact}
for deep linear networks.
Intuitively, larger matrix factors can increase the
effective response coefficient in
\eqref{eq:general_beta_drift}, amplifying the change
in $\beta$ driven by a given loss gradient.
The attention initialization scale $\sigma$ can
therefore change the growth rate of $\beta$.

In the large-$\sigma$ regime, $\beta$ grows substantially
faster while retaining approximately linear growth,
$\beta\propto t$.
The term linear in $\beta$ in the $\delta$ dynamics
of \eqref{eq:parameter_dynamics_simplest} can therefore
no longer be neglected, producing a quadratic
contribution to $\delta$ in both control and
bias-ablation conditions.
The left two panels of
Fig.~\ref{fig:scaling_drift_general} show qualitative
agreement with these predictions.

In the small-$\sigma$ regime, nonlinear effects become
important, and the dynamics of both $\beta$ and $\delta$
depart from the simple growth laws obtained from
\eqref{eq:parameter_dynamics_simplest} and become nontrivial as shown in the right two panels
of Fig.~\ref{fig:scaling_drift_general}.
\end{revblock}

\section{The memorization circuit}\label{sec:Memorization}
\subsection{Task vector intuition}

In the 2-Mem circuit, MLP2 primarily reads from two inputs to produce the logit: MLP1 and Att2, the latter of which averages the outputs of MLP1 over the sequence. Consider the fluctuations of these inputs over sequences sampled from the same chain. MLP1 can only read the current state and the output of Att1, the latter of which is almost exclusively a representation of the previous state. Thus, the input from MLP1  depends only on local sequence information and is expected to exhibit large fluctuations across sequence positions and sampled sequences, even those generated by the same chain. In contrast, Att2 averages the local representations produced by MLP1, and we expect the large-\(N\) average to concentrate on a vector determined by the stationary 2-point distribution. Thus, the input from Att2 will exhibit small fluctuations across sequences from the same chain. With this intuition in mind, we propose that MLP2 infers the task identity using only the aggregated sequence information produced by Att2, and refer to that output on a sequence as a task vector \(\varphi\). The remaining input from MLP1 provides MLP2 with the current state to condition the next-state predictions on.

\subsection{Representation geometry}
Since MLP2 infers the current task condition from $\varphi$, it is desirable for instantiations of $\varphi$ to be separable when the underlying sequences are from different tasks. We computed t-SNE for $\varphi$ and were able to observe task-specific clustering forming through training on data diversities up to $K=128$. The task vectors for this computation are the cached outputs of Att2 at the last state across many sequences.

\subsection{Task vector patching}\label{sec:tv_patch}
We propose that MLP2 makes predictions using only the task information provided by $\varphi$, and test this by analyzing the transformer behavior when the $\varphi$ task information is incongruent with any other alternative source of task information. The transformer first performed a forward pass on a batch of sequences $\mathcal S_A$ (Condition $S_A$). During the forward pass, the output of Att2 (i.e. the task vectors for each sequence and sequence position) was cached. The model then performed a second forward pass on a new batch of sequences $\mathcal S_B$ with the output of Att2 overwritten by the outputs observed on sequence $\mathcal S_A$ (Condition $S_B $ (Patch A)). Consider two predictors for comparison with the model predictions. One is the 2-Mem predictor that makes predictions on $\mathcal S_A$ directly. The other is a 2-Mem predictor which computes the posterior distribution over transition matrices given the context in $\mathcal S_A$ but predicts the next-state distribution given the current state in $\mathcal S_B$. Denote the predictors first by the sequences used to compute the posterior and second by the sequences given as the current state; thus, the two predictors outlined are AA and AB respectively. We compared the model behavior to each predictor by averaging the divergence \(D_\text{KL}^{S_N}(\hat\pi\mid\mid\hat\pi_\theta)\) (see Section \ref{sec:behave_readouts}) for each sequence in the batch, and converted to a similarity by normalizing this quantity by the divergence to a predictor that is uniform over the states and applying a negative exponential. We repeated the same procedure but with the roles of $\mathcal S_B$ and $\mathcal S_A$ swapped, in this case comparing to BB and BA predictors.

The similarity of a \(K=128\) transformer predictions in each condition to each predictor is shown in Figure \ref{fig:task_vector}\rev{b}. In condition \(\mathcal S_A\) and \(\mathcal S_B\) the transformer is similar to predictors AA and BB respectively, as expected. In the Patch A (Patch B) conditions, the transformer is similar to predictor BA (AB). Thus, MLP2 algorithmically behaves like a predictor that infers the posterior over tasks from the Att2 task vector $\varphi$ and predicts the next-state distribution accordingly.

\begin{figure}[h]
\begin{nolinenumbers}
\centering
\includegraphics[]{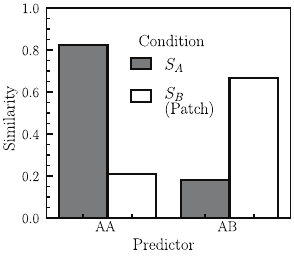}
    \caption{The task vector patching experiment performed for \(\pmo\) by replacing the output of Att1 and comparing to the 1-Mem predictor.}
    \label{fig:1mem_patching}
\end{nolinenumbers}
\end{figure}

We conduct the same patching experiment for \(\pmo\), but replace the output of Att1 and compare to the 1-Mem predictor instead. The similarities in each condition are shown in Figure~\ref{fig:1mem_patching} and indicate that the blocks computing the output prediction (Figure~\ref{fig:add_circuits}b) identify the transition matrix from the task vector alone. This task vector is likely localized to the vector passed from Att1 to MLP1 in the traced circuit.

\subsection{Identifying encoded information}
\begin{revblock}
The task vector patching experiments (Appendix~\ref{sec:tv_patch}) establish that the next-state prediction is made given the posterior task distribution inferred from the task vector $\varphi$ alone. However, this alone does not specify the information that is compressed into $\varphi$ and used to infer the posterior. Specifically, $\varphi$ could constitute either 1) a compressed representation of the posterior task distribution itself or 2) a compressed representation of the 2-point sequence statistics.

To differentiate between these two possibilities, we measured the ability for either 1) the true task identity or 2) the 2-point sequence statistics to be linearly decoded from the $\varphi$ observed in a $\pmt$ transformer. The task identity decoder $W_1 \in \mathbb{R}^{D \times K}$ was trained to minimize the cross-entropy classification loss for the latent chain identity on in-distribution sequences, averaged over sequence positions. The sequence statistics decoder $W_2 \in \mathbb{R}^{D \times C^2}$ was trained to minimize the KL-divergence from the sequence statistics on out-of-distribution sequences, restricted to observed transitions and averaged over sequence positions. We denote the final performance of each decoder for a given $K$ by $\mathcal{L}^\text{model}_1$ and $\mathcal{L}^\text{model}_2$.

To establish baseline performance, we also trained decoders on task vectors produced by jointly optimized sequence embeddings. This baseline implementation follows the minimal model for 2-point memorization (Appendix~\ref{app:minimal_network_mem}), with the following modifications: the 2-point embedding function (Equation~\ref{app:minembed}) is implemented as a learned lookup table rather than an MLP, the initial state is embedded by a separate lookup table, we fix $D_\varphi = D$, and the MLP predictor is replaced by a linear decoder. Since the decoder and task vector are jointly optimized for either the task identification or sequence statistic objective, we consider the performance of these models as optimal for the corresponding objective, given the constraints of the minimal memorization model (which follow from the observed $\pmt$ circuit, see Appendix~\ref{app:minimal_network_mem} for details). We denote the final performance of each baseline decoder by $\mathcal{L}^\text{optimal}_1$ and $\mathcal{L}^\text{optimal}_2$.

The objective of encoding either the task posterior or sequence statistics are similar, so optimizing for one may result in a task vector that contains a significant amount of information about the other. To account for this, we further trained cross-objective decoders which receive frozen task vectors produced by the final baseline model for the opposite objective. We denote the final performance of each cross-objective decoder by $\mathcal{L}^\text{cross}_1$ and $\mathcal{L}^\text{cross}_2$, where the subscript indicates the decoding objective.

The above decoder models were trained for $20$k iterations of minibatch gradient descent with the Adam optimizer and batch size of 128 sequences. Models at the same $K$ share the same training task distribution. Transformer models were trained for $10^5$ iterations following the procedure outlined in Appendix~\ref{sec:training_AR}.

 Figure~\ref{fig:lin_decoders}, shows a comparison of the rescaled relative loss
\begin{equation}
    \tilde{\mathcal{L}}_i \equiv \frac{\mathcal{L}^\text{model}_i - \mathcal{L}^\text{baseline}_i}{\mathcal{L}^\text{cross}_i - \mathcal{L}^\text{baseline}_i}
\end{equation}
for each objective across a range of $D$ and $K$. By construction, $\tilde{\mathcal{L}}_i < 1$ indicates that the transformer task vector contains more information for the corresponding objective than expected under optimization for the opposite objective.  Across the considered $D$ and $K$ values, $\tilde{\mathcal{L}}_2 > 1$ and the transformer is not observed to be optimizing the task vector to encode the 2-point sequence statistics. Instead, $\tilde{\mathcal{L}}_1 < 1$ for sufficiently small $K$ for each $D$. Thus, we conclude that the $\pmt$ task vector is constructed to represent the posterior task distribution.

\begin{figure}[h]
\begin{nolinenumbers}
\centering
\includegraphics[]{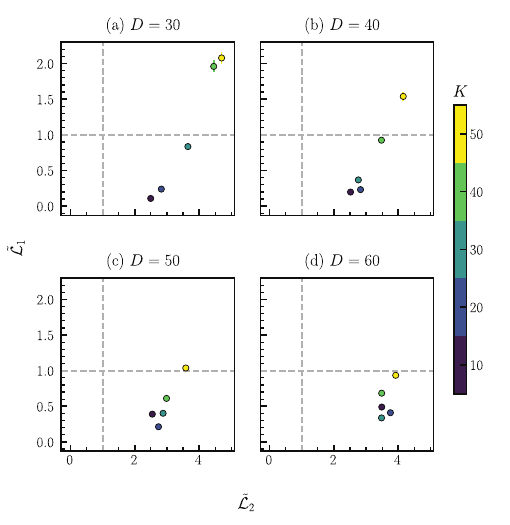}
    \caption{\rev{Comparison of linear decoding performance from the $\pmt$ task vector of task identity ($\tilde{\mathcal{L}}_1$) and 2-point sequence statistics ($\tilde{\mathcal{L}}_2$), varying $D$ and $K$. Values of $\tilde{\mathcal{L}} < 1$ indicate significantly strong decoding of the corresponding information; dashed lines indicate this threshold. Models only ever show significant encoding of task identity. Errorbars correspond to paired-bootstrapped 95\% confidence intervals  from evaluations on $32 \times K$ and $2^{13}$ sequences for $\tilde{\mathcal{L}}_1$ and $\tilde{\mathcal{L}}_2$ respectively.}}
    \label{fig:lin_decoders}
\end{nolinenumbers}
\end{figure}

\end{revblock}

\subsection{Measuring task information content}\label{sec:tv_info}
In \(\pmt\), the first layer attention weight concentrates on the previous state $A^{(1)}_{ni}\approx\delta(n-i-1)$, in which case Att1 may be seen as forming a representation of 2-point statistics in the residual stream $x^{(1)}_n = x_n + W_V^{(1)}x_{n-1}\equiv f(x_n, x_{n-1})$. The traced circuit indicates that MLP1 reads these pairs and produces an output vector for each that is then aggregated across the sequence by Att2 to form the task vector \(\varphi\). Indeed, MLP1 is a requirement for the 2-Mem circuit because of a general limitation of linear 2-point embeddings, which the attention operation is restricted to perform. For any linear embedding function $f_{\text{lin}}$, averaging the embeddings along the sequence \(\expval{f_{\text{lin}}(x_i, x_{i-1})}= f_{\text{lin}}(\expval{x_i}, \expval{x_{i-1}})\) destroys the bigram information. As a result, the second attention layer must average a nonlinear bigram embedding computed by MLP1.

In what follows, we consider MLP1 to be equivalent to a non-linear embedding function $\lambda(s_i, s_{i-1})$ of neighboring pairs of states $(s_i, s_{i-1})$ and approximate the task vector $\varphi_n$ at position $n$ as an average of these embeddings
\begin{align}\label{tvdecomp}
    \varphi_n \approx \frac{1}{n}\sum_{i\leq n} \lambda(s_i, s_{i-1}).
\end{align}
The embedding function acts on state pairs \((\tau, \mu) \in C \times C\) and forms an embedding matrix $\mathcal{E}\in\mathbb R^{D\times C^2}$. Additionally, each chain $k$ defines a distribution over the pairs \(P^{(k)}(\mu, \tau) = p_\mu^{(k)} T^{(k)}_{\tau\mu}\). Write the vector of 2-point probabilities for chain \(k\) as $p^{(k)}\in\Delta^{C^2-1}$.

We would like to analyze the optimality of a given set of embeddings. Specifically, we estimate the mutual information between the task vector as in equation \ref{tvdecomp} and the chain identity \(I(K; \varphi) = H(\varphi) - H(\varphi \mid K)\). Substituting the uniform prior over chains \(p(k) = \frac{1}{K}\) into each term of this expression gives
\begin{gather}
	H(\varphi) = \log K - \frac{1}{K} \int d^D\varphi \, \left( \sum_k p(\varphi \mid k) \right) \log \sum_k p(\varphi \mid k) \label{eq:tv_info_first} \\
	H(\varphi \mid K) = -\frac{1}{K} \int d^D\varphi \, \left( \sum_k p(\varphi \mid k) \log p(\varphi \mid k) \right) \label{eq:tv_info_last}
\end{gather}                       
and the distribution $p(\varphi\mid k)$ remains to be determined. To do so, note that the task vector resulting from a length \(N+1\) sequence may be written as a sum over transitions
\begin{equation}
    \varphi= \frac{1}{N}\mathcal{E} \boldsymbol{n},
\end{equation}
where $\boldsymbol{n} \in \mathbb N^{C^2}$ is the vector of counts for state pairs \((\tau, \mu)\) in the sequence.  Fixing a chain \(k\), the vector of counts \(\boldsymbol{n}^{(k)}\) is multinomially distributed across different sequences with pair probabilities \(p^{(k)}\). As a result, the task vector is a random variable \(\varphi^{(k)} \sim p(\varphi \mid k)\) with mean and covariance
\begin{align}
    \mathbb{E} (\varphi^{(k)}) &= \frac{1}{N}\mathcal{E}\, \mathbb{E}(\boldsymbol{n}^{(k)}) = \mathcal{E} p^k, \\
    \text{Cov}(\varphi^{(k)}) &= \frac{1}{N^2}\mathcal{E}\,\text{Cov}(\boldsymbol{n}^{(k)})\,\mathcal{E}^T = \frac{1}{N}\mathcal{E}\left(\rev{\operatorname{diag}(p^k)}-p^kp^{k\top}\right)\mathcal{E}^\top.
\end{align}
We assume $N$ is sufficiently large such that \(p(\varphi \mid k)\) is well-approximated by a \(D\)-dimensional Gaussian with mean and covariance given above. To estimate the mutual information, the integration in \eqref{eq:tv_info_first} is approximated via Monte-Carlo sampling (1000 samples) from the Gaussian mixture, and \eqref{eq:tv_info_last} is evaluated directly from the entropy of a Gaussian.

For a given model, we estimate the embedding function $\lambda$ computed by MLP1 by caching the output of the layer at the last state of 2-state sequences representing each possible pair. The information content is estimated for a sequence length of \(N=128\) (i.e., midway through the maximum sequence length) and for embeddings projected down to a PC space that preserves 90\% of the variance. This space was found to have dimension \(\leq 32\) at all checkpoints for the \(K\) range shown in Figure \ref{fig:task_vector}c. We omit the \(K=64\) setting for clarity, which displays an anomalous reduction in estimated information late in training due to a falloff of the PC space dimension that is likely an artifact of the method for estimating the embedding function.

\section{Memorization to generalization transition}
\label{sec:mem_to_gen}

\begin{figure}[h]
\begin{nolinenumbers}
\centering
\includegraphics[]{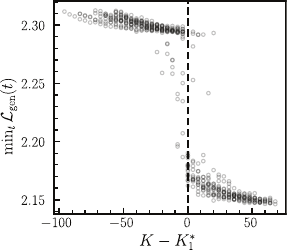}
    \caption{The minimum achieved value of the generalization loss \(\mathcal{L}_\text{gen}\) drops sharply  as the data diversity crosses over \(K_1^\ast\). As in Figure~\ref{fig:K1}a, each seed has been shifted along the horizontal axis by the value of \(K_1^\ast\) determined for that seed by the \(\phi_\beta^{(2)}\) threshold criteria. Results derived using the same models used for Figure~\ref{fig:K1}a.}
    \label{fig:k1_add}
\end{nolinenumbers}
\end{figure}

We find a critical data diversity $K_1^*$ below which the model transitions from \(\pgo\) to \(\pmo\) and above which the model instead abruptly transitions to \(\pgt\). Our findings (detailed below) indicate a kinetic mechanism for this transition.

We specify the transition point $K_1^*$ as follows. The induction-head order parameter $\phi_\beta$ is nonzero only in $\pgt$, and its maximum value over training is approximately sigmoidal as the data diversity crosses through $K_1^*$ with the inflection point occurring approximately when $\phi_\beta = \phi_\beta^* = 0.45$. Given these observations, we consider a transformer to be in $\pgt$ when $\phi_\beta>\phi_\beta^*$. We compute $K_1^*$ for a transformer by performing a binary search in the data diversity with the condition that $K>K_1^*$ if $\max \phi_\beta(t) > \phi_\beta^*$ over $10^4$ training iterations. To check the validity of this criteria, we measure the minimum value of the generalization loss \(\mathcal{L}_\text{gen}\) achieved over training as the data diversity crosses over \(K_1^\ast\) and find that the minimum \(\mathcal{L}_\text{gen}\) drops sharply as the data diversity crosses \(K_1^\ast\) (Figure~\ref{fig:k1_add}).

The kinetic mechanism driving \(K_1^\ast\) can be understood from a view of the transformer as constituting a mixture-of-experts. These experts are realized as subcircuits implementing 1-Mem and 2-Gen that develop in parallel in the transformer, and which have weighted contributions to the next-state prediction. The subcircuits are in competition in the sense that improved performance of one circuit reduces the gradient driving the development of the other. In such a case, one would expect the transition to be highly sensitive to the learning kinetics of the two subcircuits. To this end, we developed two experiments to independently perturb the kinetics of each subcircuit.

\subsection{Task injection perturbation}
To implement a memorizing predictor, the model must in part learn to differentiate between sequences from different tasks. The rate at which the model learns this can be increased by providing perfect task information to the model. To do this, we allowed the model to learn $D$-dimensional embeddings for each chain that were injected into \rev{MLP2} via addition to the original input vector to indicate the ground truth generating chain. This task injection perturbation is seen to increase the value of $K_1^*$ for the model by a factor \(\sim 10\) (Figure~\ref{fig:K1}c).

\subsection{Gradient reweighting perturbation}
We sought to perturb the model by modifying the effective learning rate for the \(2\)-Gen circuit independent of the 1-Mem circuit. One way to accomplish this is by reducing the rate at which the first attention layer learns to attend to the previous state. We implemented a modified attention head in Python with a fixed weight factor \(w \in [0,1]\) such that the previous-state attention weights include a term that does not contribute to the gradient
\begin{equation}
	A_{i,i-1}^{(1)} = w \cdot A_{i, i-1}^{\prime(1)} + (1-w) \cdot \text{detach}\left( {A}_{i,i-1}^{\prime(1)} \right) \quad \text{for } i \in [2, N]
\end{equation}
where \(\text{detach}(\cdot)\) denotes detachment from the computational graph used for automatic differentiation. As a result, gradients of these attention weights during backpropagation are reduced by a factor of \(w\) while the attention weights used for the forward pass are unmodified. Using this method to decrease the previous-state attention weights' learning rate by a factor of \(w=1/10\) relative to that of the rest of the model is seen to increase $K_1^*$ by a factor \(\sim 2\) (Figure~\ref{fig:K1}c). During the numerical simulations performed to obtain this result, we increased the total training iterations by a factor of $1/w$ to broadly account for the slower dynamics.

\begin{figure}[h]
\begin{nolinenumbers}
\centering
\includegraphics[]{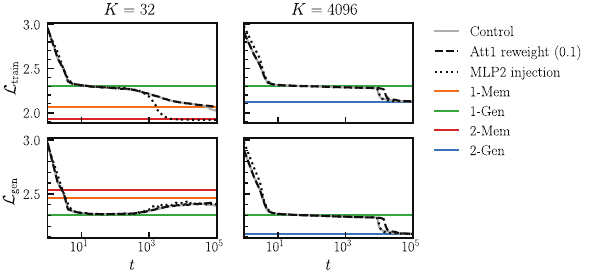}
    \caption{
    \rev{
    The kinetic perturbations show the desired effect on the target circuit, and little off-target effect. Specifically, the task injection perturbation speeds up learning of 1-Mem for small $K < K_1^\ast$ while learning of 2-Gen at large $K \gg K_1^\ast$ remains unaffected. Likewise, the gradient reweighting perturbation delays learning of 2-Gen while learning of 1-Mem is not affected. Dynamics are shown for vanilla SGD.
    }
    }
    \label{fig:kinetic_demo}
\end{nolinenumbers}
\end{figure}

\rev{
To check the specificity of each perturbation's effect on the target circuit kinetics, we compare the learning dynamics at small and large $K$ (Figure~\ref{fig:kinetic_demo}).
}

\section{Transient generalization and partial memorization}
\label{sec:partial_mem}

\begin{figure}[h]
\begin{nolinenumbers}
\centering
\includegraphics[width=0.45\linewidth]{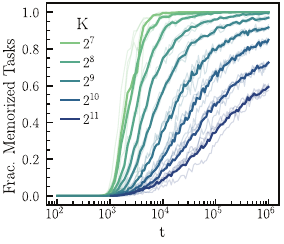}
    \caption{The fraction of memorized tasks displays sigmoidal training dynamics across data diversities. Individual results for 4 seeds are plotted as thin traces, and the mean is plotted by a thick line.}
    \label{fig:partial_mem_dynamics}
\end{nolinenumbers}
\end{figure}

We hypothesize the existence of a critical data diversity \(K_2^*\), above which the model will converge to \(\pgt\) and never enter \(\pmt\). In an attempt to estimate this transition point, we measured the length of the interval \(\Delta\tau_K \equiv \tau_{\text{2-Mem}} - \tau_{\text{2-Gen}}\) between the sharp drop in the loss at the onset of \(\pgt\) and the first crossing of the model loss with the 2-Gen predictor \(\mathcal{L}_\text{train} < \mathcal{L}^{2\text{-Gen}}\). This quantity is observed to diverge as \((K_2^\ast - K)^{-\gamma}\) with \(K_2^\ast \approx 7000\) and \(\gamma \approx 2\) (Figure~\ref{fig:mem}a).

To develop a complementary picture of \(K_2^\ast\), we investigated the nature of the model memorization at large data diversities. To do so, we computed the average loss difference between the model and the 2-Gen predictor on sequences drawn from each chain. This quantity forms a per-task memorization score
\begin{equation}
	\Delta_\text{2-Mem}^k = \expval{\mathcal{L}(\hat\pi_\theta, S_{N+1}) - \mathcal{L}\left( \hat\pi^\text{2-Gen}, S_{N+1} \right)}_{S_{N+1}\sim T^{(k)}}
\end{equation}
where \(\mathcal{L}(\hat{\pi}, S_{N+1})\) is the autoregressive cross-entropy loss of predictor \(\hat{\pi}\) on sequence \(S_{N+1}\). Thus, a negative score \(\Delta_\text{2-Mem}^k < 0\) indicates that the model outperforms the 2-Gen predictor on sequences from task \(k\). We estimated \(\Delta_\text{2-Mem}^k\) at each training checkpoint by evaluating the model and 2-Gen predictor on a batch of \(64 \times K\) training sequences. The proportion of tasks with \(\Delta_\text{2-Mem}^k < \epsilon\) for small \(\epsilon\) (detailed below) shows sigmoidal training dynamics across a range of \(K\) \rev{(Figure~\ref{fig:partial_mem_dynamics})}, and we show in Figure \ref{fig:mem}(\rev{b}) the asymptotic values of sigmoid functions fit to these dynamics. These data reveal that for data diversities \(K \gtrsim 2^8\) the model can only memorize a subset of the training chains, and a linear extrapolation of these data suggests that the model will not memorize any chains for \(K \gtrsim 5000\). This result is consistent with the estimate for \(K_2^\ast\) obtained from the diverging memorization time in the previous paragraph.

The model in \(\pgt\) may exhibit per-chain over-performance relative to the 2-Gen predictor due to biases in the model predictions. To account for this possibility, we train 3 models at \(K = \infty\) for \(10^6\) iterations and evaluate the memorization score \(\Delta_\text{2-Mem}^k\) of the final checkpoint on \(10^4\) randomly fixed tasks. The distribution of scores for each model each have mean and standard deviation \(\approx 10^{-3}\), indicating that typical fluctuations of \(\Delta_\text{2-Mem}^k\) are \(\sim 10^{-3}\) for a \(\pgt\) model. Therefore, we require that \(\Delta_\text{2-Mem}^k < \epsilon\) with \rev{\(\epsilon = -10^{-3}\)} for a chain to be considered memorized. Removing the threshold by setting \(\epsilon = 0\) modifies the slope of the decay in Figure \ref{fig:mem}(\rev{b}) but leaves the qualitative features unchanged.

\section{Minimal model for 2-point memorization}\label{app:minimal_network_mem}

\subsection{Path expansion in $\pmt$}

The circuit-tracing results for \(\pmt\) in Figure~\ref{fig:circuits_raw} indicate that the full transformer computation can be reduced to the dominant information flow shown schematically in Figure~\ref{fig:mem}(c). This reduced path can be written as
\begin{align}
x_n^{(0)} &= W_E x_n, \\
y_n^{(1)} &= x_n^{(0)} + \mathrm{Att}^{(1)}\!\left(x^{(0)}_{\leq n}\right), \\
x_n^{(1)} &= x_n^{(0)} + \mathrm{MLP}^{(1)}\!\left(y_n^{(1)}\right), \\
y_n^{(2)} &= \mathrm{MLP}^{(1)}\!\left(y_n^{(1)}\right) + \mathrm{Att}^{(2)}\!\left(x^{(1)}_{\leq n}\right), \\
x_n^{(2)} &= \mathrm{MLP}^{(2)}\!\left(y_n^{(2)}\right), \\
x_n^{(3)} &= \mathrm{Linear}\!\left(x_n^{(2)}\right).
\end{align}

Our mechanistic hypothesis may be summarized as follows. Att1 extracts the preceding state and MLP1 embeds the ordered pair $(x_{n-1},x_n)$ into a high-dimensional representation that distinguishes different tasks. Next, Att2 pools these pair embeddings across the sequence to form a task vector. Finally, MLP2 combines this task vector with the current-state information encoded by MLP1 to retrieve the transition probabilities conditioned on the current state. Under this hypothesis, the computation can be further simplified without loss of essential functionality.

\subsection{Minimal network construction}

First, we assume that Att1 extracts the preceding state and maps it to a subspace orthogonal to that of the current state. This allows us to replace
\begin{equation}
y_n^{(1)} = x_n^{(0)} + \mathrm{Att}^{(1)}\!\left(x^{(0)}_{\leq n}\right)
\;\longrightarrow\;
y_n^{(1)} = x_n^{(0)} \oplus x_{n-1}^{(0)} ,
\end{equation}
where $\oplus$ denotes concatenation into orthogonal subspaces.

Second, we assume that Att2 performs a uniform pooling operation over the pair embeddings produced by MLP1, while the current-state information is passed directly to MLP2. This yields
\begin{equation}
y_n^{(2)} =
\mathrm{MLP}^{(1)}\!\left(y_n^{(1)}\right)
+
\mathrm{Att}^{(2)}\!\left(x^{(1)}_{\leq n}\right)
\;\longrightarrow\;
x_n^{(0)} \oplus
\left\langle
\mathrm{MLP}^{(1)}\!\left(x^{(0)}_{i+1} \oplus x^{(0)}_{i}\right)
\right\rangle_{i<n}.
\end{equation}

Combining these assumptions, the resulting minimal network is
\begin{align}
x_n^{(0)} &= W_E x_n, \\
y_n^{(1)} &= x_n^{(0)} \oplus x_{n-1}^{(0)}, \\
x_n^{(1)} &= \mathrm{MLP}^{(1)}\!\left(y_n^{(1)}\right), \label{app:minembed} \\
y_n^{(2)} &= x_n^{(0)} \oplus \left\langle x_i^{(1)} \right\rangle_{i\le n}, \\
x_n^{(2)} &= \mathrm{MLP}^{(2)}\!\left(y_n^{(2)}\right), 
\end{align}
which is equivalent to the architecture depicted in Figure~\ref{fig:mem}c upon the replacement of the two attention circuits by their corresponding functional forms. In the last step, the final linear readout has been absorbed into MLP2, which now directly outputs the C-dimensional logit.
For this minimal model, $W_E$ is initialized randomly from a Gaussian distribution and kept fixed throughout training.

Relative to the full transformer, the minimal model introduces a single additional parameter: the dimension $D_\varphi$ of the task vector
\begin{equation}
\varphi = \frac{1}{n-1}\sum_{i=2}^n\left\langle x^{(0)}_{i-1} \oplus x^{(0)}_{i} \right\rangle,
\end{equation}
where $\varphi \in \mathbb{R}^{D_\varphi}$.

Unless otherwise stated, the reference model discussed in this section has an MLP2 with two hidden layers, in a departure from the full transformer model. We use autoregressive training for the minimal model with the maximum sequence length $N=256$ to match the training process described in Appendix \ref{sec:training_AR}.

\subsection{Phase diagram}

Figures~\ref{fig:mem}d and~\ref{fig:mem}f show that the minimal model training loss can fall below the 2-Gen baseline $\mathcal{L}^{\mathrm{2\text{-}Gen}}$ while the generalization loss remains significantly above it. This separation demonstrates that the minimal network successfully reproduces the 2-Mem predictor.

As data diversity $K$ increases, Figure~\ref{fig:mem}d exhibits a clear crossover behavior: for small $K$, the minimal network outperforms the 2-Gen predictor, whereas for sufficiently large $K$ it does not. Because the minimal network does not implement the induction-head mechanism required for 2-Gen prediction (which is present in the full transformer), this crossover is consistent with the existence of a critical data diversity $K_2^*$ separating $\pmt$ and $\pgt$ in the full model. This agreement suggests that the minimal network captures the essential structure of the \(\pmt\) transformer and can therefore serve as a useful framework for analyzing the mechanisms underlying 2-Mem prediction. 

We further observe that MLP2 can implement the 2-Gen prediction in certain regimes. In the limit $K \to \infty$, both training and generalization losses approach the 2-Gen baseline, provided that the task-vector dimension $D_\varphi$ is sufficiently large. A large $D_\varphi \sim C^2$ is required to encode the complete empirical transition matrix of the sequence in one vector.

\begin{figure}[h]
\begin{nolinenumbers}
\centering
\includegraphics[width=\linewidth]{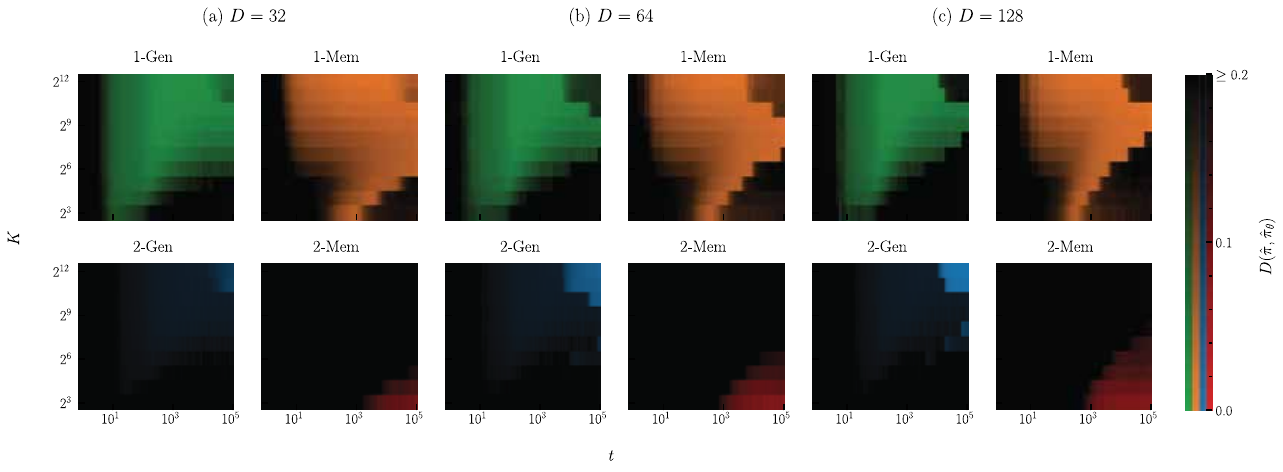}
    \caption{
    \rev{
    Behavioral readouts of the full transformer trained with the second attention block restricted to uniform attention. As the model dimension $D$ crosses over $C^2$ in panel (c), the full four phases appear. Note that these diagrams include up to $K = 2^{12}$.
    }
    }
    \label{fig:pool_phases}
\end{nolinenumbers}
\end{figure}

\rev{
Indeed, if we restrict the second attention block in the full transformer model to uniform attention, we observe the usual four-phase picture when the model dimension $D$ becomes sufficiently large (Figure~\ref{fig:pool_phases}).
}

\subsection{Dependence of $\hat{K}_2^*$ on encoder and decoder capacity}

We denote the estimated critical data diversity for the minimal model by $\hat{K}_2^*$ to distinguish it from $K_2^*$ in the full model.

We hypothesize that MLP1 is primarily responsible for extracting a task vector from the sequence, while MLP2 must memorize the collection of task-specific transition matrices and select the appropriate one conditioned on the task vector. Consequently, the capacities of both MLPs can act as bottlenecks and influence the critical data diversity $\hat{K}_2^*$.

In Figure~\ref{fig:mem}(e), we vary the depths of MLP1 and MLP2 to have one or two hidden layers while keeping their widths fixed. For each architecture, we estimate $\hat{K}_2^*$ via binary search using the threshold
\begin{equation}
\left\langle
\mathcal{L}_{\mathrm{train}} - \mathcal{L}^{\mathrm{2\text{-}Gen}}
\right\rangle_{1000}
< 0,
\end{equation}
where the average is taken over 1000 iterations with a $4\times10^5$-iteration run.

We find that \rev{$\hat{K}_2^*$} depends only weakly on the depth of MLP1. For a fixed MLP2 depth, increasing MLP1 from one to two layers leaves $\hat{K}_2^*$ essentially unchanged (see Figure~\ref{fig:mem}e). In contrast, increasing the depth of MLP2 substantially enlarges \rev{$\hat{K}_2^*$}, raising it from $\sim 10^2$ to $\sim 10^3$-$10^4$. This indicates that memorization capacity in MLP2 is the primary bottleneck for sustaining $\pmt$ at large $K$. This behavior aligns with the intuition that constructing pair embeddings (the role of MLP1) is simple compared to retrieving the transition matrix given the chain identity (the role of MLP2), which evidently demands significantly greater representational capacity. We also observe a dependence on the embedding dimension, with \(\hat{K}_2^\ast\) increasing more rapidly with \(D_\varphi\) for $D_\varphi \gtrsim 60$. The scaling between the capacity of the neural network and the embedding dimension is thus non-trivial.

\end{document}